\documentclass[10pt,english]{article}
\usepackage[T1]{fontenc}
\usepackage[latin9]{inputenc}
\usepackage{geometry}
\usepackage[active]{srcltx}
\usepackage{babel}
\usepackage{verbatim}
\usepackage{float}
\usepackage{mathtools}
\usepackage{bm}
\usepackage{amsmath}
\usepackage{amssymb}
\usepackage{graphicx}
\usepackage[unicode=true,
 bookmarks=false,
 breaklinks=false,pdfborder={0 0 1},colorlinks=false]
 {hyperref}
\hypersetup{
 colorlinks,linkcolor=red,anchorcolor=blue,citecolor=blue}

\makeatletter

\usepackage{natbib}
\usepackage{amsthm}
\usepackage{dsfont}
\usepackage{array}
\usepackage{mathrsfs}
\usepackage{comment}
\usepackage{color}
\usepackage{xfrac}
\usepackage{enumitem}
\usepackage{algorithm}
\usepackage[noend]{algpseudocode}
\usepackage{arydshln}
\usepackage[outdir=./]{epstopdf}

\makeatother

\theoremstyle{plain} 
\newtheorem{lemma}{Lemma} 

\newtheorem{theorem}{Theorem}

\theoremstyle{remark}
\newtheorem{assumption}{Assumption} 
\newtheorem{definition}{Definition}

\newtheorem*{example}{Example}

\newtheorem{remark}{Remark}

\allowdisplaybreaks

\newcommand{\argmin}{\mathop{\mathrm{argmin}}}
\newcommand{\argmax}{\mathop{\mathrm{argmax}}}

\newcommand{\defn}{\coloneqq}
\newcommand{\tree}{\mathsf{tr}}
\newcommand{\lf}{\mathsf{lf}}
\newcommand{\adp}{\mathsf{a}}
\newcommand{\dm}{\mathsf{dm}}
\newcommand{\se}{\mathsf{se}}
\newcommand{\cA}{\mathcal{A}}

\newcommand{\cD}{\mathcal{D}}
\newcommand{\cE}{\mathcal{E}}

\newcommand{\bE}{\mathbb{E}}
\newcommand{\bP}{\mathbb{P}}
\newcommand{\hatbeta}{\widehat{\beta}}
\newcommand{\olc}{\overline{c}}

\begin{document}

\title{Transfer Learning for Contextual Multi-armed Bandits}

\author{
	Changxiao Cai\thanks{Department of Industrial and Operations Engineering, University of Michigan, Ann Arbor, MI 48109, USA.} \\
%; email: \texttt{ccai@princeton.edu} 	
 UMich \\
 	\and
	T.~Tony Cai\thanks{Department of Statistics and Data Science,  Wharton School, University of Pennsylvania, Philadelphia, PA 19104, USA.}\\
	UPenn  \\
  	\and 
	Hongzhe Li\thanks{Department of Biostatistics, Perelman School of Medicine, University of Pennsylvania, Philadelphia, PA 19104, USA.
	}\\
	UPenn  \\ 
	}
	
\date{November 2022; ~~ Revised: January 2024}	

\maketitle

\begin{abstract}

Motivated by a range of applications, we study in this paper the problem of transfer learning for nonparametric contextual multi-armed bandits under the covariate shift model, where we have data collected from source bandits before the start of the target bandit learning. The minimax rate of convergence for the cumulative regret is established and a novel transfer learning algorithm that attains the minimax regret is proposed. The results quantify the contribution of the data from the source domains for learning in the target domain in the context of nonparametric contextual multi-armed bandits. 

In view of the general impossibility of adaptation to unknown smoothness, we develop a data-driven algorithm that achieves near-optimal statistical guarantees (up to a logarithmic factor) while automatically adapting to the unknown parameters over a large collection of parameter spaces under an additional self-similarity assumption. A simulation study is carried out to illustrate the benefits of utilizing the data from the source domains for learning in the target domain.
\end{abstract}

\medskip


\tableofcontents{}

\section{Introduction}
\label{sec:intro}
   
Inspired by the human intelligence of leveraging prior experiences to tackle novel problems, \emph{transfer learning}, which aims to improve the learning performance in a target domain by transferring the knowledge contained in different but related source domains, has become an active and promising area of research in machine learning.  Transfer learning has achieved significant success in a wide range of practical applications such as computer vision \citep{quattoni2008transfer,kulis2011you,li2013learning}, genomic and genetic studies \citep{wang2019data,peng2021integration}, and medical imaging \citep{raghu2019,YU2022230},  to mention a few. We refer interested readers to \citet{pan2009survey,weiss2016survey} for detailed surveys on transfer learning.
Motivated by the success in these applications, substantial progress has also been made recently in the theoretical quantification for transfer learning in supervised and unsupervised settings. A partial list of examples includes classification \citep{cai2021transfer,kpotufe2021marginal,maity2020minimax,reeve2021adaptive}, high-dimensional linear regression \citep{li2020transfer}, graphical model \citep{li2022transfer}, and nonparametric regression \citep{Cai2022TransferNPR,ma2022optimally,pathak2022new}.
  
In this paper, we consider transfer learning for nonparametric contextual multi-armed bandits. Since the seminal formulation in \citet{robbins1952some}, the multi-armed bandit (MAB) and its various extensions have been widely used in numerous fields related to sequential decision-making, including personalized medicine \citep{tewari2017ads,rabbi2018feasibility,zhou2019tumor,shrestha2021bayesian,demirel2022escada}, recommendation system \citep{li2010contextual,kallus2020dynamic}, and dynamic pricing \citep{rothschild1974two,kleinberg2003value,wang2021multimodal}. 
In the classical nonparametric contextual multi-armed bandit problem, a decision-maker sequentially and repeatedly chooses an arm from a set of available arms and receives a random reward generated by the selected arm. The goal is to develop an arm selection policy that maximizes the expected cumulative rewards over a finite time horizon. Motivated by the common scenario where the decision-maker often has access to side information to assist arm selection, covariates are introduced to encode the features that affect the reward yielded by each arm at each time step.  
The expected reward of each arm conditioned on the context is assumed to follow a nonparametric form, allowing for a more flexible and robust formulation in real-world applications. 

However, collecting enough reward feedback to design an optimal arm selection strategy is often challenging in practice.
For instance, the contextual multi-armed bandit framework is widely used in precision medicine that aims to tailor the medical care to each patient \citep{rindtorff2019biologically,zhou2019tumor}. Every time a patient visits, a healthcare provider needs to determine a treatment based on the patient's profile, including genetics, biomarkers, environment, and demographic information. The objective is to optimize the post-treatment health outcomes of all patients. In this case, patient profiles, treatments, and health outcomes correspond to covariates, arms, and rewards, respectively. However, there is no shortage of cases where biomedical data of minority populations are underrepresented in certain healthcare institutions \citep{sudlow2015uk}. Given this limited availability of data in clinical research, it is common to resort to the healthcare records of other patients with similar characteristics. In such a scenario, the task of transfer learning for contextual multi-armed bandits naturally arises.

In addition to applications in precision medicine,  contextual multi-armed bandits are also frequently used in online recommendation systems that seek to learn dynamically the preferences of an individual customer for a collection of products based on the demographics and purchase histories \citep{agrawal2019mnl,kallus2020dynamic}. Since each user can only purchase a small set of products, the availability of transactional data is often limited in practice. Therefore, it is natural to explore the information of different but related customers, in order to better predict the possibility of an individual customer purchasing a specific product. 
Similarly, anomaly detection systems often rely on a limited number of interactions with human experts for verification to maximize accurate anomaly detection. Due to its trade-off between exploration (e.g.~investigation of various anomalies to improve prediction) and exploitation (e.g.~queries of the most suspicious one), anomaly detection has also been formulated by a contextual multi-armed bandit framework \citep{ding2019interactive,soemers2018adapting}. In credit card fraud identification systems, if the account history of a single card holder is short, it would be advisable to utilize the information of similar types of transactions and customers to increase the detection accuracy of fraudulent transactions.

In this work, we consider the following setting of transfer learning for contextual multi-armed bandits. Let $Q$ and $P$ be two probability distributions over $[0,1]^{d}\times[0,1]^{K}$ that generate a sequence of independent random vectors $(X_{t},Y^{(1)}_{t},\cdots,Y^{(K)}_{t})_{t \geq 1}$ associated with a contextual $K$-armed bandit. Here, $(X_{t})_{t \geq 1}$ is a sequence of i.i.d.~random vectors in $\mathcal{X}\coloneqq[0,1]^{d}$ representing the covariates. For each $1\leq k \leq K$ and $t \geq 1$, $Y^{(k)}_{t}$ is a random variable in $[0,1]$ indicating the reward yielded by arm $k$ at time $t$, with conditional expectation given by $\mathbb{E}[ Y_{t}^{(k)} \,|\, X_{t}] = f_{k}(X_{t})$, where $f_{k}:\mathcal{X}\rightarrow[0,1]$ is referred to as a reward function.
The bandit game operates as follows: at each time step $t$, given side information $X_t$, the decision-maker pulls one of the $K$ arms, denoted by $\pi_{t}$, and receives the random reward $Y_t^{(\pi_{t})}$.
Suppose that before the $Q$-bandit game starts, we have access to pre-collected samples $\{ (X_{i}^{P}, \pi_{i}^{P}, Y_{i}^{(\pi_{i}^{P})}) \}_{i=1}^{n_{P}}$, generated from a source bandit with underlying distribution $P$. 
Throughout the paper, $Q$ refers to the target distribution about which we wish to make statistical inferences; $P$ stands for the source distribution from which we have collected data to improve the decision-making under $Q$. We use the $Q$-bandit (resp.~$P$-bandit) to represent the bandit with distribution $Q$ (resp.~$P$). In addition, we use superscripts $Q$ and $P$ to refer to quantities associated with the $Q$-bandit and the $P$-bandit, respectively.
Our goal is to design a policy $\pi = \{\pi_{t} \}_{t \geq 1}$ that maximizes the expected cumulative rewards under the target distribution $Q$.
% Equivalently, 
The performance of the policy $\pi$ can be measured by its cumulative regret over $n_{Q}$ time steps, given by 
\begin{align}
\label{eq:regret}
    \mathbb{E} [ R_{n_{Q}} (\pi) ] \coloneqq \mathbb{E} \bigg[ \sum_{t=1}^{n_{Q}} \Big(Y_{t}^{Q,(\pi^{\star}(X_t^{Q}))} (X_t^{Q}) - Y_{t}^{Q,(\pi_{t}(X_t^{Q}))} (X_t^{Q})\Big) \bigg],
\end{align}
where $\pi^{\star}$ is the oracle policy with complete knowledge of the reward functions $\{ f_{k} \}_{k=1}^{K}$.
One can expect that as long as distributions $P$ and $Q$ are similar, the source data from the $P$-bandit can improve the decision-making in the $Q$-bandit. Therefore, it is natural to quantify the improvement in the cumulative regret, which can be viewed as the amount of information transferred from the source distribution $P$ to the target distribution $Q$. 

This paper focuses on transfer learning under the covariate shift model, where the marginal distributions of covariates $P_{X}$ and $Q_{X}$ differ (i.e.~$P_{X}\neq Q_{X}$), but the conditional reward distributions of $Y^{(k)}$ given $X$ are identical under $P$ and $Q$ (i.e.~$P_{Y^{(k)}|X}=Q_{Y^{(k)}|X}$) for all $1\leq k\leq K$. This framework is well motivated by many practical applications, typically arising from the scenarios when the same study is conducted among different populations. For instance, healthcare providers in a hospital may utilize medical records from other healthcare centers to better guide medical treatments. While the patient characteristics (captured by the marginal context distributions) tend to differ across different hospitals, given the same patient profile, the effects of the same treatment (which can be modeled as the conditional reward distributions) in various medical institutions can be identical in many cases. Therefore, it is natural to model this scenario as a covariate shift.

The covariate shift model hinges on the characterization of the similarity between the marginal distributions $P_{X}$ and $Q_{X}$. Various assumptions on the similarity have been proposed in the literature. In the present paper, we adopt the concept of \emph{transfer exponent}---introduced in \citet{kpotufe2021marginal} to study transfer learning for nonparametric classification---that measures the discrepancy between $P_{X}$ and $Q_{X}$ in terms of the ball mass ratio of the respective distributions.  It is assumed that there exists a transfer exponent $\gamma \geq 0$ such that $P_{X}(B(x, r)) \gtrsim r^{\gamma} Q_{X}(B(x, r))$ holds for any $\ell_{\infty}$-ball of center $x \in \mathcal{X}$ and radius $r \in (0,1]$ (see Definition~\ref{def:transfer}). Informally, the transfer exponent $\gamma$ gauges how locally singular $P_{X}$ is with respect to $Q_{X}$. When $\gamma$ is small, this condition ensures that the source data adequately covers important regions under the target distribution $Q$ (i.e.~with large $Q_X$ mass), thus facilitating the transfer of information from the source domain. In a recent study by \citet{suk2021self}, contextual multi-armed bandits were considered in a setting where the distribution of covariates changes over time. The authors employed a similar framework to capture the distribution shift. They proposed an algorithm that provably achieves the near-optimal minimax regret while automatically adapting to the unknown change point in time and covariate shift level. However, the results therein suffer some deficiencies. First, the proposed algorithm is designed for Lipschitz reward functions (i.e.~smoothness parameter $\beta = 1$, see Section~\ref{sec:Problem-formulation} for the formal definition). Therefore, it falls short of accommodating the general settings $\beta \neq 1$. Moreover, this limitation raises a more challenging question---is it possible to design a data-driven procedure that can adapt to the smoothness level, which is typically \emph{a priori} unknown in practice?

Moving beyond concerns about smoothness, it is noteworthy that \citet{suk2021self} focused on a scenario where the decision-maker is allowed to explore the bandit game amid a covariate shift actively. However, it is more common in practice that one does not have the freedom to interact with the source bandit. Instead, one has to rely on a fixed pre-collected batch dataset. In this setup, it is natural to expect a potentially larger regret, as the arm selection policy in the source dataset might be uninformative (e.g.~a suboptimal arm is predominantly selected in the source dataset). To account for this challenge, we introduce an \textit{exploration coefficient} $\kappa$ (see Definition~\ref{def:explore-coef}) that quantifies the extent to which each arm is explored in the source bandit. Specifically, the exploration coefficient $\kappa$ ensures that each arm is pulled with probability at least $\kappa / K$ across the covariate space in the source dataset. Consequently, when $\kappa$ is not vanishingly small, each arm is explored more or less sufficiently, thereby enabling the extraction of valuable information about the reward functions from the source domain. A natural question arising from such a scenario is: what is the minimax regret in the transfer learning setting when dealing with pre-collected batch data? 
Finally, we note a logarithmic gap exists between the upper and lower bounds in \citet{suk2021self}, and it remains unknown whether the minimax lower bound on the regret can be attained. 

In the present paper, we aim to address the following questions: given a pre-collected source dataset, what is the minimax rate of convergence of the regret for nonparametric contextual multi-armed bandits in the covariate shift setting? Can we design a rate-optimal policy that achieves the minimax regret?
Moreover, is it possible to develop a data-driven procedure that achieves near-optimal statistical guarantees while at the same time automatically adapting to the unknown smoothness of the reward functions and the covariate shift of the source distribution? Encouragingly, the answers to these questions are affirmative.

\subsection{Main Contribution}

Our main contribution is twofold. We first establish the minimax rate of convergence of the cumulative regret for nonparametric contextual multi-armed bandits under the covariate shift model. In addition to the standard assumptions that the reward functions are $\beta$-H\"{o}lder (see Assumption~\ref{assumption:smooth}) and the target distribution $Q$ satisfies a margin assumption with parameter $\alpha$ (see Assumption~\ref{assumption:margin}), we assume that the source and target distributions $P$ and $Q$ satisfy transfer learning conditions with transfer exponent $\gamma$ and exploration coefficient $\kappa$ (see Definitions~\ref{def:transfer} and \ref{def:explore-coef}). 
Given $n_{P}$ source samples collected from the source bandit, we show that the minimax regret over $n_Q$ time steps in the target $Q$-bandit is of order $ n_{Q}\big(n_{Q} + (\kappa n_{P})^{\frac{d+2\beta}{d+2\beta+\gamma}}\big)^{-\frac{\beta(1+\alpha)}{d+2\beta}}$.  
In the classical setting where one has no auxiliary data from the source domain, i.e.~$n_{P} = 0$, the minimax regret is known to be of order $n_{Q}^{1-\frac{\beta(1+\alpha)}{d+2\beta}}$.
Therefore, the term $(\kappa n_{P})^{\frac{d+2\beta}{d+2\beta+\gamma}}$ in the minimax regret captures the contribution from the source data to the target bandit, which depends on the amount of covariate shift between $P_{X}$ and $Q_{X}$ as well as the degree of arm exploration in the source dataset.

We also develop a novel transfer learning algorithm and prove that it achieves the minimax regret. However, the constructed procedure depends on the knowledge of the smoothness and transfer learning parameters. Unfortunately, it has been widely recognized that adaptation to unknown smoothness is generally infeasible in nonparametric bandit problems \citep{locatelli2018adaptivity,gur2022smoothness,cai2022stochastic}. To this end, we choose to focus on the bandits with reward functions that satisfy the self-similarity condition---an assumption extensively used in the statistics literature \citep{picard2000adaptive,gine2010confidence}. We develop a data-driven algorithm and show that it simultaneously achieves the near-optimal minimax regret at the penalty of an additional logarithmic factor over a large class of parameter spaces. Moreover, we demonstrate that the self-similarity assumption does not decrease the complexity of the problem by establishing the minimax lower bound that remains the same as in the general case.

\subsection{Related works}
\paragraph*{Contextual multi-armed bandits}
The framework of contextual multi-armed bandits was first introduced in \citet{woodroofe1979one}. 
Among the parametric approaches, an important line of work assumes a linear reward function \citep{abe1999associative,auer2002using,bastani2020online,goldenshluger2013linear}. 
In this setup, \citet{bastani2021mostly} proposed a rate-optimal greedy strategy with regret logarithmic in the length of the time horizon. %provided a margin condition holds.
For nonparametric contextual multi-armed bandits, it is typical to assume H\"older smooth reward functions. 
\citet{yang2002randomized} developed a greedy policy and showed that its regret goes to zero as the time horizon tends to infinity. 
\citet{rigollet2010nonparametric} studied the two-armed bandits and proposed an upper-bound-confidence (UCB) type policy that attains a near-optimal minimax regret. This result was further refined in \citet{perchet2013multi} where a rate-optimal policy called ABSE was proposed in the multi-armed setting. Additionally, \citet{reeve2018k} combined a UCB-type policy with the nearest neighbor method to design a near-optimal algorithm capable of adapting to the low intrinsic dimension of contexts. It is worth noting that all the aforementioned methods are tailored for $\beta$-H\"{o}lder smooth reward with $\beta \in (0,1]$. \citet{hu2019smooth} extended the theory to accommodate smoother reward functions with $\beta>1$.

\paragraph*{Adaptivity}
It has been demonstrated in various bandit settings that adaptation to the unknown smoothness of reward functions is generally impossible. This means that no policy can achieve the minimax regrets simultaneously over different classes of reward functions \citep{locatelli2018adaptivity,gur2022smoothness,cai2022stochastic}. We note that this phenomenon is closely related to the impossibility of constructing adaptive confidence intervals in nonparametric function estimation \citep{low1997nonparametric,Cai2004adaptation,cai2012minimax}. Fortunately, adaptive statistical inference can be accomplished under certain shape constraints, such as monotonicity and convexity \citep{hengartner1995finite,dumbgen1998new,genovese2005confidence,cai2013adaptive}.
Self-similarity---first introduced by \citet{picard2000adaptive} for adaptive nonparametric confidence intervals---is another widely used condition that allows adaptivity. This concept finds applications in various fields, including density estimation \citep{gine2010confidence}, sparse regression \citep{nickl2013confidence}, and $\ell_{p}$-confidence sets \citep{nickl2016sharp}.
It was first introduced to the nonparametric contextual bandit setting by \citet{qian2016randomized}, where a UCB-type policy based on Lepski's method \citep{lepski1997optimal} was proposed and shown to achieve the minimax regret up to a logarithmic factor. The drawback, however, is that its cost of adaptation tends to infinity as the covariate dimension grows.
\citet{gur2022smoothness} improved upon this result by reducing the adaptation cost to a logarithmic factor independent of the dimension.

\paragraph*{Transfer learning}
Transfer learning has been explored using information measures such as KL-divergence and total variation to quantify the distinction between target and source distributions \citep{ben2006analysis,blitzer2007learning,mansour2009domain}. Generalization bounds are then established based on these metrics.
Despite its generality, such results are often not tight when applied to specific statistical models. Recent work has imposed more structured assumptions on the similarity between target and source distributions, such as covariate shift and posterior drift \citep{cai2021transfer,Cai2022TransferNPR, hanneke2019value,kpotufe2021marginal,maity2020minimax,reeve2021adaptive}, thereby leading to more refined theoretical guarantees.
Finally, our work is also closely related to hybrid reinforcement learning that aims to combine offline datasets with online interaction to improve statistical / computational efficiency \citep{ross2012agnostic,xie2021policy,song2022hybrid,wagenmaker2023leveraging,li2023reward,nakamoto2023cal}.

\subsection{Organization}

The rest of the paper is organized as follows. Section~\ref{sec:Problem-formulation} formulates the problem and introduces definitions and assumptions. We then establish the minimax optimal rate of the regret and develop a rate-optimal algorithm in Section~\ref{sec:Main-Results}. In Section~\ref{sec:Adaptivity}, we propose a data-driven adaptive procedure that achieves the minimax regret up to logarithmic factors. The proofs of our theorems, technical lemmas, and numerical experiments are deferred to the Supplementary Material~\citep{cai2022transfer-supp}. 
% Section~\ref{sec:Numerical-experiments} presents the numerical performance of the proposed algorithms. 
We conclude with a discussion of future directions in Section~\ref{sec:Discussion}.  

\subsection{Notation}

\label{subsec:Notations}
For any $a,b\in\mathbb{R}$, we define $a\vee b\defn\max\{a,b\}$ and $a\wedge b\defn\min\{a,b\}$.
We denote by $\| \cdot \|_2 $ and $\| \cdot \|_\infty$ the $\ell_2$ norm and the $\ell_\infty$ norm, respectively. The notation $B_\infty(x,r) \defn \{y : \| y - x \|_\infty \leq r \}$ refers to the $\ell_\infty$ ball of center $x$ and radius $r$, and we define the shorthand $B(x,r) \defn B_\infty(x,r)$. Denote by $[K]:=\{1,2,\cdots,K\}$. We use $\mathds{1}\{ \cdot \}$ to represent the indicator function, and we define $\log^{+}(x)\defn\log(x)\vee1$. Let $\mathsf{supp}(\cdot)$ denote the support of any probability distribution. For any distributions $P, Q$, the notation $\mathsf{KL}(P\|Q)$ stands for the KL-divergence. For any $a \in \mathbb{R}$, denote by $\lfloor a \rfloor$ (resp.~$\lceil a \rceil$) the largest (resp.~smallset) integer that is strictly smaller (resp.~larger) than $a$.  The notation $\mathbb{N}$ stands for the set of the natural numbers, and we denote $\mathcal{X}\coloneqq[0,1]^{d}$.
Throughout the paper, we denote by $C$ or $c$ some constants independent of $n_P$ and $n_Q$ which may vary from line to line.

For any two functions $f(n), g(n) > 0$, the notation $f(n)\lesssim g(n)$ (resp.~$f(n)\gtrsim g(n)$) means that there exists a constant $C>0$ such that $f(n)\leq Cg(n)$ (resp.~$f(n)\geq Cg(n)$). The notation $f(n)\asymp g(n)$ means that $C_{0}f(n)\leq g(n)\leq C_{1}f(n)$ holds for some constants $C_{0},C_{1}>0$. In addition, $f(n)=o(g(n))$ means that $\limsup_{n\rightarrow\infty}f(n)/g(n)=0$, $f(n)\ll g(n)$ means that $f(n)\leq c_{0}g(n)$ for some small constant $c_{0}>0$, and $f(n)\gg g(n)$ means that $f(n)\geq c_{1}g(n)$ for some large constant $c_{1}>0$.

\section{Problem formulation}

\label{sec:Problem-formulation}

\subsection{Transfer learning for nonparametric contextual multi-armed bandits}
 
Let $Q$ be a probability distribution over $\mathcal{X}\times[0,1]^{K}$ that generates a sequence of independent random vectors $(X^{Q},Y^{Q,(1)},\cdots,Y^{Q,(K)})$.
At each time point $t$, based on the covariate $X_{t}^{Q}\in\mathcal{X}$ drawn from the marginal distribution $Q_{X}$, a decision-maker selects an arm $k\in[K]$ and receives a random reward $Y_{t}^{Q,(k)}\in[0,1]$ associated with the chosen arm according to the conditional distribution $Q_{Y^{(k)}| X_{t}^{Q}}$. We assume that for any $k\in[K]$ and $t\geq1$, the random reward $Y_{t}^{Q,(k)}$ is a random variable with conditional expectation given by
\begin{align*}
\mathbb{E}\big[Y_{t}^{Q,(k)}\,|\,X_{t}^{Q}\big]=f_{k}^{Q}(X_{t}^{Q}),
\end{align*}
where $f_{k}^{Q}:\mathcal{X}\rightarrow[0,1]$ is an unknown function called a reward function. A policy $\pi$ is a collection of functions $\{\pi_{t}\}_{t\geq1}$ where $\pi_{t}: \mathcal{X}\rightarrow[K]$ prescribe the arm to pull at time $t$.

In the context of transfer learning, we assume that the decision-maker is given a batch dataset $\mathcal{D}^{P} \coloneqq \{(X_{i}^{P},\pi_{i}^{P}, Y_{i}^{P,(\pi_{i}^{P})})\}_{i=1}^{n_{P}}$. This dataset is collected from a contextual $K$-armed bandit over $n_{P}$ rounds, of which the underlying probability distribution $P$ generates a sequence of independent random vectors $(X^{P},Y^{P,(1)},\cdots,Y^{P,(K)})\in\mathcal{X}\times[0,1]^{K}$.
Here, $X_{i}^{P}\in\mathcal{X}$ represents the covariate observed at time $i$, policy $\pi_{i}^{P}:\mathcal{X}\rightarrow[K]$ denotes the selected arm at time $i$, and $Y_{i}^{P,(\pi_{i}^{P})}$ corresponds to the observed random reward at time $i$. Similar to the $Q$-bandit, it is assumed that for any $k\in[K]$ and $i\geq1$, the random reward $Y_{i}^{P,(k)}$ of the $P$-bandit obeys
\begin{align*}
\mathbb{E}\big[Y_{i}^{P,(k)}\,|\,X_{i}^{P}\big]=f_{k}^{P}(X_{i}^{P}),
\end{align*}
for an unknown function $f_{k}^{P}:\mathcal{X}\rightarrow[0,1]$. Throughout the paper, we denote $n \coloneqq n_{Q} \vee n_{P}$.

As mentioned in the introduction, this paper focuses on the covariate shift model. To be specific, it is assumed that the marginal distributions of covariates in the $P$-bandit and $Q$-bandit are different (i.e.~$P_{X} \neq Q_{X}$) while the distributions of rewards conditioned on the covariate value are identical (i.e.~$P_{Y^{(k)}|X}=Q_{Y^{(k)}|X}$ for all $1\leq k\leq K$). In particular, the latter implies that the reward functions of the two bandits are also identical. We denote these common reward functions as $f_{k}(x)\coloneqq f_{k}^{P}(x)\equiv f_{k}^{Q}(x)$ for all $k\in[K]$ and $x\in\mathcal{X}$.

Recall that $\pi^{\star}$ is the oracle policy with access to full knowledge of the reward functions $\{f_{k}\}_{k=1}^{K}$.
It is straightforward to see that given a covariate value $x$, the oracle policy $\pi^{\star}$ selects any arm with the largest expected reward, with ties broken arbitrarily. In other words,
\begin{align*}
\pi^{\star}(x)\in\argmax_{k\in[K]}f_{k}(x).
\end{align*}
Therefore, for any policy $\pi=\{\pi_{t}\}_{t\geq1}$, the regret of $\pi$ in the $Q$-bandit defined in (\ref{eq:regret}) has the following expression:
\begin{align}
\mathbb{E}[R_{n_Q}(\pi)] & =\mathbb{E}\bigg[ \sum_{t=1}^{n_{Q}} \Big( f_{\pi^{\star}(X_t^Q)}(X_{t}^{Q})-f_{\pi_{t}(X_t^Q)}(X_{t}^{Q}) \Big) \bigg] \nonumber \\
& =\mathbb{E}\bigg[ \sum_{t=1}^{n_{Q}} \Big( \max_{k\in[K]} f_{k}(X_{t}^{Q})-f_{\pi_{t}(X_t^Q)}(X_{t}^{Q}) \Big)\bigg]. \label{eq:def-regret}
\end{align}
In the remainder of the paper, we may drop the subscript $n_Q$ whenever there is no confusion.

Finally, we would like to emphasize that the policy $\pi_{t}$ at time $t$ depends on both the observations of the $Q$-data prior to time $t$ (i.e.~$\{(X_{i}^{Q},\pi_{i},Y_{i}^{Q,(\pi_{i})})\}_{i=1}^{t-1}\cup\{X_{t}^{Q}\}$) and the complete $P$-data (i.e.~$\{(X_{i}^{P},\pi_{i}^{P},Y_{i}^{P,(\pi_{i}^{P})})\}_{i=1}^{n_{P}}$).

\subsection{Assumptions}

It is noteworthy that one cannot hope to distinguish the optimal arm of a contextual multi-armed bandit with arbitrary covariate and reward distributions. In order to guarantee provably small cumulative regrets, we impose the following model assumptions, which have become standard in the literature on nonparametric contextual multi-armed bandits \citep{rigollet2010nonparametric,perchet2013multi}.

We begin by imposing a H\"older smoothness assumption on the reward functions $\{f_{k}\}_{k=1}^{K}$ as follows.
\begin{assumption}[Smoothness]
\label{assumption:smooth}
The reward functions $\{f_{k}\}_{k=1}^{K}$ are $\left(\beta,C_{\beta}\right)$-H\"{o}lder continuous for some constants $0<\beta\leq1,C_{\beta}>0$, i.e. for any $k\in[K]$,
\begin{align*}
\big|f_{k}(x)-f_{k}(x')\big|\leq C_{\beta} \| x-x'\|_{\infty}^{\beta},\quad\forall x,x'\in\mathcal{X}.
\end{align*}
\end{assumption}

\begin{remark}
	By the equivalence of $\ell_{p}$ norms $(p\geq1)$ in $\mathcal{X}$, the results in this work continue to hold if $\ell_{\infty}$ norm is replaced with any $\ell_{p}$ norm ($p\geq1$).
\end{remark}

\begin{remark}
	Given the primary focus of this work is to illustrate the potential for reducing cumulative regrets through the utilization of source data, we confine our attention to the case $0 < \beta \leq 1 $ for simplicity of presentation. Notably, the insights and findings here can be extended to accommodate the case $\beta > 1$. A detailed discussion of this generalization is deferred to Section~\ref{sec:Discussion}.
\end{remark}

Next, it is natural to expect that the gap between the reward functions is a pivotal measure of a contextual multi-armed bandit problem's complexity. To this end, let $f_{(1)}$ (resp.~$f_{(2)}$) denote the pointwise maximum (resp.~the second pointwise maximum) of the reward functions $\{f_{k}\}_{k=1}^{K}$, namely
\begin{align*}
f_{(1)}(x) & \coloneqq\max_{k\in[K]}f_{k}(x),
\end{align*}
and
\begin{align*}
f_{(2)}(x)\coloneqq\begin{cases}
\max\limits_{k\in[K]}\big\{ f_{k}(x):f_{k}(x)<f_{(1)}(x)\big\}, & \text{if }\min\limits_{k\in[K]}f_{k}(x)\neq\max\limits_{k\in[K]}f_{k}(x),\\
f_{(1)}(x), & \text{otherwise.}
\end{cases}
\end{align*}

Equipped with these notations, we introduce the following margin assumption to quantify the interplay between the reward gap and the covariate distribution in the target bandit $Q$.
\begin{assumption}[Margin]\label{assumption:margin}
	There exist constants $\alpha\geq0,C_{\alpha}>0$ such that the reward functions $\{f_{k}\}_{k=1}^{K}$ and marginal distribution $Q_{X}$ satisfy
	\begin{align*}
		Q_X\big(0<f_{(1)}(X)-f_{(2)}(X)\leq\delta\big)\leq C_{\alpha}\delta^{\alpha},\quad\forall0<\delta\leq1.
	\end{align*}
\end{assumption}

Assumption~\ref{assumption:margin} bears a resemblance to the margin condition initially introduced in classification \citep{mammen1999smooth,tsybakov2004optimal,audibert2007fast}, and has been widely used in contextual multi-armed bandits \citep{goldenshluger2009woodroofe,perchet2013multi} and dynamic treatment regimes \citep{qian2011performance,luedtke2016statistical,shi2020breaking}.
Roughly speaking, the margin condition encodes the distribution behavior of the contexts near the decision boundary. It is easy to see that the margin condition is inherently satisfied for $\alpha = 0$ and holds for $\alpha = 1$ when $f_{(1)}(X)-f_{(2)}(X)$ has a bounded probability density near zero.
If the margin parameter~$\alpha$ is large, it implies that, with low probability, the reward gap between the optimal arm and other arms is small but bounded away from zero. This means that the reward functions of different arms are well-separated over a region of large probability mass, which, in turn, reduces the difficulty of distinguishing between the arms.

\begin{remark}
As discussed in \citet[Proposition 3.1]{perchet2013multi}, when $\alpha \beta > d$, there exists a single arm that dominates others across the entire covariate space. In such a case, a contextual multi-armed bandit problem degenerates into a static multi-armed bandit problem that falls beyond the scope of interest for this work. Therefore, we shall assume $\alpha \beta \leq d$ in the remainder of the paper.
\end{remark}

\begin{remark} \label{remark:smoothness}
Assumptions \ref{assumption:smooth} and \ref{assumption:margin} are commonly found in the nonparametric contextual multi-armed bandit literature. However, the assumption that all reward functions are smooth may not be valid in certain practical applications. In such cases, it might be possible to relax Assumption \ref{assumption:smooth} by imposing the smoothness assumption solely on the best few arms while introducing a more delicate condition on the reward gap to replace Assumption~\ref{assumption:margin}. We leave the development of suitable models for such settings to future investigation.
\end{remark}

In addition, we impose a regularity condition on the marginal distribution $Q_X$. It ensures that the support of $Q_{X}$ is regular and that the density is bounded away from zero and infinity on the support.

\begin{assumption}[Bounded density]\label{assumption:bounded-density} There exist constants $\overline{q}>\underline{q}>0$ such that $\underline{q} r^d \leq Q_{X}(B(x,r)) \leq \overline{q} r^d$ for any $x\in \mathsf{supp}(Q_{X})$ and $r\in(0,1]$.
\end{assumption}

With these conditions pertaining to the target bandit $Q$ in place, let us turn to the assumptions that enable reliable transfer learning.
As previously discussed in Section~\ref{sec:intro}, we focus on the covariate shift setting and deploy the concept of the transfer exponent. This notion was originally introduced in \citet{kpotufe2021marginal}, and numerous variants have emerged in the transfer learning literature \citep{hanneke2019value,cai2021transfer,suk2021self,pathak2022new}. 

\begin{definition}[Transfer exponent]\label{def:transfer}Define the transfer exponent $\gamma \in \mathbb{R}_{+} \cup \{0,\infty\}$ of $P_{X}$ with respect to $Q_{X}$ to be the smallest constant such that
\begin{align}
\label{eq:tran-exp}
P_{X}(B(x, r)) \geq c_{\gamma}r^{\gamma} Q_{X}(B(x, r)),\quad\forall x\in\mathsf{supp}(Q_{X}), r \in (0, 1],
\end{align}
for some constant $0 < c_{\gamma} \leq 1$.
\end{definition}

Note that for an arbitrary probability distribution pair $(P, Q)$, condition (\ref{eq:tran-exp}) always holds with $\gamma = \infty$. Also, given that the radius $r$ is always upper bounded by one in $[0,1]^{d}$, the probability mass $P_{X}(B(x,r))$ increases as $\gamma$ approaches to $0$. Intuitively, this implies that the source data cover a larger subset of the covariate regime of interest, allowing more effective information to be transferred from the source distribution $P$ to the target distribution $Q$. 

We now give an example for Definition~\ref{def:transfer}. Let $Q_{X}$ be the uniform distribution over $[0,1]$. Suppose the density function $p_{X}(x)$ of $P_{X}$ takes the form $p_{X}(x) = C x^{\gamma}$ for some normalization constant $C>0$. Then it is easy to verify that the transfer exponent of $P_{X}$ with respect to $Q_{X}$ equals $\gamma$. We refer to \citet{kpotufe2021marginal} for a more in-depth discussion of this transfer exponent.

In addition, the covariate-arm pairs $\big\{\big(X^{P}_{i}, \pi_{i}^{P}\big)\big\}_{i=1}^{n_P}$ in the pre-collected source dataset are assumed to be generated i.i.d.~according to $X^{P}_{i} \sim P_{X}$ and $\pi_{i}^{P}(X^{P}_{i}) \sim \mu(\cdot \, | \, X^{P}_{i})$, where $\{\mu(\cdot \, | \, x) \}_{x\in \mathcal{X}}$ is a collection of probability distributions over the arm set $[K]$. We make a note that this i.i.d.~assumption prevails in the literature on bandits and reinforcement learning where one seeks to exploit offline data \citep{rashidinejad2021bridging}, which is well-motivated by the data randomization procedure in experience replay \citep{mnih2015human}. To gauge the degree of exploration over the arm set in the source dataset, we introduce the exploration coefficient as defined below.
\begin{definition}[exploration coefficient]
\label{def:explore-coef}
Define the exploration coefficient $ \kappa  \in [0, 1]$ of a collection of distributions over the arm set $\{\mu(\cdot \, | \, x) \}_{x\in\mathcal{X}}$ with respect to $Q_X$ as
	\begin{align} 
\label{eq:explore-coef}
	 \kappa  \defn \inf_{\substack{k\in [K],\,x\in \mathsf{supp}(Q_{X})}} K \mu(k \, | \, x).
\end{align}
\end{definition}
Note that $ \kappa  / K$ is the lowest probability of an arm being selected over the support of $Q_X$ in the $P$-data. Intuitively, when the exploration coefficient $ \kappa $ is not vanishingly small, each arm has been extensively tested by the source policy within the regions of interest. This, in turn, provides the decision-maker with greater confidence regarding the reward function associated with each arm, thereby facilitating the decision-making in the target bandit.
We make a note that Definition~\ref{def:explore-coef} exhibits close ties to the positivity assumption in dynamic treatment regimes \citep{shi2020breaking}, as well as the notion of uniformly bounded concentrability coefficient in offline reinforcement learning \citep{munos2007performance,farahmand2010error,chen2019information,xie2021batch,wagenmaker2023leveraging}.

Finally, we assume the number of arms $K$ is constant throughout this paper.

Denote by $\Pi(K,\beta,C_{\beta},\alpha,C_{\alpha},\underline{q},\overline{q},\gamma,c_{\gamma}, \kappa )$ the class of nonparametric contextual $K$-armed bandits that satisfy Assumptions~\ref{assumption:smooth}--\ref{assumption:bounded-density} and Definitions~\ref{def:transfer}--\ref{def:explore-coef}. Here and throughout the paper, we may use the shorthands $\Pi(K,\beta,\alpha,\gamma, \kappa )$ and $\Pi$ if there is no confusion.

\section{Minimax Rate of Convergence}

\label{sec:Main-Results}

In this section, we establish the minimax regret for transfer learning under the covariate shift model and develop a rate-optimal procedure Algorithm~\ref{alg:UCB-TL} to achieve the minimax regret.

\subsection{Algorithm}
\label{subsection:algorithm}

The key to solving nonparametric contextual multi-armed bandit problems lies in accurately estimating the values of the reward functions $\{ f_k \}_{k=1}^{K}$ at each observed point $X_t^Q$.
Inspired by the success of \citet{rigollet2010nonparametric,perchet2013multi} in the classical setting, the high-level idea of Algorithm~\ref{alg:UCB-TL} is fairly straightforward. It dynamically partitions the covariate space $\mathcal{X}$ into a set of hypercubes (bins), and uses local constant estimators for the reward functions in each bin. This reduces the original contextual multi-armed bandit into a collection of (static) multi-armed bandits (without covariates). Subsequently, we can apply a successive elimination algorithm within each bin separately and independently. 
To be more specific, Algorithm~\ref{alg:UCB-TL} generates a sequence of nested partitions $\{ \mathcal{L}_{t} \}_{t \geq 1}$ of the covariate space $\mathcal{X}$ over time, where the partition $\mathcal{L}_{t}$ at time $t$ consists of a set of bins (of potentially different side lengths) in $\mathcal{X}$. 
Here, for any non-negative integer $l \geq 0$, we define a collection of bins $\mathcal{B}_{l}\defn \{B_{\mathsf k}\}_{\mathsf{k} \in [2^l]^d}$ where
\begin{align}
    B_{\mathsf{k}}\defn\{x\in\mathcal{X}:(k_{i}-1)2^{-l}\leq x_{i}\leq k_{i}2^{-l},\,k_{i}\in[2^{l}],\,i\in[d]\},\quad\forall\mathsf{k}=(k_{1},\dots,k_{d}).\label{def:bin}
\end{align}
Throughout the paper, we use $|B|$ to denote the side length of any bin $B$, i.e.~$|B|=2^{-l}$ for any $B\in\mathcal{B}_{l}$. 
As an important observation, for any bin $B\in\mathcal{L}_{t}$ with $Q_{X}(B)>0$, if we restrict our focus to samples of which the covariates fall in bin $B$, it is not hard to see that the corresponding observed rewards $(Y^{Q,(k)}_{s}(B))_{s\geq 1}$ generated by arm $k$ are i.i.d.~random variables with expectation equal to the conditional expectation of the reward of arm $k$ over bin $B$, namely
\begin{align}
\label{eq:cond-mean-reward-func}
    \bar{f}_{k}^{Q}(B)\defn \mathbb{E}\big[f_{k}(X_{t}^{Q}) \,|\,X_{t}^{Q} \in B\big] =  \frac{1}{Q_{X}(B)} \int_{B}f_{k}(x) \, \mathrm{d}Q_{X}(x). 
\end{align}
As a consequence, at each time $t$, given the covariate $X_t^Q$, we first find the bin $B$ in the current partition $\mathcal{L}_t$ that contains $X_t^Q$. We then invoke Procedure~\ref{alg:EA-TL}---a transfer learning procedure tailored to multi-armed bandits that yields a policy $\{\widetilde{\pi}_{s}(B)\}_{s\geq1}$ for bin $B$---to determine $\pi_{t}$, i.e.~the arm to pull at time $t$. 

In order to present the policy $\pi = \{\pi_{t}\}_{t \geq 1}$ generated by Algorithm~\ref{alg:UCB-TL}, we introduce several notations. 
First, for any $x\in\mathcal{X}$ and $t \geq 1$, let $B_{t}(x)\in\mathcal{L}_{t}$ denote
the bin in the partition $\mathcal{L}_{t}$ at time $t$ such that $x\in B_{t}(x)$. If there are multiple bins, we choose $B_{t}(x)$ to be the one whose center is closest to the origin. 
Next, for any bin $B$ and time $t \geq 1$, denote by $N_{t}(B)$ the number of times the covariate fell into $B$ prior to time $t$, i.e.~$N_{t}(B)\defn\sum_{1\leq s\leq t}\mathds{1}\{X_{s}^Q\in B\}$.
With these definitions in place, the policy $\pi = \{\pi_{t}\}_{t \geq 1}$ yielded by Algorithm~\ref{alg:UCB-TL} can be described by $\pi_{t} = \widetilde{\pi}_{N_{t}(B)}(B)$ with $B = B_{t}(X_t^Q)$ for any time $t \geq 1$.

Before delving into the details of Algorithm~\ref{alg:UCB-TL}, we pause to introduce some additional notations. As Algorithm~\ref{alg:UCB-TL} maintains an adaptive partition of the covariate space $\mathcal{X}$ over time, it is convenient to describe the partition using a tree.
To this end, denote by $\mathcal{T}^{(l)}$ the perfect tree with root node $\mathcal{X}$ and depth $l \geq 0$, where there are $2^{i d}$ nodes in each depth $0 \leq i < l$ and each node $B$ represents a bin in set $\mathcal{B}_{i}$. 
The set of children of any bin $B \in \mathcal{B}_{i}$ with $i \geq 0$ is defined as $\mathsf{child}(B) \defn \{ B' \in \mathcal{B}_{i + 1} : B' \subset B \}$. Then at each time~$t$, the partition $\mathcal{L}_t$ induced by Algorithm~\ref{alg:UCB-TL} can be described as a set of leaf nodes of a subtree of $\mathcal{T}^{(l)}$ for some $l>0$.
% As we shall see momentarily, at each time~$t$, the partition $\mathcal{L}_t$ induced by Algorithm~\ref{alg:UCB-TL} can be described as a set of leaf nodes of some subtree $\mathcal{T}$ of $\mathcal{T}_{l^{\star}}$ for some integer $l^{\star} > 0$ to be specified later. 
Throughout the paper, the terms bin and node are used interchangeably.

Next, given a subset $\mathcal{D}$ of the source dataset $\mathcal{D}^{P}$, for any bin $B$ and arm $k$, let $n_{k}^{P}(B;\mathcal{D})$ denote the number of samples in dataset $\mathcal{D}$ such that the covariate falls in bin $B$ and arm $k$ is pulled, i.e.
\begin{align}
    \label{eq:n-k-B-Pdata}
    n_{k}^{P}(B;\mathcal{D}) & \defn\sum_{(X_{i},\pi_{i},Y_{i})\in\mathcal{D}}\mathds{1}\{X_{i}\in B,\,\pi_{i}=k\}.
\end{align}
Let $\overline{Y}_{k}^{P}(B;\mathcal{D})$  denote the empirical mean of the reward of arm $k$ over bin $B$ in dataset~$\mathcal{D}$, namely
\begin{align}
\label{eq:Y-bar-k-B-Pdata}
\overline{Y}_{k}^{P}(B;\mathcal{D})\defn
\begin{cases}
    \frac{1}{n_{k}^{P}(B;\mathcal{D})}\sum\limits_{(X_{i},\pi_{i},Y_{i})\in\mathcal{D}}Y \mathds{1}\{X_{i}\in B,\,\pi_{i}=k\}, & \text{if }  n_{k}^{P}(B;\mathcal{D}) \neq 0, \\
    0, & \text{otherwise.}
\end{cases}
\end{align}
 In addition, for any non-negative integer $\tau \geq 0$, bin $B$ and arm $k$, let us define
\begin{align}
U_{k}(\tau,B;\mathcal{D})\coloneqq
\begin{cases}
2\sqrt{\frac{2}{\tau+n_{k}^{P}(B;\mathcal{D})}\log^{+}(\frac{n_{Q}|B|^{d}}{\tau})} \vee 2C_{\beta}|B|^{\beta}, & \text{if } \tau > 0, \\
2\sqrt{\frac{2}{n_{k}^{P}(B;\mathcal{D})}\log^{+}(n_{Q}|B|^{d+2\beta}\vee\kappa n_{P}|B|^{d+2\beta+\gamma})} \vee 2C_{\beta}|B|^{\beta}, & \text{if } \tau = 0,
\end{cases} \label{eq:UCB}
\end{align}
where we recall the notation $\log^{+}(x)\defn\log(x)\vee1$ and use the convention $1/0 = \infty$. The $U_{k}(\tau,B;\mathcal{D})$ can be essentially viewed as a confidence bound that quantifies the uncertainty of the reward function estimator used in Algorithm~\ref{alg:UCB-TL}.

With these notations in place, the transfer learning algorithm for nonparametric contextual multi-armed bandits is summarized in Algorithm~\ref{alg:UCB-TL}. Let us discuss its details, along with some intuition. As briefly mentioned earlier, Algorithm~\ref{alg:UCB-TL} aims to segment the covariate space based on the local margins of the reward functions. Smaller bins are employed in areas where the gaps between the reward functions of different arms are small, whereas coarser partitioning is used in regions where arms are easily distinguishable. Once this partition is established, each bin is treated as an index for a sequence of static multi-armed bandit problems, and Procedure~\ref{alg:EA-TL} is executed in each bin with parameters specific to that bin.

\begin{algorithm}[t]
\caption{Transfer learning algorithm for contextual multi-armed bandits}
\label{alg:UCB-TL} \begin{algorithmic}[1]
\State{\textbf{Input:} arm set $\mathcal{I}$, horizon length
$n_{Q}$, smoothness parameters $\beta, C_{\beta}$, transfer parameters $\gamma$, exploration coefficient $\kappa$, $P$-data $\mathcal{D}^{P}$.}
\State{Initialize $\mathcal{L}_{1}\gets\{\mathcal{X}\}$, $\mathcal{I}(\mathcal{X})\gets\mathcal{I}$.} 
    \Comment{initialize partition and arm set}
\State{Initialize the policy $\widetilde{\pi}(\mathcal{X})$ by Procedure~\ref{alg:EA-TL}$\big(\mathcal{X},\mathcal{I}(\mathcal{X}),U,\mathcal{D}^{P} \big)$.}
\State{Initialize $N(\mathcal{X})\gets0$.}
    \Comment{initialize time for policy $\widetilde{\pi}(\mathcal{X})$}
\State{Initialize $\tau_{k}(\mathcal{X}) \gets 0$, and $\tau_{k}^{\star}(\mathcal{X};\mathcal{D}^{P}) $ as in \eqref{eq:tau-star}, $\forall k\in \mathcal{I}(\mathcal{X})$.} 
    \Comment{initialize rounds and set round upper bounds}
\For{$t=1,\dots,n_{Q}$}
\State{Draw a sample $X_{t}^{Q} \sim Q_{X}$.}
\State{Find the bin $B\in\mathcal{L}_{t}$ such that $X_{t}^{Q}\in B$.} 
\While{$|\mathcal{I}(B)|>1$
% , $|B|>2^{-l^{\star}}$
and $\tau_{k}(B)\geq\tau_{k}^{\star}(B;\mathcal{D}^{P}), \forall k\in\mathcal{I}(B)$}
    \Comment{keep partitioning $B$ until reaching suitable scale}
\If{$\tau_{k}^{\star}(B;\mathcal{D}^{P}) = 0, \forall k\in\mathcal{I}(B)$}
    \Comment{no exploration needed in $B$: discard suboptimal arms}
\State{Set $\underline{Y}^{\star}(B;\mathcal{D}^{P})\gets\max\limits_{k\in\mathcal{I}(B)}\big\{\overline{Y}_{k}^P(B;\mathcal{D}^{P})-  U_{k}(0,B;\mathcal{D}^{P}) \big\}$.}
    \Comment{set largest reward lower bound}
\State{Set $\mathcal{I}(B) \gets \big\{ k \in \mathcal{I}(B) : \overline{Y}_k^P(B;\mathcal{D}^{P})+U_{k}(0,B;\mathcal{D}^{P}) \geq \underline{Y}^{\star}(B;\mathcal{D}^{P}) \big\}$.}
    \Comment{update arm set}
\EndIf
\For{$B'\in\mathsf{child}(B)$}
\State{Set $\mathcal{I}(B')\gets\mathcal{I}(B)$.}
    \Comment{assign remaining arms in $B$ as initial arms in its children}
\State{Initialize the policy $\widetilde{\pi}(B')$ by Procedure~\ref{alg:EA-TL} $ \big(B',\mathcal{I}(B'),U,\mathcal{D}^{P} \big)$.}
\State{Set $N(B')\gets0$.}
    \Comment{initialize time for policy $\widetilde{\pi}(B')$}
\State{Set $\tau_{k}(B')\gets0$ and $\tau_{k}^{\star}(B';\mathcal{D}^{P})$ as in \eqref{eq:tau-star}, $\forall k\in\mathcal{I}(B')$.}
    \Comment{initialize rounds and set round upper bounds}
\EndFor
\State{Set $\mathcal{L}_{t}\gets (\mathcal{L}_{t}\setminus B)\cup\mathsf{child}(B)$.}
    \Comment{replace $B$ with its children in partition}
\State{Find the bin $B\in\mathcal{L}_{t}$ such that $ X_{t}^{Q}\in B$.}
\EndWhile
\State{Set $N(B)\gets N(B)+1$.}
    \Comment{update times $X_{t}^{Q} \in B$}
\State{Set $\pi_{t}\gets\widetilde{\pi}_{N(B)}(B)$.}
    \Comment{choose arm by policy $\widetilde{\pi}(B)$}
\State{Set $\mathcal{I}(B)\gets\widetilde{\mathcal{I}}_{N(B)} (B)$.}
    \Comment{update arm set by policy $\widetilde{\pi}(B)$}
\State{Set $\tau_{k}(B) \gets \widetilde{\tau}_{N(B),k}(B), \forall k\in\mathcal{I}(B)$.}
    \Comment{update numbers of rounds by policy $\widetilde{\pi}(B)$}
\EndFor
\State{\textbf{Output:} policy $\{\pi_{t}\}_{t\geq1}$.}
\end{algorithmic}
\end{algorithm}

\addtocounter{algorithm}{-1}
\floatname{algorithm}{Procedure}
\begin{algorithm}[t]
\caption{Successive elimination procedure for a static bandit with source data}
\label{alg:EA-TL} 
\begin{algorithmic}[1]
\State{\textbf{Input:} bin $B$, arm set $\mathcal{I}$, confidence bound function $U$, source data $\mathcal{D}$.}
\State{Set $n_{k}^{P}(B;\mathcal{D}),\overline{Y}_{k}^{P}(B;\mathcal{D}),\tau_{k}^{\star}(B;\mathcal{D})$ as in (\ref{eq:n-k-B-Pdata}), (\ref{eq:Y-bar-k-B-Pdata}), (\ref{eq:tau-star}), respectively, $\forall k\in\mathcal{I}$.}
% \State{Initialize
% $t\gets0$ For each $k\in\mathcal{I}$ $(\tau_{0,k})_{k\in \mathcal{I}}\gets(0, \cdots, 0)$, and $(\overline{Y}_{k})_{k\in\mathcal{I}} \gets (\overline{Y}_{k}^{P} )_{k\in\mathcal{I}}$}
\State{Set $n_{k}^{P} \gets n_{k}^{P}(B;\mathcal{D})$, $\overline{Y}_{k} \gets \overline{Y}_{k}^{P}(B;\mathcal{D})$, and $\tau_{k}^{\star} \gets \tau_{k}^{\star}(B;\mathcal{D}), \forall k\in\mathcal{I}$.}
\State{Initialize $t\gets0$.}
\State{Initialize $\tau_{k} \gets 0, \forall k\in\mathcal{I}$.} 
    \Comment{initialize pull counts}
\State{Initialize $\underline{Y}^{\star}\gets\max\limits_{k\in\mathcal{I}}\big\{\overline{Y}_{k}-U_{k}(0,B;\mathcal{D})\big\}$.}
    \Comment{initialize largest reward lower bound}
\Loop
\If{$\tau_{k}\geq\tau_{k}^{\star}, \forall k\in\mathcal{I}$} 
	\Comment{condition to stop exploration}
\State{Set $t \gets t+1$.}
\State{Select arm $\widetilde{\pi}_{t}\gets\argmax\limits_{k\in\mathcal{I}}\overline{Y}_{k}$ (with ties broken arbitrarily) and receive reward $Y^{Q,({\widetilde{\pi}_{t}})}$.}
\State{Set $\tau_{\widetilde{\pi}_{t}}\gets\tau_{\widetilde{\pi}_{t}} + 1$.}
	\Comment{update pull count}
\State{Set $\tau_{t,k}\gets \tau_{k}, \forall k\in\mathcal{I}$, and $\mathcal{I}_{t}\gets\mathcal{I}$.}
	\Comment{record pull count and active arm set}
\State{Set $\overline{Y}_{\widetilde{\pi}_{t}}\gets\frac{1}{n_{\widetilde{\pi}_{t}}^{P}+\tau_{\widetilde{\pi}_{t}}}\big(Y^{Q,({\widetilde{\pi}_{t}})}+(n_{\widetilde{\pi}_{t}}^{P}+\tau_{\widetilde{\pi}_{t}}-1)\overline{Y}_{\widetilde{\pi}_{t}}\big)$.}
    \Comment{update estimated reward}
\Else
\For{$k\in\mathcal{I} \text{ such that } \tau_{k}<\tau_{k}^{\star}$}
% \If{$\tau_{t,k}<\tau_{k}^{\star}$}
    \Comment{only explore arms s.t.~pull counts less than upper bounds}
\If{$\overline{Y}_{k}+  U_{k}(\tau_{k},B;\mathcal{D}) \geq\underline{Y}^{\star}$}
    \Comment{eliminate arm s.t.~reward upper bound is smaller than largest reward lower bound}
\State{Set $t \gets t+1$.} 
\State{Select arm $\widetilde{\pi}_{t}\gets k$ and receive reward $Y^{Q,({\widetilde{\pi}_{t}})}$.}
\State{Set $\tau_{\widetilde{\pi}_{t}}\gets\tau_{\widetilde{\pi}_{t}} + 1$.}
	\Comment{update pull count}
\State{Set $\tau_{t,k}\gets \tau_{k}, \forall k\in\mathcal{I}$, and $\mathcal{I}_{t}\gets\mathcal{I}$.}
	\Comment{record pull count and active arm set}
\State{Set $\overline{Y}_{\widetilde{\pi}_{t}}\gets\frac{1}{n_{\widetilde{\pi}_{t}}^{P}+\tau_{\widetilde{\pi}_{t}}}\big(Y^{Q,({\widetilde{\pi}_{t}})}+(n_{\widetilde{\pi}_{t}}^{P}+\tau_{\widetilde{\pi}_{t}}-1)\overline{Y}_{\widetilde{\pi}_{t}}\big)$.}
    \Comment{update estimated reward}
\State{Set $\underline{Y}^{\star}\gets\max_{k\in\mathcal{I}}\big\{\overline{Y}_{k}-U_{k}(\tau_{k},B;\mathcal{D})\big\}$.}
    \Comment{update largest reward lower bound}
% \State{Set $\tau_{t}\gets\tau_{t-1}$, $\tau_{t,\widetilde{\pi}_{t}}\gets\tau$, $\mathcal{I}_{t}\gets\mathcal{I}$.}
\Else \State{Eliminate arm $k$ from active arm set: $\mathcal{I} \gets\mathcal{I}\setminus\{k\}$.}
\EndIf
% \EndIf
\EndFor
\EndIf
\EndLoop
\State{\textbf{Output:} policy $\{\widetilde{\pi}_{t}\}_{t\geq1}$, arm pull counts $\{(\tau_{t,k})_{k\in\mathcal{I}_t}\}_{t\geq1}$, sets of active arms $\{\mathcal{I}_{t}\}_{t\geq1}$.}
\end{algorithmic}
\end{algorithm}

\begin{figure}[t]
\centering
\includegraphics[width=0.75\textwidth]{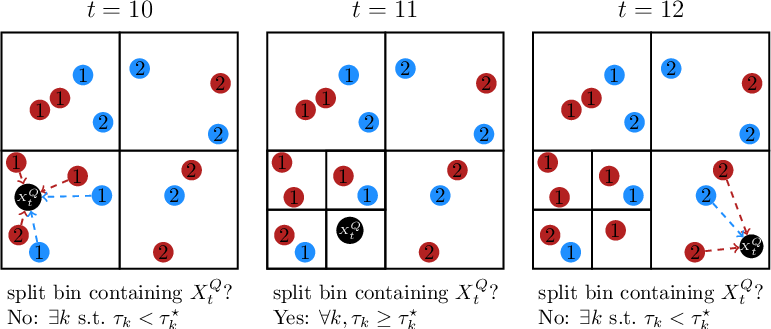}  
\caption{An illustration of Algorithm~\ref{alg:UCB-TL} for $d=2$ and $K=2$. The target samples (resp.~source samples) are represented by the red (resp.~blue) points. The coordinates of each point correspond to the covariate $X$, and the number in the point stands for the arm that is pulled. In each time step $t$, Algorithm~\ref{alg:UCB-TL} first assesses if the bin containing $X_t^Q$ requires splitting. It then utilizes the samples located in the same bin as $X_t^Q$ to execute a static MAB procedure to select an arm. For example, at time $t = 11$, one has (say) $\tau_1^\star= 3$, $\tau_2^\star = 1$, and both arms are active. In this case, we need to split the lower left bin and run Procedure~\ref{alg:EA-TL} in the bin containing $X_t^Q$ to choose an arm.
\label{fig:alg}}
\end{figure}

In order to achieve such an adaptive partition, for each bin $B$ and arm $k$, we assign a non-negative upper bound on the number of pulls, given by
\begin{align}
\tau^\star_k(B;\mathcal{D}) \defn \min_{\tau \in \{0\}\cup\mathbb{N}}\big\{ \tau : U_{k}(\tau,B;\mathcal{D}) 
% \leq 2c_{2}|B|^{\beta}\label{eq:tau-star}
\leq 2C_{\beta}|B|^{\beta}\big\}, \label{eq:tau-star}
\end{align}
where we recall the definition of $U_{k}(\tau,B;\mathcal{D})$ in \eqref{eq:UCB}. Note that the confidence bound $U_{k}(\tau, B;\mathcal{D})$ is composed of two terms. The first term represents the standard deviation of the reward function estimator owing to finite samples, while the second component stands for the bias term, as we attempt to approximate the reward function using its conditional expectation over bin $B$.
Roughly speaking, the value of $\tau^\star_k(B;\mathcal{D})$ is chosen to ensure that, after arm $k$ has been pulled $\tau^\star_k(B;\mathcal{D})$ times in bin $B$, the standard deviation and bias of its reward function estimator is balanced. In particular, if the conditional mean reward of arm $k$ over bin $B$ is low, Procedure~\ref{alg:EA-TL} executed in bin $B$ can identify and eliminate it by the end of $\tau_{k}^{\star}(B;\mathcal{D})$ rounds with high probability. Combined with the smoothness assumption, this procedure guarantees that the eliminated arms are uniformly suboptimal over bin $B$ and that none of the remaining arms dominates the others. 
Therefore, if multiple arms remain active in bin $B$ after each arm $k$ has been pulled $\tau_{k}^{\star}(B;\mathcal{D})$ times, one knows that the reward functions of the remaining arms are locally close to each other, and hence need more refined estimation. Consequently, we split the node $B$ by replacing $B$ with its children $\mathsf{child}(B)$ in the partition tree, and the set of the active arms in node $B$ is passed on to each $B'\in \mathsf{child}(B)$. 
An illustration of Algorithm~\ref{alg:UCB-TL} can be found in Figure~\ref{fig:alg}.

Next, let us move on to take a closer look at Procedure~\ref{alg:EA-TL}. Since it is designed for each bin, in addition to the bin index~$B$, set of arms~$\mathcal{I}$ and confidence bound function~$U$, Procedure~\ref{alg:EA-TL} also requires the information of the source samples that fall in the bin. Specifically, Procedure~\ref{alg:EA-TL} needs the sample size $n_{k}^{P}$, empirical mean of the reward $\overline{Y}^{P}_{k}$, and upper bound on the play rounds $\tau_{k}^{\star}$, for each arm $k \in \mathcal{I}$. 

Before any arm reaches its play round limit, Procedure~\ref{alg:EA-TL} runs similarly to a standard successive elimination algorithm (see e.g.~\citet{auer2010ucb,perchet2013multi}). It operates in rounds and maintains a set of active arms that are potentially optimal. In each round, each arm in the active arm set is pulled once. Given access to the source dataset, the observed rewards from the $Q$-bandit and $P$-bandit are combined to calculate the empirical average of the reward of each active arm. It then seeks to eliminate suboptimal arms from the set of active arms by comparing their upper and lower confidence bounds of the sample mean rewards.

However, once arm $k$ has been pulled for $\tau_k^\star$ times, Procedure~\ref{alg:EA-TL} stops selecting it temporarily. In fact, $\tau^\star_k$ can be viewed as the maximum horizon length of the exploration phase for arm $k$. The rationale is simple: given the additional information from the source data, one should be able to gain more certainty about whether arm $k$ is optimal compared to the standard case. Consequently, we can reduce the cumulative regret by exiting from the exploration stage earlier. 
Once each active arm reaching its play round limit, Procedure~\ref{alg:EA-TL} advances to the second phase, where we select only the arm with the highest empirical average reward. In this case, a previously suspended arm might be pulled again if it happens to be the only active arm or has the highest empirical mean award.
Therefore, one critical difference between this work and previous results on classical contextual multi-armed bandits (without source data), such as \citet{perchet2013multi}, is that our successive elimination procedure in each bin involves a more complicated early stopping stage, hence requiring much a more sophisticated analysis. In addition, achieving the exact minimax regret demands more carefully designed uncertainty estimates that take account of both the source sample size and transfer learning parameters.

It is noteworthy that Algorithm~\ref{alg:UCB-TL} has three critical steps to integrate the source data. First, we combine the target and source data to estimate the reward function of each arm, and the increased sample size leads to an improvement in estimation accuracy.
Second, note that the upper bound on the play rounds $\tau_{k}^{\star}(B;\mathcal{D})$ in (\ref{eq:tau-star}) incorporates the information from the source data $\mathcal{D}$. It is easy to see that $\tau_{k}^{\star}(B;\mathcal{D})$ is a decreasing function of the source sample size $n_k^P(B;\mathcal{D})$. Therefore, if a suboptimal arm has been pulled sufficiently many times in the $P$-bandit so that one is certain about its suboptimality, we no longer need to test it in the $Q$-bandit. This implies a shorter exploration phase and thus reduces the regret. 
Finally, closely related to the second point, the source data allows us to build a deeper partition tree.
We can then discretize the covariate space more finely, thus facilitating the identification of suboptimal arms and incurring a smaller regret.

We would like to remark that the covariate shift of the source data is characterized by the transfer exponent~$\gamma$, which plays a crucial role in Algorithm~\ref{alg:UCB-TL}.
For instance, as $\gamma$ decreases, the partition tree becomes deeper, resulting in a more refined partition of the covariate space.
Therefore, when the covariate shift is mild, Algorithm~\ref{alg:UCB-TL} can construct more accurate local estimators for the reward functions. This aids in distinguishing the optimal arm, thereby reducing the cumulative regret.
Additionally, the confidence bound $U_{k}(\tau,B;\mathcal{D})$ in \eqref{eq:UCB} for arm $k$ in bin $B$ depends on the number of source samples $n_{k}^{P}(B;\mathcal{D})$ such that the covariate falls within bin $B$ and arm $k$ is pulled. This number depends on the source marginal distribution $P_{X}$ and, consequently, the transfer exponent $\gamma$.  As $\gamma$ approaches zero, $n_{k}^{P}(B;\mathcal{D})$ tends to increase with high probability, leading to a tighter confidence bound $U_{k}(\tau,B;\mathcal{D})$. According to the elimination criterion in Algorithm~\ref{alg:UCB-TL} and Procedure~\ref{alg:EA-TL}, this enhances the accuracy of distinguishing the optimal arm, resulting in a reduction in the cumulative regret.
Moreover, Algorithm~\ref{alg:UCB-TL} guarantees that in each bin $B$, each arm $k$ is played for a maximum of $\tau^\star_k(B;\mathcal{D})$ times. As the upper bound on the number of pulls $\tau^\star_k(B;\mathcal{D})$ in \eqref{eq:tau-star} depends on the confidence bound $U_{k}(\tau,B;\mathcal{D})$, it is also influenced by the transfer exponent $\gamma$. As $\gamma$ goes to zero, the upper bound on the number of pulls $\tau^\star_k(B;\mathcal{D})$ decreases.  Therefore, when the covariate shift is slight, Algorithm~\ref{alg:UCB-TL} selects suboptimal arms less frequently, leading to a decrease in the cumulative regret.

Finally, we note that Algorithm~\ref{alg:UCB-TL} takes the horizon length $n_Q$ as input that may be unknown in practice. Fortunately, this issue can be circumvented by the well-known doubling trick (see \citet{auer1995gambling}).

\subsection{Minimax rate of convergence}

We proceed to discuss the theoretical guarantees of Algorithm~\ref{alg:UCB-TL}.
To begin with, Theorem~\ref{thm:upper-bound} gives an upper bound
on the regret of the $Q$-bandit. The proof is postponed to Appendix~\ref{sec:Proof-of-Upper-Bound} in the supplementary material \citep{cai2022transfer-supp}.

\begin{theorem}[Upper bound]\label{thm:upper-bound}
Assume that $\alpha \beta \leq d$. Then the expected regret of the policy $\pi$ given by Algorithm~\ref{alg:UCB-TL} satisfies 
\begin{align}
\sup_{\Pi(K,\beta,\alpha,\gamma, \kappa)} \mathbb{E}[R_{n_Q}(\pi)]\leq Cn_{Q}\Big(n_{Q} + (\kappa n_{P})^{\frac{d+2\beta}{d+2\beta+\gamma}}\Big)^{-\frac{\beta(1+\alpha)}{d+2\beta}}\label{eq:upper-bound-oracle}
\end{align}
where $C>0$ is a constant independent of $n_Q$ and $n_P$.
\end{theorem}

In addition, Theorem~\ref{thm:lower-bound} below shows that the regret of Algorithm~\ref{alg:UCB-TL} matches the minimax lower bound, thereby justifying Algorithm~\ref{alg:UCB-TL} is rate-optimal.
The proof can be found in Appendix~\ref{sec:Proof-of-Lower-Bound} in the supplementary material~\citep{cai2022transfer-supp}.

\begin{theorem}[Lower bound]\label{thm:lower-bound}Assume that $\alpha \beta \leq d$. Then one has 
\begin{align}
\inf_{\pi}\sup_{\Pi(K,\beta,\alpha,\gamma, \kappa )}\mathbb{E}[R_{n_Q}(\pi)]\geq c n_{Q}\Big(n_{Q} + (\kappa n_{P})^{\frac{d+2\beta}{d+2\beta+\gamma}}\Big)^{-\frac{\beta(1+\alpha)}{d+2\beta}}\label{eq:lower-bound-oracle}
\end{align}
where $c>0$ is a constant independent of $n_Q$ and $n_P$.
\end{theorem}
Here, the infimum is taken over the class of admissible policies obeying the selected arm~$\pi_t$ at time $t$ depends only on observations prior to time $t$, i.e.~$\{(X^{Q}_s, \pi_s, Y^{Q,(\pi_s)})\}_{s<t}\cup \{X^{Q}_{t}\}$.

We now discuss several important implications.
\begin{itemize}
\item Theorems~\ref{thm:upper-bound} and \ref{thm:lower-bound} together establish the minimax regret for transfer learning under the covariate shift model when the number of arms $K\asymp 1$:
\begin{align}
\inf_{\pi}\sup_{\Pi(K,\beta,\alpha,\gamma, \kappa )}\mathbb{E}[R_{n_Q}(\pi)]\asymp n_{Q}\Big(n_{Q} + (\kappa n_{P})^{\frac{d+2\beta}{d+2\beta+\gamma}}\Big)^{-\frac{\beta(1+\alpha)}{d+2\beta}}.\label{eq:minimax-rate}
\end{align}

\item In the classical setting where one has no access to the source $P$-data,
the minimax regret is established in \citet{perchet2013multi}, given
by
\begin{align}
\inf_{\pi}\sup_{\Pi(K,\beta,\alpha)}\mathbb{E}[R_{n_Q}(\pi)]\asymp n_{Q}^{1-\frac{\beta(1+\alpha)}{d+2\beta}}.\label{eq:minimax-rate-conventional}
\end{align}
We can see that (\ref{eq:minimax-rate-conventional}) is a special
case of (\ref{eq:minimax-rate}) by setting $n_{P}=0$.
\item Comparing the minimax regret (\ref{eq:minimax-rate}) in the transfer learning setting with the minimax regret (\ref{eq:minimax-rate-conventional}) in the standard setting, it becomes evident that the incorporation of source data leads to a faster convergence rate of the regret. 
The reduced regret quantifies the information allowed to be transferred from the source data to the target bandit, with a precise characterization of the dependence on the smoothness $\beta$, dimension $d$, transfer exponent $\gamma$, and exploration coefficient $\kappa$.
\item Due to the difference between the source and target distributions, it is reasonable to anticipate that the value of the source data differs from that of the target data. This intuition is elucidated in (\ref{eq:minimax-rate}) where we can interpret $(\kappa n_{P})^{\frac{d+2\beta}{d+2\beta+\gamma}}$ as the effective sample size provided by the source data. Moreover, given that $\kappa \in [0,1]$ and $\frac{d+2\beta}{d+2\beta+\gamma}\leq1$ for $\gamma\geq0$, the minimax regret (\ref{eq:minimax-rate}) further suggests that the samples from the source dataset are always inferior unless $P = Q$ in the context of nonparametric contextual multi-armed bandits.
\item We proceed to examine the roles of the parameters introduced for transfer learning. As we intuitively discussed in Section~\ref{sec:Problem-formulation}, the challenge of transfer learning becomes more formidable as the transfer exponent $\gamma$ increases. This observation is validated theoretically in (\ref{eq:minimax-rate}), demonstrating that a smaller value of $\gamma$ results in a faster convergence rate of the regret.
\item 
Shifting our focus to the exploration coefficient $\kappa$, we can see that the dependence of the minimax rate (\ref{eq:minimax-rate}) on $\kappa$ underscores the importance of extensive exploration in the source data in order to achieve reliable transfer learning. When $\kappa = 0$, the effective source sample size is zero. To see this, let us consider a three-armed bandit problem where the policy in the source domain exclusively selects arm $3$. In such a case, even if the source sample size $n_P$ were to go to infinity which provides us with precise knowledge of the reward function $f_{3}(x)$ of arm $3$, we would still be confronted with a two-armed bandit problem, with minimax rate of convergence of the regret given by $n_{Q}^{1-\frac{\beta(1+\alpha)}{d+2\beta}}$. This indicates that a substantial reduction in regret cannot be expected unless the source dataset is highly exploratory.
\item We would like to remark that the minimax lower bound (\ref{eq:lower-bound-oracle}) in Theorem~\ref{thm:lower-bound} is the same as the one established for the classification setting in \citet{kpotufe2021marginal}. This means that even if we can observe the rewards generated by all the arms in each round, it remains impossible to design a policy that can improve the regret upper bound (\ref{eq:lower-bound-oracle}) in terms of $n_P$ and $n_Q$. Intuitively, this implies that although we are faced with the challenge of sequential decision-making, the hardness of nonparametric estimation ultimately dictates the complexity of transfer learning for nonparametric contextual multi-armed bandits.

\end{itemize}

% Now we briefly discuss the high-level idea of the proof of Theorem~\ref{thm:upper-bound}. As shall be clear in Section~\ref{sec:Proof-of-Upper-Bound}, we upper bound the regrets incurred on the non-leaf and leaf nodes separately. Thanks to the source dataset, the regret on a non-leaf node is negligible as long as its depth is smaller than some threshold that depends on the source sample size and covariate shift. Also, the regret accumulated on the leaf nodes is controlled by the depth of the partition tree. Combining these two regrets and optimizing over the tree depth leads to the minimax regret. As for the minimax lower bound, we construct a set of contextual multi-arm bandit instances with Bernoulli random rewards such that they are close in distribution but distant in terms of the induced regret. One can then apply the standard tool such as Fano's inequality to establish Theorem~\ref{thm:lower-bound}.

We conclude this section by comparing our work with some intimately connected prior research in the literature on transfer learning. As mentioned earlier, the transfer exponent used in the current paper was originally proposed in \citet{kpotufe2021marginal}. Along with its variants (see e.g.~\citet{cai2021transfer,pathak2022new}), they have been broadly deployed in transfer learning for various supervised learning problems, where the primary focus is on leveraging the source dataset to develop optimal estimators for the regression functions of interest. In contrast, the sequential decision-making nature of bandit problems introduces new algorithmic and technical challenges in data integration compared to these prior works. For example, in order to address the trade-off between exploration and exploitation, it is essential not only to construct suitable reward function estimators but also to utilize the source data intelligently to develop confidence intervals for their uncertainties.
Moreover, the samples for each arm are collected adaptively in the bandit setting. This results in more complicated distributions of samples and reward function estimators, necessitating more careful statistical analysis.

\section{Adaptivity}
\label{sec:Adaptivity}

The primary drawback of Algorithm~\ref{alg:UCB-TL} is its reliance on prior knowledge of the smoothness parameter $\beta$ and transfer parameters $\gamma,\kappa$, which are often unknown in practical applications. Therefore, it is natural to ask if one can develop a data-driven algorithm that achieves the minimax optimal rate of convergence while adapting to a wide range of parameter spaces $\Pi(K,\beta,\alpha,\gamma,\kappa)$. However, it is widely acknowledged in the bandit literature that one cannot hope to develop a smoothness-adaptive algorithm that attains the minimax regret simultaneously over different classes of multi-armed bandits with varying smoothness \citep{locatelli2018adaptivity,gur2022smoothness,cai2022stochastic}. Additional structural assumptions on the reward functions are needed to achieve smoothness adaptivity. 

Given the general impossibility of adaptation, we focus on a setting where the reward functions satisfy the self-similarity condition. This condition has been used for adaptive confidence intervals in the nonparametric regression literature \citep{picard2000adaptive,gine2010confidence} and the adaptive multi-armed bandit literature \citep{gur2022smoothness,cai2022stochastic}. In the remainder of this section, we first introduce the self-similarity condition in Section~\ref{subsec:The-self-similarity-assumption} and subsequently present a data-driven algorithm with theoretical guarantees in Section \ref{subsec:Adaptive-Algorithm}.
As a side note, we believe it is also possible to develop adaptive strategies for the bandits with shape-constrained reward functions (e.g.~concavity), and we defer the discussion to Section~\ref{sec:Discussion}.

\subsection{The self-similarity condition}
\label{subsec:The-self-similarity-assumption}
 
For any function $f(\cdot)$ on $\mathcal{X}$, bin $B$ in $\mathcal{X}$, and probability distribution $\lambda$ over $\mathcal{X}$, let $\Gamma_{B}f(\cdot ; \lambda)$ be the $L_{2}(\lambda)$-projection of $f$ onto the class of piecewise-constant functions over $B$, namely
\begin{align}
\label{eq:def-poly-proj}
\Gamma_{B}f(x ; \lambda) \defn \frac{1}{\lambda(B)} \int_{B} f(u)  \, \mathsf{d}\lambda(u)
\end{align}
if $\lambda(B) > 0$, and $\Gamma_{B}f(x ; \lambda) \defn 0$ otherwise.

Recall that $\mathcal{B}_{l}$ denotes the partition of $\mathcal{X}$ that consists of bins with side length $2^{-l}$. We now present the self-similarity condition as follows.

\begin{definition}[Self-similarity]\label{def:self-similar} Let $f:\mathcal{X} \rightarrow \mathbb{R}$ be a H\"{o}lder continuous function in $\mathcal{H}(\beta, C_\beta)$ with $0<\beta \leq 1$. For any probability distribution $\lambda$ and constants $l_{0}\geq 0$, $b >0$, we say $f$ is self-similar under $\lambda$ with parameters $l_{0}$ and $b$, if the following holds for any integer $l\geq l_{0}$:
\begin{align*}
\sup_{B\in\mathcal{B}_{l}}\sup_{x\in B}\big|\Gamma_{B}f(x ; \lambda)-f(x)\big|\geq b2^{-\beta l}.
\end{align*}
\end{definition}

In a nutshell, the self-similarity condition imposes a global lower bound on the approximation error of a function using piecewise-constant functions. Therefore, it can be viewed as a complement to the H\"older smoothness condition, which implies an upper bound on the error. We note that the difficulty of smoothness adaptation arises from the possibility of functions being highly irregular on small scales, a scenario precluded by the self-similarity condition.
Interested readers are referred to \citet{bull2012honest,nickl2016sharp} for more in-depth discussions on the self-similarity conditions. Below, we present an example of the self-similar functions.

\begin{example}
\label{example:self-similar-func}
Fix some constants $a,c \in \mathbb{R}$. The function $f(x)=ax^{\beta}+c$ is self-similar under the uniform distribution over $[0,1]$ with parameters $l_0 = 0$ and $b = \frac{a}{\beta + 1}$.
\end{example}

To verify this, let us fix an arbitrary $l \geq 0$ and denote $B_{0} = [0,2^{-l}]$. It is straightforward to compute
\begin{align*}
&\sup_{B\in\mathcal{B}_{l}}\sup_{x\in B}\big|\Gamma_{B}f(x ; \mathsf{Unif}[0,1])-f(x)\big|  \geq \big|\Gamma_{B_0}f(0;\mathsf{Unif}[0,1])-f(0)\big| \\
& \quad= \bigg|  \int_{B_{0}} f(x) 2^{l} \, \mathrm{d}x - c \bigg|  = a \int_{0}^{2^{-l}} x^{\beta} 2^{l}\, \mathrm{d}x  = \frac{a }{\beta + 1 } 2^{-\beta l}.
\end{align*}
As a consequence, this shows that $f(x)=ax^{\beta}+c$ is self-similar by Definition \ref{def:self-similar}.

We now introduce the self-similarity assumption regarding the reward functions $\{f_{k}\}_{k=1}^{K}$ as follows.
\begin{assumption}[Self-similarity]\label{assumption:self-similar} 
Assume that there exists some $k\in[K]$ such that the reward function $f_{k}$ is self-similar under both $Q_{X}$ and $P_{X\,|\,\pi=k}$ with some parameters $l_{0} \geq 0$ and $b > 0$.
\end{assumption}

We denote by $\Pi(K,\beta,C_{\beta},\alpha,C_{\alpha},\underline{q},\overline{q},\gamma,c_{\gamma}, \kappa,l_{0},b)$ the class of nonparametric contextual $K$-armed bandits that satisfy Assumptions~\ref{assumption:smooth}--\ref{assumption:self-similar} and Definitions~\ref{def:transfer}--\ref{def:explore-coef}.
Whenever clear from the context, the shorthand $\Pi(K,\beta,\alpha,\gamma, \kappa,l_{0},b)$ and $\Pi$ are used.

\subsection{Adaptive algorithm}

\label{subsec:Adaptive-Algorithm}

In this section, we present a two-stage adaptive transfer learning algorithm designed for contextual multi-armed bandits with self-similar reward functions. 
In essence, this adaptive algorithm begins by computing a reasonably precise estimate $\widehat{\beta}$ for the H\"{o}lder smoothness parameter $\beta$ (see Procedure~\ref{alg:smoothness}). It then uses $\widehat{\beta}$ as an input to a variant of Algorithm~\ref{alg:UCB-TL}, 
% (using a new confidenc function $\widehat{U}_{k}(\tau,B;\mathcal{D}^{P}_{\dm})$)
which guarantees a near-optimal statistical performance if given an accurate smoothness estimate. A comprehensive description of this algorithm is summarized in Algorithm \ref{alg:UCB-TL-adaptive}.

\floatname{algorithm}{Algorithm}
\begin{algorithm}[t]
\caption{Adaptive transfer learning algorithm for contextual multi-armed bandits}
\label{alg:UCB-TL-adaptive} 
\begin{algorithmic}[1]
\State{\textbf{Input:} arm set $\mathcal{I}$, horizon length $n_{Q}$, lower and upper bounds on smoothness $\underline{\beta}$ and $\overline{\beta}$, upper bound on Lipschitz constant $\overline{C}_\beta$, upper bound on transfer exponent $\overline{\gamma}$, $P$-data $\mathcal{D}^{P}$.}

\State{Run Procedure~\ref{alg:smoothness} ($K, n_{Q}, \underline{\beta}, \overline{\beta}, \overline{\gamma}, \mathcal{D}^{P}$) to get the smoothness estimate $\widehat{\beta}$, time steps $s_{P}$ and $s_{Q}$, and policy $\{\pi_{t}^{\adp}\}_{1\leq t <s_{Q}}$.}

\State{Split the source data $\mathcal{D}^{P}_{\mathsf{dm}} \gets \Big\{ \big(X_{i}^{P},\pi_{i}^{P},Y^{P,(\pi_{i}^{P})} \big) \Big\}_{i=s_{P}}^{n_{P}}$ to aid in decision-making in the $Q$-bandit.}

\State{Set $\mathcal{L}_{s_{Q}}\gets\{\mathcal{X}\}$, and $\mathcal{I}(\mathcal{X})\gets\mathcal{I}$.}
    \Comment{initialize partition and arm set}
\State{Initialize the policy $\widetilde{\pi}(\mathcal{X})$ by Procedure~\ref{alg:EA-TL}
$\big(\mathcal{X}, \mathcal{I}(\mathcal{X}), \widehat{U}, \mathcal{D}^{P}_{\mathsf{dm}} \big)$.}
\State{Initialize $N(\mathcal{X})\gets0$.}
    \Comment{set time to $0$ for policy $\widetilde{\pi}(\mathcal{X})$}
\State{Initialize $\tau_{k}(\mathcal{X}) \gets 0$, and $\widehat{\tau}_{k}^{\star}(\mathcal{X};\mathcal{D}^{P}_{\mathsf{dm}})$ as in \eqref{eq:tau-star-adaptive}, $\forall k \in \mathcal{I}(\mathcal{X})$.}
    \Comment{initialize rounds and round upper bounds}
\For{$t=s_{Q},\dots,n_{Q}$}
    \State{Draw a sample $X_{t}^{Q} \sim Q_{X}$.}
    \State{Find the bin $B\in\mathcal{L}_{t}$ such that $X_{t}^{Q}\in B$.}
    \While{$|\mathcal{I}(B)|>1$ and $\tau_{k}(B)\geq\widehat{\tau}_{k}^{\star}(B;\mathcal{D}^{P}_{\mathsf{dm}}), \forall k\in\mathcal{I}(B)$}
        \Comment{keep partitioning $B$ until reaching suitable scale}
    \If{$\widehat{\tau}_{k}^{\star}(B;\mathcal{D}^{P}_{\mathsf{dm}}) = 0, \forall k\in\mathcal{I}(B)$}
        \Comment{no exploration needed in $B$: discard suboptimal arms}
    \State{Set $\underline{Y}^{\star}(B;\mathcal{D}^{P}_{\mathsf{dm}})\gets\max\limits_{k\in\mathcal{I}(B)}\big\{\overline{Y}_{k}^P(B;\mathcal{D}^{P}_{\mathsf{dm}})-\widehat{U}_{k}(0,B;\mathcal{D}^{P}_{\mathsf{dm}})\big\}$}
        \Comment{set largest reward lower bound}
    \State{Set $\mathcal{I}(B) \gets \{ k \in \mathcal{I}(B) : \overline{Y}_k^P(B;\mathcal{D}^{P}_{\mathsf{dm}}) + \widehat{U}_{k}(0,B;\mathcal{D}^{P}_{\mathsf{dm}}) \geq \underline{Y}^{\star}(B;\mathcal{D}^{P}_{\mathsf{dm}}) \}$.}
        \Comment{update arm set}
    \EndIf
    \For{$B'\in\mathsf{child}(B)$}
    \State{Set $\mathcal{I}(B')\gets\mathcal{I}(B)$.}
        \Comment{assign remaining arms in $B$ as initial arms in its children}
    \State{Initialize the policy $\widetilde{\pi}({B'})$ by Procedure~\ref{alg:EA-TL} $\big(B', \mathcal{I}(B'),\widehat{U},\mathcal{D}^{P}_{\mathsf{dm}} \big)$.}
    \State{Set $N(B')\gets0$.}
        \Comment{initialize time for policy $\widetilde{\pi}(B')$}
    \State{Set $\tau_{k}(B') \gets 0$, and  $\widehat{\tau}_{k}^{\star}(B';\mathcal{D}^{P}_{\mathsf{dm}}) $ as in \eqref{eq:tau-star-adaptive}, $\forall k\in\mathcal{I}(B')$.}
        \Comment{initialize rounds and round upper bounds}
\EndFor
\State{Set $\mathcal{L}_{t}\gets (\mathcal{L}_{t}\setminus B)\cup\mathsf{child}(B)$.}
    \Comment{replace $B$ with its children in partition}
\State{Find the bin $B\in\mathcal{L}_{t}$ such that $X_{t}^{Q}\in B$.}
\EndWhile
\State{Set $N(B)\gets N(B)+1$.}
    \Comment{update times $X_{t}^{Q} \in B$}
\State{Set $\pi^{\adp}_{t}\gets\widetilde{\pi}_{N(B)}(B)$.}
    \Comment{choose arm by policy $\widetilde{\pi}(B)$}
\State{Set $\mathcal{I}(B)\gets\widetilde{\mathcal{I}}_{N(B)} (B)$.}
    \Comment{update arm set by policy $\widetilde{\pi}(B)$}
\State{Set $\tau_{k}(B) \gets \widetilde{\tau}_{N(B),k}(B), \forall k\in\mathcal{I}(B)$.}
    \Comment{update numbers of rounds by policy $\widetilde{\pi}(B)$}
\EndFor
\State\textbf{Output:} policy $\pi^{\adp} = \{\pi^{\adp}_{t}\}_{t\geq1}$.

\end{algorithmic}
\end{algorithm}

\addtocounter{algorithm}{-1}
\floatname{algorithm}{Procedure}
\begin{algorithm}[t]
\caption{Smoothness estimation procedure}
\label{alg:smoothness} \begin{algorithmic}[1]

\State{\textbf{Input:} arm number $K$, horizon length $n_{Q}$, lower and upper bounds on smoothness $\underline{\beta}$ and $\overline{\beta}$, upper bound on transfer exponent $\overline{\gamma}$, $P$-data $\mathcal{D}^{P}$, tuning parameters $C_{1}, C_{2}$.}
\State{Set $n\gets n_{P}\vee n_{Q}$.}
\If{$n_{P}>n_{Q}$}
\State{Set $l_{1}\gets\Big\lceil\frac{\underline{\beta}\log_{2}(n)}{(d+2\overline{\beta}+\overline{\gamma})^2}\Big\rceil$, $l_{2}\gets l_{1}+\Big\lceil\frac{1}{d}\log_{2}(\log(n))\Big\rceil$, $l_{3}\gets\Big\lceil \frac{\overline{\beta}}{\underline{\beta}}l_{1}+\frac{1}{\underline{\beta}}\log_{2}(\log(n))\Big\rceil$.}
    \Comment{bandwidths of reward function estimators}
\State{Set $T\gets \Big\lceil C_{1} K n^{\frac{\underline{\beta}}{d+2\overline{\beta}}}\log^{\frac{d+\overline{\gamma}}{d}}(Kn)\Big\rceil$.}
    \Comment{sample size for smoothness estimation}
\State{Set $\mathcal{D}_{\mathsf{se}} \gets \Big\{\big(X_{i}^{P},\pi_{i}^{P},Y_{i}^{P,(\pi_{i}^{P})}\big)\Big\}_{i=1}^{T}$, $s_{P}\gets T+1$, $s_{Q}\gets1$.} 
    \Comment{samples for smoothness estimation}
\Else
\State{Set $l_{1}\gets\Big\lceil\frac{\underline{\beta}\log_{2}(n)}{(d+2\overline{\beta})^2}\Big\rceil$, $l_{2}\gets l_{1}+\Big\lceil\frac{1}{d}\log_{2}(\log(n))\Big\rceil$, $l_{3}\gets\Big\lceil \frac{\overline{\beta}}{\underline{\beta}}l_{1}+\frac{1}{\underline{\beta}}\log_{2}(\log(n))\Big\rceil$.}
    \Comment{bandwidths of reward function estimators}
\State{Set $T\gets \Big\lceil C_1 n^{\frac{\underline{\beta}}{d+2\overline{\beta}}}\log(Kn)\Big\rceil$.}
    \Comment{sample size for smoothness estimation}
\For{$t = 1,\dots, T$}
\State{Draw a sample $X_{t}^{Q}$, pull arm $\pi_{t}$ uniformly at random from $[K]$, and get the reward $Y_{t}^{Q,(\pi_{t})}$.}
\EndFor
\State{Set $\mathcal{D}_{\mathsf{se}} \gets \Big\{\big(X_{t}^{Q},\pi_{t},Y_{t}^{Q,(\pi_{t})}\big)\Big\}_{t=1}^{T}$, $s_{P}\gets1$, $s_{Q}\gets T+1$.}
    \Comment{samples for smoothness estimation}
\EndIf

\State Define the grid $\mathcal{M}$
\begin{align*}
\mathcal{M}\coloneqq\big\{x\in\mathcal{X}:x_{i}=(k_{i}-1/2)2^{-l_{3}},\,k_{i}=[2^{l_{3}}],i=[d]\big\}.
\end{align*}

% \For{$i\in\{1,2\},$}
% \For{$k\in[K]$}
\For{$i\in\{1,2\},k\in[K]$}
\State{Set $\mathcal{D}_{\mathsf{se}}^{(k)} \gets \{(X_{t},Y_{t}) : (X_{t},\pi_{t},Y_{t}) \in\mathcal{D}_{\mathsf{se}}\,\, \text{ such that }\,\, \pi_{t} = k \}$.}
\For{$x\in\mathcal{M}$}
\State{Find the bin $B_{i}(x)\in\mathcal{B}_{l_{i}}$ such that $x\in B_{i}(x)$.}
\State{Define the reward function estimator $\widehat{f}_{k}(x;2^{-l_{i}}) \defn \widehat{\eta}_{k}\big(x;B_{i}(x)\big)$, where
\begin{align}
\widehat{\eta}_{k}(x;B) \defn \frac{\sum_{(X_{t},Y_{t})\in\mathcal{D}_{\mathsf{se}}^{(k)}}Y_{t}\mathds{1}\{X_{t}\in B\}}
{1 \vee \sum_{(X_{t},Y_{t})\in\mathcal{D}_{\mathsf{se}}^{(k)}}\mathds{1}\{X_{t}\in B\}},\quad \forall B\in\mathcal{B}_{l_{i}}. \label{def:local-estimator}
\end{align}
% if $\sum_{(X_{t},Y_{t})\in\mathcal{D}_{\mathsf{se}}^{(k)}}\mathds{1}\{X_{t}\in B\} > 0$, and $\widehat{\eta}_{k}(x;B) \defn 0$ otherwise.
}
% \EndFor
\EndFor
\EndFor
\State {Set $\mathsf{b}\gets\max_{k\in[K]}\max_{x\in\mathcal{M}}\big|\widehat{f}_{k}(x;2^{-l_{1}})-\widehat{f}_{k}(x;2^{-l_{2}})\big|$.} 
\State {Set $\widehat{\beta}\gets-\frac{1}{l_{1}}\log(\mathsf{b})-C_{2}\frac{\log_{2}(\log (n))}{\log_{2}(n)}$.}
\State {\textbf{Output:} smoothness estimate $(\underline{\beta}\vee\widehat{\beta})\wedge\overline{\beta}$, time steps $s_{P}$ and $s_{Q}$, policy $\{\pi_{t}\}_{1\leq t<s_{Q}}$.}
\end{algorithmic}
\end{algorithm}

In the first phase, our primary objective is to estimate the H\"{o}lder smoothness parameter $\beta$ under the self-similarity condition. The procedure is rooted in the critical insight that the local piecewise-constant regression estimator of a function $f$ over a bin $B$ is sufficiently close to its piecewise-constant projection $\Gamma_{B}f$ with high probability. 
Hence, when applying the local piecewise-constant regression method to estimate the reward function, the self-similarity condition ensures that the estimation bias does not decay too rapidly. Combined with the H\"older smooth condition, this yields a tight bound on the estimation bias, which depends on the smoothness parameter $\beta$. 
Even though we lack direct access to the estimation bias, we can adapt Lepski's method \citep{lepskii1991problem,lepskii1992asymptotically,lepskii1993asymptotically,lepski1997optimal} suitably to obtain a reliable estimate by comparing the difference between estimators with different bin side lengths. To this end, Procedure~\ref{alg:smoothness} first creates two partitions of the covariate space by using bins of different sizes. Next, based on these two partitions, it collects independent samples $\mathcal{D}_{\mathsf{se}}$ to construct two separate local piecewise-constant regression estimators for each reward function in each bin. Clearly, the estimation bias will be larger in the bin with a larger side length. As long as we collect adequate samples such that the estimation bias dominates the standard deviation, the maximal difference between the two regression estimates is approximately of the same order as the larger estimation bias. This allows us to infer the smoothness of the reward functions. In particular, with high probability, we can obtain a smoothness estimate $\widehat{\beta}$ with statistical guarantee $\beta - O(\log_2 (\log (n)) / \log_2 (n)) \leq \widehat{\beta} \leq \beta$, which suffices to attain a near-optimal regret. The detailed smoothness estimation procedure is presented in Procedure~\ref{alg:smoothness}. 
It is worth noting that depending on the relationship between $n_{P}$ and $n_{Q}$, the samples for the smoothness estimation are collected from either the $Q$-bandit or $P$-bandit. 
In the former case, the smoothness estimation phase only takes a vanishingly small portion of the time horizon length $n_Q$ in the $Q$-bandit. Thus, the regret incurred during this stage is negligible compared to the minimax regret. On the other hand, when $n_{Q} < n_{P}$, we split the source samples for the smoothness estimation and for the decision-making in the target bandit separately. Similarly, the sample size used for the smoothness estimation is considerably smaller than the total source sample size $n_{P}$ and has no impact on the minimax regret. 

With the smoothness estimate $\widehat{\beta}$ in hand, the second stage of Algorithm~\ref{alg:UCB-TL-adaptive} takes it as an input and operates in a manner similar to Algorithm~\ref{alg:UCB-TL}. We would like to highlight several pivotal differences. To begin with, as a portion of the source data may have been used in the smoothness estimation process, Algorithm~\ref{alg:UCB-TL-adaptive} utilizes a subset of the source dataset $\mathcal{D}^{P}_{\mathsf{dm}}\subset \mathcal{D}^{P}$ to assist in distinguishing the optimal arm in the target bandit. Additionally, note that the confidence bound $U_{k}(\tau,B;\mathcal{D})$ (cf.~\eqref{eq:UCB}) used in Algorithm~\ref{alg:UCB-TL} requires the knowledge of the unknown parameters $\gamma$ and $\kappa$. To construct an adaptive procedure in Algorithm~\ref{alg:UCB-TL-adaptive}, we substitute it with the confidence bound $\widehat{U}_{k}(\tau,B;\mathcal{D})$ defined as follows. Given a subset $\mathcal{D}$ of the source data $\mathcal{D}^{P}$ and the smoothness estimate $\widehat{\beta}$ returned by Procedure~\ref{alg:smoothness}, for any non-negative integer $\tau \geq 0$, bin $B$, and arm $k$, we define
\begin{align}
\widehat{U}_{k}(\tau,B;\mathcal{D})\coloneqq
\begin{cases}
2\sqrt{\frac{2}{\tau+n_{k}^{P}(B;\mathcal{D})}\log^{+}\big(\frac{n_{Q}|B|^{d}}{\tau}\big)} \vee 2\overline{C}_{\beta}|B|^{\widehat{\beta}}, & \text{if } \tau > 0, \\
2\sqrt{\frac{2}{n_{k}^{P}(B;\mathcal{D})}\log(n|B|^{d+2\widehat{\beta}})} \vee 2\overline{C}_{\beta}|B|^{\widehat{\beta}}, & \text{if } \tau = 0,
\end{cases}
 \label{eq:UCB-adapt}
\end{align}
where we recall the notations $\log^{+}(x)\defn\log(x)\vee1$, $n\defn n_{P}\vee n_{Q}$, and $1/0 = \infty$.
The associated non-negative upper bound on play rounds is defined by
\begin{align}
\widehat{\tau}^\star_k(B;\mathcal{D}) \defn \min_{\tau \in \{0\}\cup\mathbb{N}}\big\{ \tau : \widehat{U}_{k}(\tau,B;\mathcal{D}) \leq 2\overline{C}_{\beta}|B|^{\widehat{\beta}}\big\}. \label{eq:tau-star-adaptive}
\end{align}
It is noteworthy that the source sample size within each bin in the partition tree is highly concentrated around its expectation, and that our choice of $\widehat{\tau}_{k}^{\star}(B;\mathcal{D})$ achieves a balanced bias-variance trade-off for estimating the reward functions in each bin. This allows Algorithm~\ref{alg:UCB-TL-adaptive} to automatically adapt to the unknown parameters $\gamma$ and $\kappa$, leading to a near-optimal regret. 

Finally, it is worth mentioning that while Algorithm~\ref{alg:UCB-TL-adaptive} requires an upper bound on the transfer exponent $\overline{\gamma}$, this information is solely utilized in the smoothness estimation stage (i.e.~Procedure~\ref{alg:smoothness}). As previously discussed, our smoothness estimation procedure uses $T$ independent samples to evaluate the estimation bias of the local regression estimator under the self-similarity condition. To guarantee a good statistical performance of the local estimator in a bin with side length $h$, it is crucial to gather enough samples to balance the standard deviation (of order $(1 / \sqrt{Th^{d+\gamma}}$) with the estimation bias (of order $h^{\beta}$). On the other hand, we also need to ensure that the regret incurred during the smoothness estimation phase is relatively small compared to the minimax regret. Therefore, achieving these two objectives necessitates the knowledge of an upper bound on the transfer exponent $\overline{\gamma}$. As an important remark, once the smoothness parameter estimate $\widehat{\beta}$ is generated by~Procedure~\ref{alg:smoothness}, $\overline{\gamma}$ is no longer used in the second phase of Algorithm~\ref{alg:UCB-TL-adaptive}. 
% In other words, suppose we have an access to an accurate smoothness estimate $\widehat{\beta}$, then our transfer learning algorithm is fully adaptive to the unknown transfer exponent $\gamma$.

We now present the theoretical guarantees of Algorithm \ref{alg:UCB-TL-adaptive}.
The proof is postponed to Appendix~\ref{sec:Proof-of-Adaptivity} in the supplementary material~\citep{cai2022transfer-supp}.

\begin{theorem}[Upper bound]\label{thm:upper-bound-adaptive}
Let $0 < \underline{\beta} < \overline{\beta} \leq 1$ and $\overline{\gamma} \geq 0$. Suppose that $\kappa\asymp1$ and $\alpha \beta \leq d$.
Then the policy $\pi^{\adp}$ given by Algorithm~\ref{alg:UCB-TL-adaptive} satisfies that for all $\beta \in [\underline{\beta}, \overline{\beta}]$ and $\gamma \in [0, \overline{\gamma}]$,
\begin{align}
\sup_{\Pi(K,\beta,\alpha,\gamma, \kappa ,l_{0},b)}\mathbb{E}[R_{n_{Q}}(\pi^{\adp})]\leq C_{1}n_{Q}\Big(n_{Q} + (\kappa n_{P})^{\frac{d+2\beta}{d+2\beta+\gamma}}\Big)^{-\frac{\beta(1+\alpha)}{d+2\beta}}\log^{C_{2}}(n_{P}+n_{Q}), \label{eq:upper-bound-adaptive}
\end{align}
for some constants $C_{1} > 0$ and $C_{2}>0$ independent of $n_Q$ and $n_P$.
\end{theorem}

All in all, Theorem \ref{thm:upper-bound-adaptive} demonstrates that Algorithm~\ref{alg:UCB-TL-adaptive} achieves the near-optimal minimax regret simultaneously for all $\beta\in[\underline{\beta},\overline{\beta}]$ and $\gamma \in [0, \overline{\gamma}]$ when $K,\kappa\asymp1$. 
In comparison with Theorem \ref{thm:upper-bound}, the regret upper bound (\ref{eq:upper-bound-adaptive}) contains an additional logarithmic factor, which can be viewed as the cost paid for smoothness adaptation. The condition $\kappa\asymp1$ essentially assumes that the sample sizes corresponding to each arm in the source data are roughly of the same order, where one can achieve the most effective transfer learning. 

Moreover, Theorem \ref{thm:lower-bound-adaptive} below shows that the self-similarity assumption does not reduce the minimax complexity of the problem. The proof is deferred to Appendix~\ref{sec:Proof-of-Lower-Bound} in the supplementary material \citep{cai2022transfer-supp}.

\begin{theorem}[Lower bound]\label{thm:lower-bound-adaptive}Assume that $\alpha \beta \leq d$.
For any constant $\beta\in(0,1]$ and $ l_0 \geq 0$, there exists a constant $b>0$ that only depends on $\beta, C_\beta, \underline{q}, \overline{q}$ and $d$ such that
\begin{align}
\inf_{\pi}\sup_{\Pi(K,\beta,\alpha,\gamma, \kappa ,l_{0},b)}
\mathbb{E}[R_{n_Q}(\pi)]\geq c n_{Q}\Big(n_{Q} + (\kappa n_{P})^{\frac{d+2\beta}{d+2\beta+\gamma}}\Big)^{-\frac{\beta(1+\alpha)}{d+2\beta}}.
\end{align}
for some constant $c>0$ independent of $n_Q$ and $n_P$.
\end{theorem}

Similar to Theorem~\ref{thm:lower-bound}, the infimum is taken over the class of admissible policies. Recognizing that the self-similar function space $\Pi(K,\beta,\alpha,\gamma, \kappa,l_{0},b)$ is a subset of the general space $\Pi(K,\beta,\alpha,\gamma, \kappa )$, Theorem \ref{thm:lower-bound-adaptive} and \ref{thm:lower-bound} together demonstrates that the minimax regret under the self-similar condition is the same as that in the general case. As a result, the self-similar condition does not reduce the complexity of the problem.

\begin{remark}
    \label{rem:diff}
    We would like to remark that \citet{suk2021self} has also studied nonparametric contextual multi-armed bandits under the covariate shift model, with a particular focus on Lipschitz reward functions ($\beta =1$). However, several significant distinctions exist between the analysis in our current work and that in \citet{suk2021self}. For instance, a central challenge in our work is achieving smoothness adaptivity. Integrating the source dataset to attain the minimax regret in the target bandit while at the same time adapting to the unknown smoothness parameter requires a substantially more complicated algorithmic design and technical analysis. 
    Also, \citet{suk2021self} assumed the permission to collect data from the source bandit, allowing for active exploration. In contrast, our work deals with a fixed, pre-collected source dataset. This limitation means that the source data might have been generated by a certain behavior policy, which might not provide sufficiently many data samples for important context-arm pairs. Effectively handling this limited data coverage becomes a critical challenge that governs the statistical efficiency of transfer learning.
\end{remark}

\section{Discussion}

\label{sec:Discussion}

In this paper, we have studied transfer learning for nonparametric contextual multi-armed bandits under the covariate shift model. We establish the minimax regret that captures the amount of information transferred from the source domains to the target domain. A novel transfer learning algorithm is proposed to attain the minimax regret. Moreover, we also develop a data-driven algorithm that achieves within a logarithmic factor of the minimax regret while adapting to the unknown smoothness over a large class of parameter spaces under the self-similarity assumption.

There are several possible extensions worth pursuing. To begin with, while the current paper focuses on ``rough'' reward functions with smoothness parameter $\beta \in (0,1]$, it is conceivable to generalize the algorithmic ideas to the case $\beta > 1$---one can adaptively partition the covariate space coupled with static multi-armed bandit procedures. However, we would like to remark on several critical differences. First, in contrast to the case $\beta \in (0,1]$ where local piecewise constant estimators suffice to estimate the reward functions, one needs to use more complicated local polynomial estimators in the case $\beta > 1$. In addition, in our case $\beta \in (0,1]$, as reward functions may be non-differentiable, only samples close to the observed covariate are informative about their corresponding reward functions. Consequently, a fully localized learning strategy---running static multi-armed bandit procedures separately within each bin---guarantees to attain the minimax regret. Unfortunately, this rate-optimality no longer holds in the case $\beta > 1$. As the reward functions become smoother, we need to leverage global information and utilize observations from neighboring bins to extrapolate the reward functions efficiently, as introduced in \citet{hu2019smooth}. Furthermore, since samples are used across bins, this also leads to the statistical dependence between decision-making in different bins. Therefore, the resulting policy is rather complicated and requires more careful statistical analysis. We leave it to future investigation due to the space limit.

Additionally, the regret upper bound of our adaptive policy mismatches the minimax rate by a logarithmic term. Whether this logarithmic factor is an inherent consequence of not knowing smoothness or an artifact of the proof remains unclear. It has been widely recognized in the literature on nonparametric function estimation that sharp adaptation is often achievable under global integrated squared error and that a logarithmic penalty arises under pointwise squared error. In contrast, when constructing confidence intervals, adaptation to unknown smoothness without additional structural assumptions is typically impossible unless self-similarity or shape constraints are present. Notably, it was shown in \citet{cai2013adaptive} that adaptive confidence intervals can be constructed for regression functions under shape constraints, such as concavity, and are near optimal for every individual function. Therefore, it is interesting to investigate further the cost of adaptation in the bandit problems, especially in the context of transfer learning. For example, developing adaptive transfer learning procedures for nonparametric contextual multi-armed bandits with concave reward functions presents an appealing direction.

Lastly, it is intriguing to study transfer learning for nonparametric contextual multi-armed bandits under other models. For instance, one avenue for future study is the posterior drift model, where the marginal distributions of the target and source bandits are identical whereas the conditional reward distributions differ. In this framework, it is also of interest to characterize the similarity between the reward distributions, establish the minimax rate of convergence that quantifies transferable information, and develop data-driven adaptive algorithms.

\appendix

\section{Numerical experiments}

\label{sec:Numerical-experiments}

In this section, we present a series of numerical experiments to demonstrate the effectiveness of our proposed transfer learning procedures. All results displayed here are averaged over $200$ independent Monte Carlo trials.

Set $d = 2$. We generate data under the covariate shift model, where the marginal distributions $P_X$ and $Q_X$ differ whereas the reward distributions are the same under $P$ and~$Q$. 

For the marginal distributions, we set $Q_{X}$ as the uniform distribution over $[0,1]^2$. And $P_{X}$ is chosen such that the density function $p_X(x)$ obeys $p_X(x) = c \, \| x - (1/2, 1/2) \|_{\infty}^\gamma$ for some absolute constant $c > 0$. 

Turning to the reward distributions, we assume that conditioned on the covariate $x$, the observed reward $Y^{(k)} \sim \mathcal{N}(f_k(x), \sigma^2)$ is a Gaussian random variable with mean $f_k(x)$ and standard deviation $\sigma = 0.05$ for any $k\in[K]$. We make a note that although the current paper focuses on bounded rewards for simplicity of presentation, the theoretical guarantees can be easily extended to the sub-Gaussian reward case. 
As for the reward functions, let us define a function $\varphi_\beta(x):[0, \infty) \rightarrow [0,1] $ by $\varphi_{\beta}(x) = (1-x)^\beta \mathds{1}\{x \leq 1\}$. Based on function $\varphi_{\beta}(x)$, we set the reward function $f_{k}:[0,1]^2\rightarrow [0,1]$ as $f_1(x) \equiv 1/2$, and for any $k\neq 1$,
\begin{align*}
f_{k}(x)  = \frac{1}{2} + \frac{1}{2} \sum_{i = 1}^{4} \omega_{k,i} \varphi_\beta( 8 \| x - c_{k,i} \|_\infty),
\end{align*}
where the sign $\omega_{k,i} = \pm 1$ is drawn i.i.d.~with probability $1/2$ for any $i\in[4]$, and the centers $ \{ c_{k, i} \}_{i\in[4]}$ are set as $ \{ c_{k, i} \} = \{ (1/2 \pm 1/4, 1/2 \pm 1/4) + \mathsf{Unif}([0,1/8]^2) \}$. 

In addition, the pulled arm in the source data is generated i.i.d.~with arm selection distribution satisfying $\mu(\cdot \,|\,x) \equiv \mu(\cdot)$ for any $x \in \mathcal{X}$, namely, each arm $k$ is pulled with probability $\mu(k)$ at any $x\in\mathcal{X}$. 
One can verify that the construction of distributions $P$ and $Q$ satisfies the assumptions introduced in Section~\ref{sec:Problem-formulation}.

For the sake of comparison, we also plot the numerical performance of the ABSE algorithm proposed in \citet{perchet2013multi}, in addition to our proposed transfer learning methods. The ABSE algorithm is known to attain the minimax regret in the standard setup where no additional source data is available. Therefore, it serves as a benchmark to illustrate the performance improvement gained from transfer learning.

\begin{figure}[t]
\centering
\begin{tabular}{ccc}
\includegraphics[width=0.31\textwidth]{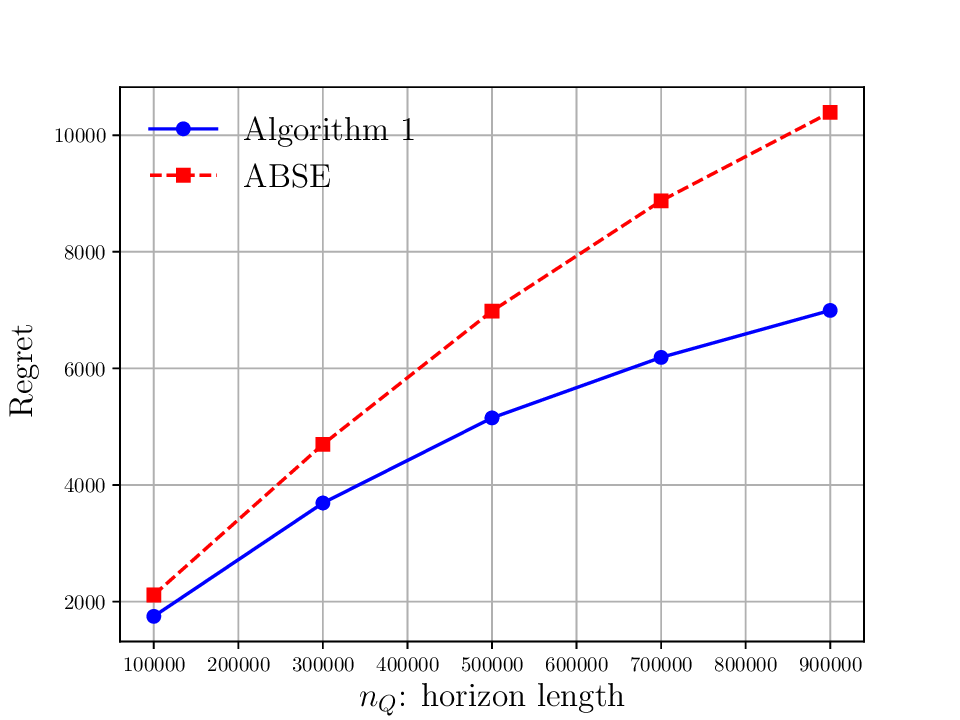} 
& \includegraphics[width=0.31\textwidth]{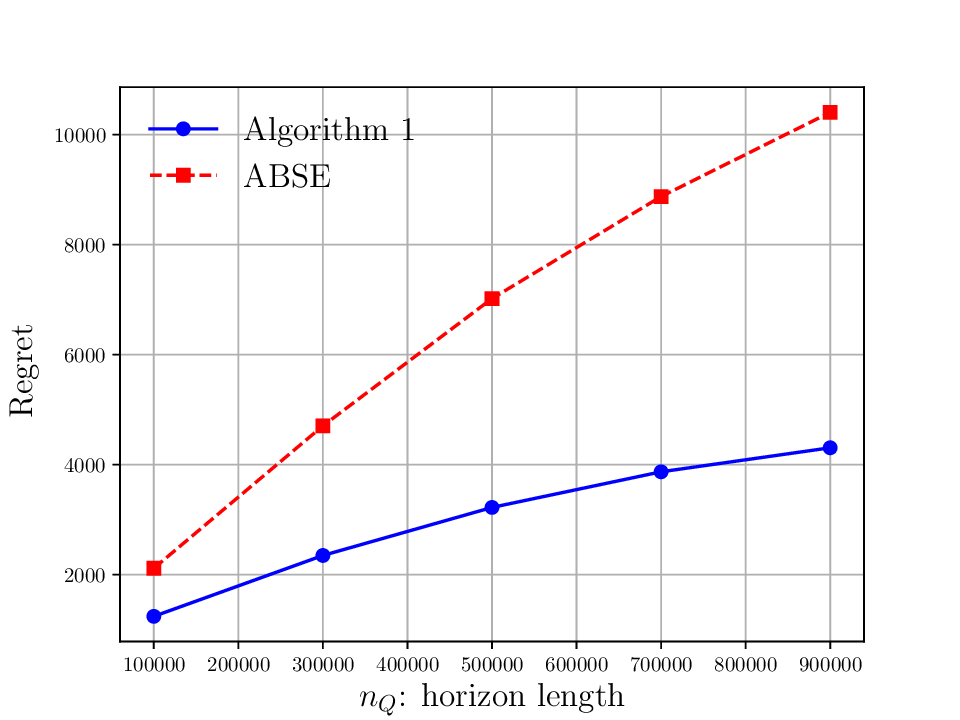}
& \includegraphics[width=0.31\textwidth]{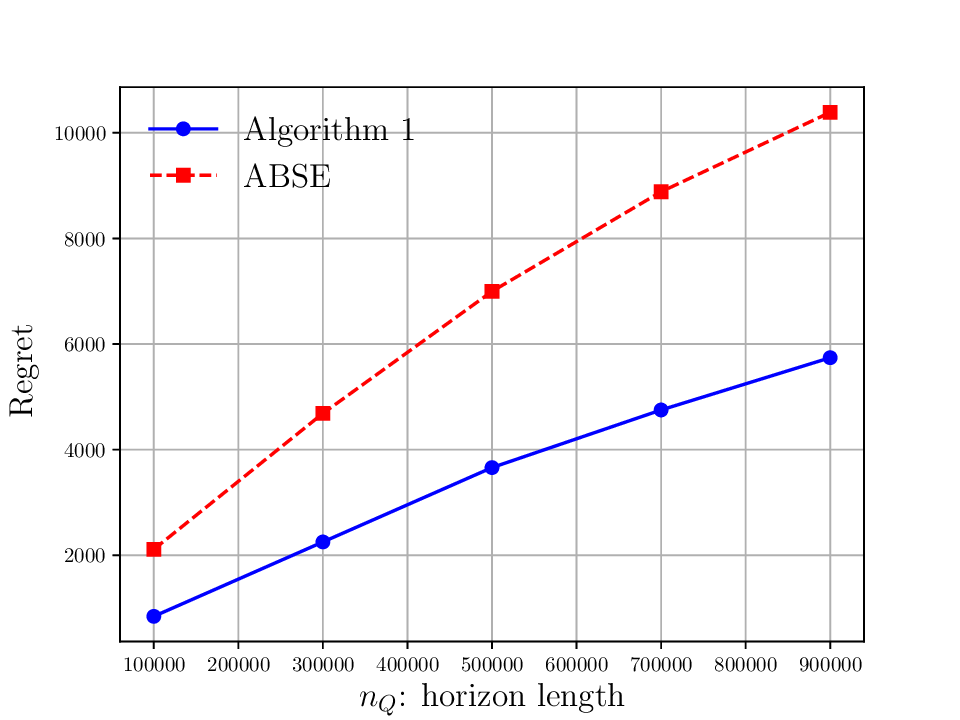} 

\tabularnewline
(a) & (b) & (c) \tabularnewline
\end{tabular}
\caption{(a) Regret vs.~horizon length $n_Q$ with $n_P =  n_Q / 2$; (b) Regret vs.~horizon length $n_Q$ with $n_P = 3 n_Q$; (c) Regret vs.~horizon length $n_Q$ with $n_P = 10 \times 10^{5}$. Here, $d=2, K=2, \beta = 0.8$, $ \gamma = 1$, and $\kappa = 1$. \label{fig:reg_nQ}}
\end{figure}

\begin{figure}[t]
        \centering
        \begin{tabular}{ccc}
        \includegraphics[width=0.31\textwidth]{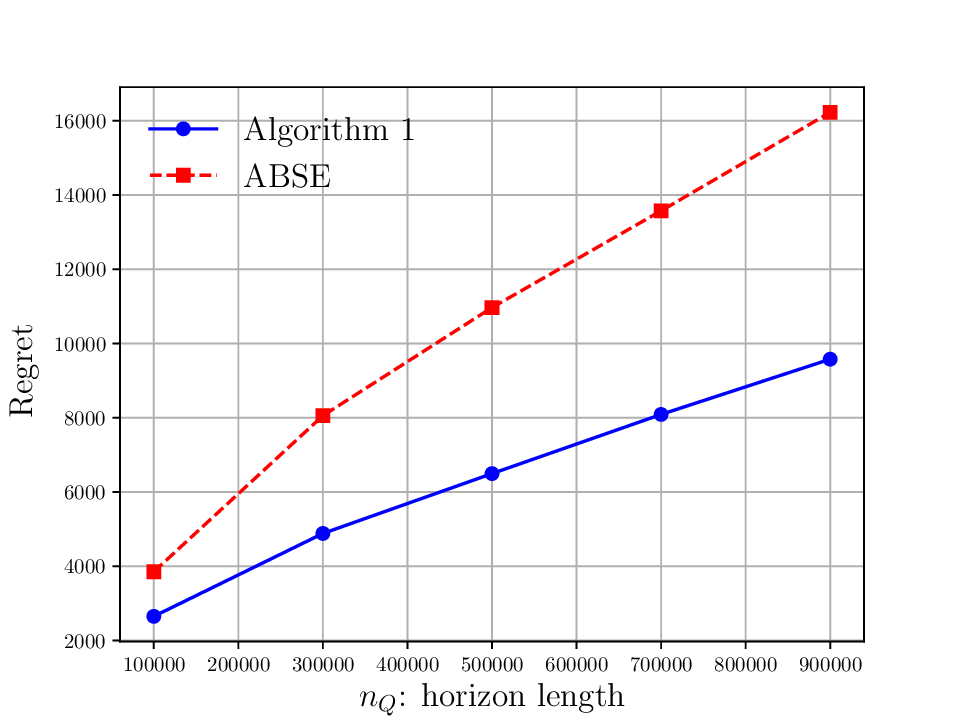} 
         & \includegraphics[width=0.31\textwidth]{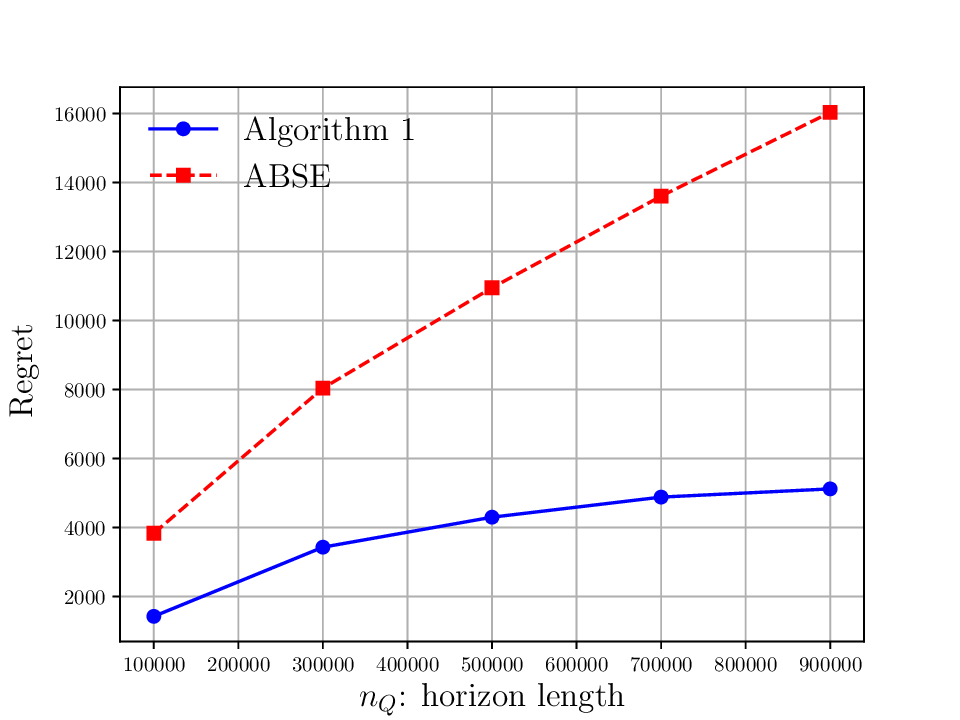}
        & \includegraphics[width=0.31\textwidth]{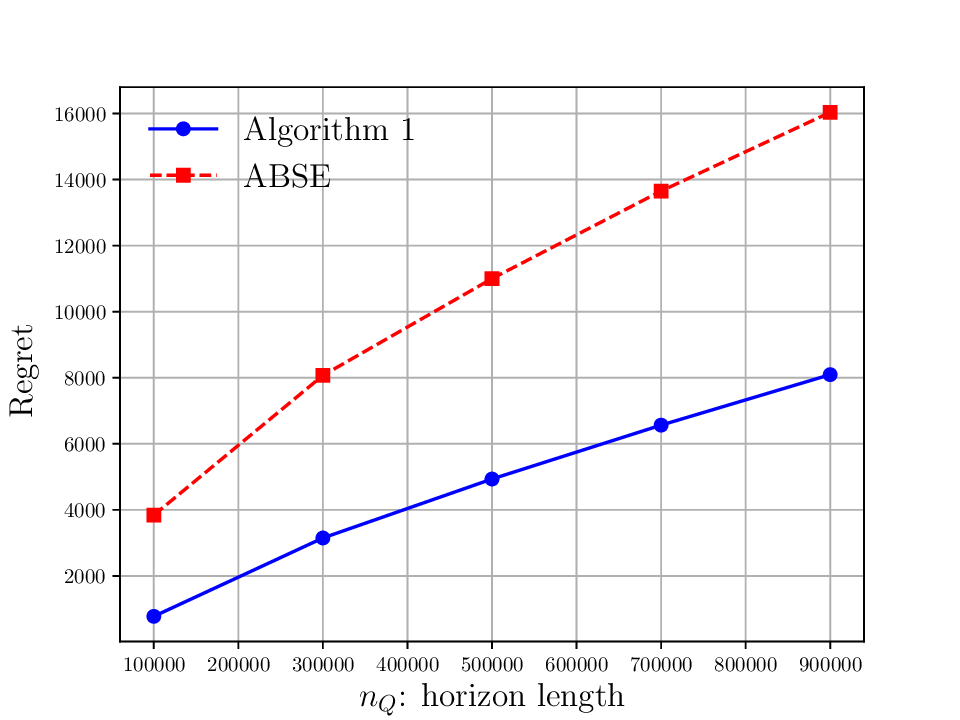} 

        \tabularnewline
        (a) & (b) & (c) \tabularnewline
        \end{tabular}
        \caption{(a) Regret vs.~horizon length $n_Q$ with $n_P =  n_Q / 2$; (b) Regret vs.~horizon length $n_Q$ with $n_P = 3 n_Q$; (c) Regret vs.~horizon length $n_Q$ with $n_P = 10 \times 10^{5}$. Here, $d=2, K=4, \beta = 0.8$, $ \gamma = 1$, and $\kappa = 1$. \label{fig:reg_nQ_K4}}
        \end{figure}

\paragraph*{Minimax optimal algorithm}
We start with the numerical performance of Algorithm~\ref{alg:UCB-TL} when given complete knowledge of parameters $\beta$, $\gamma$, and $\kappa$, which provably achieves the minimax regret in the transfer learning setting. For the case $K=2$, Figure~\ref{fig:reg_nQ}~(a), (b), and (c) plot the regret vs.~the horizon length $n_Q$ for $n_P = n_Q / 2$, $n_P = 3 n_Q $, and $n_P = 3 \times 10^{5}$, respectively. In addition, for the case $K=4$, Figure~\ref{fig:reg_nQ_K4}~(a), (b), and (c) plot the regret vs.~the horizon length $n_Q$ for $n_P = n_Q / 2$, $n_P = 3 n_Q $, and $n_P = 10 \times 10^{5}$, respectively.  As can be seen, Algorithm~\ref{alg:UCB-TL} consistently outperforms the ABSE algorithm, thereby demonstrating the advantage of incorporating the source data.

\begin{figure}[t]
\centering
\begin{tabular}{ccc}
\includegraphics[width=0.31\textwidth]{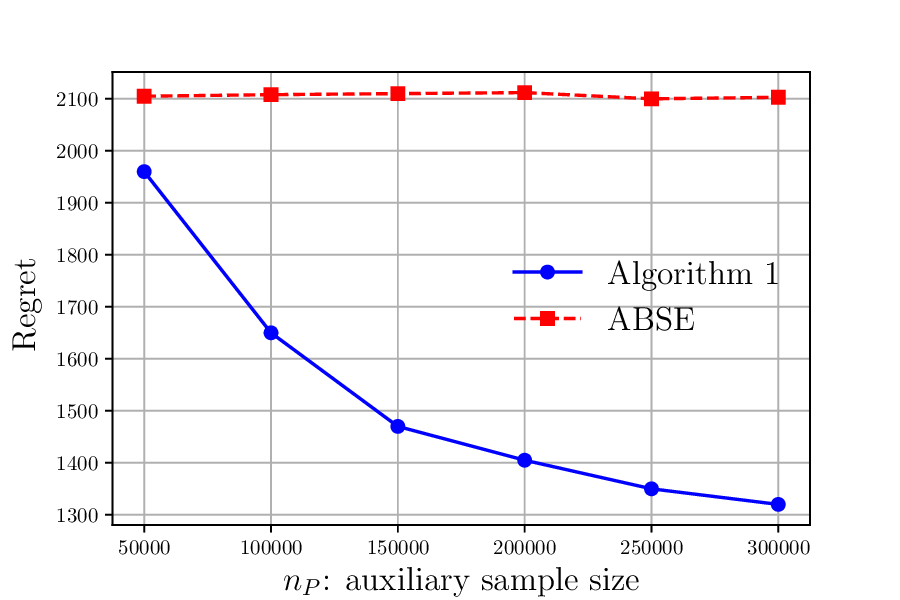} 
& \includegraphics[width=0.31\textwidth]{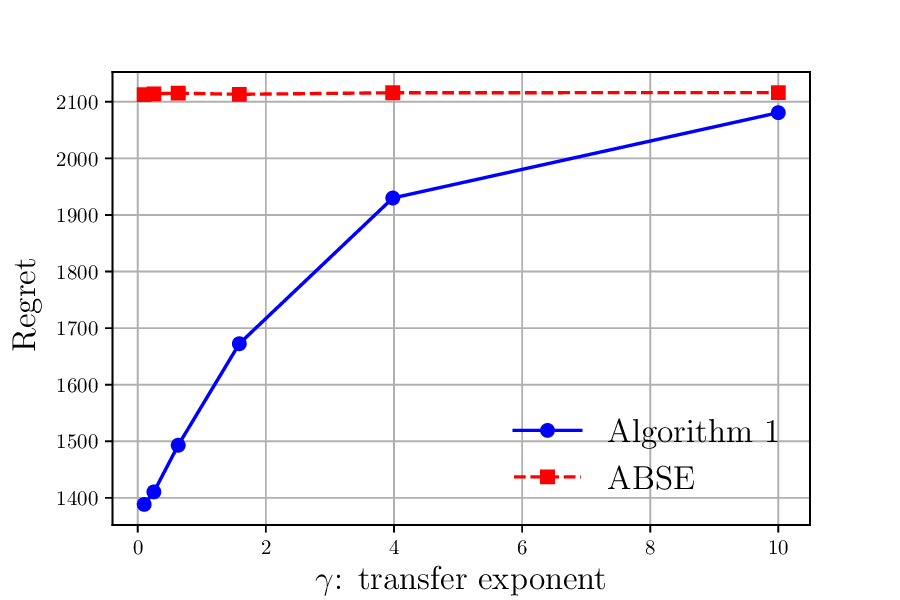} 
& \includegraphics[width=0.31\textwidth]{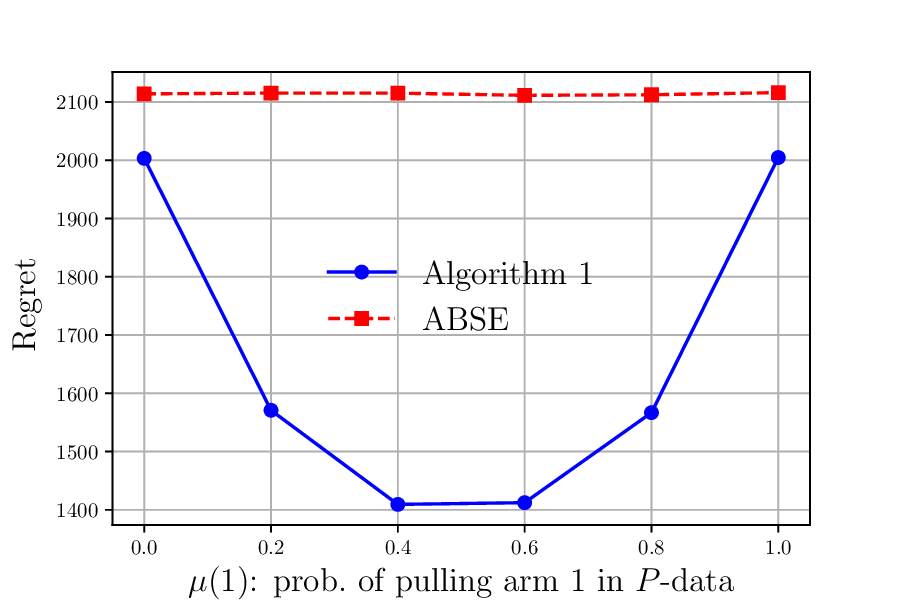} 
\tabularnewline
(a) & (b) & (c) \tabularnewline
\end{tabular}
\caption{(a) Regret vs.~auxiliary sample size $n_P$ with $\gamma = 1, \kappa = 1$; (b) Regret vs.~transfer exponent $\gamma$ with $n_P = 2 \times 10^{5}, \kappa = 1$; (c) Regret vs.~auxiliary policy probability $\mu(1)$ with $n_P = 2 \times 10^{5}, \gamma = 1$. Here, $d=2, K=2, \beta = 0.8$, and $n_Q = 1 \times 10^{5}$. \label{fig:reg_P}}
\end{figure}

We proceed to investigate the roles of the parameters that govern transfer learning from source domains. Let us begin by studying the influence of the sample size of the source dataset $n_P$. Set $K=2$. Figure~\ref{fig:reg_P}~(a) displays the regret vs.~the source sample size $n_P$. As one can see, there is an evident gap between the regret of our procedure and that of the ABSE algorithm. Naturally, one can anticipate an increasingly noticeable improvement as we collect more source data, a trend clearly reflected in the plot. Next, we move on to explore how the degree of covariate shift affects the effectiveness of transfer learning. Figure~\ref{fig:reg_P}~(b) depicts the regret vs.~the transfer exponent~$\gamma$. It can be seen from the plot that the regret grows as the transfer exponent $\gamma$ increases (which means that $P_{X}$ and $Q_{X}$ are more distinct). Hence, this observation confirms our theoretical guarantees regarding the impact of the transfer exponent. Moreover, let us examine the influence of the exploration coefficient $\kappa$. Figure~\ref{fig:reg_P}~(c) plots the regret vs.~the probability that arm $1$ is pulled in the source data. As predicted by our theories, we can see that the regret curve exhibits a U shape with rough mirror symmetry. In particular, it reaches its minimum when $\mu(1) = 1/2$, corresponding to the case $\kappa = 1$.

\begin{figure}[t]
\centering
\begin{tabular}{ccc}
\includegraphics[width=0.31\textwidth]{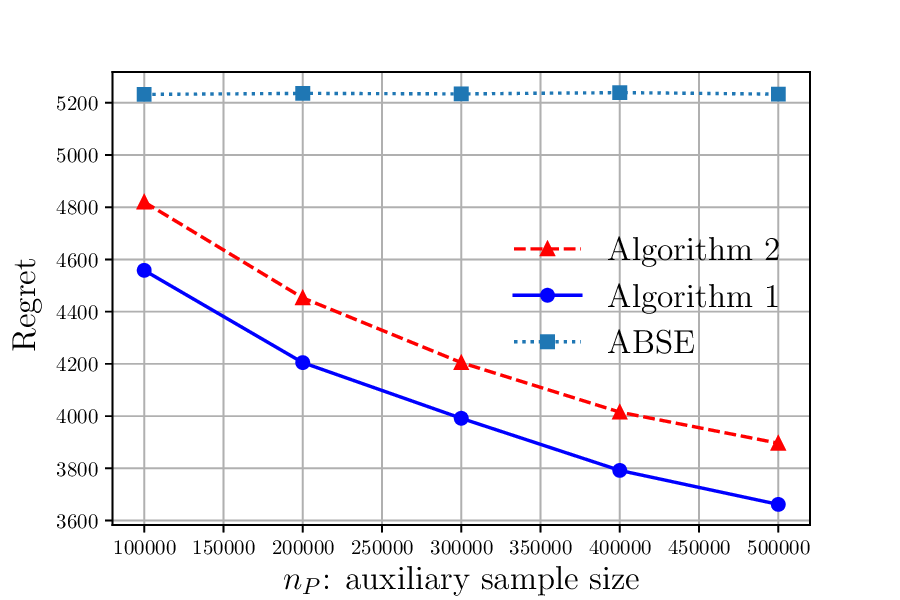} 
& \includegraphics[width=0.31\textwidth]{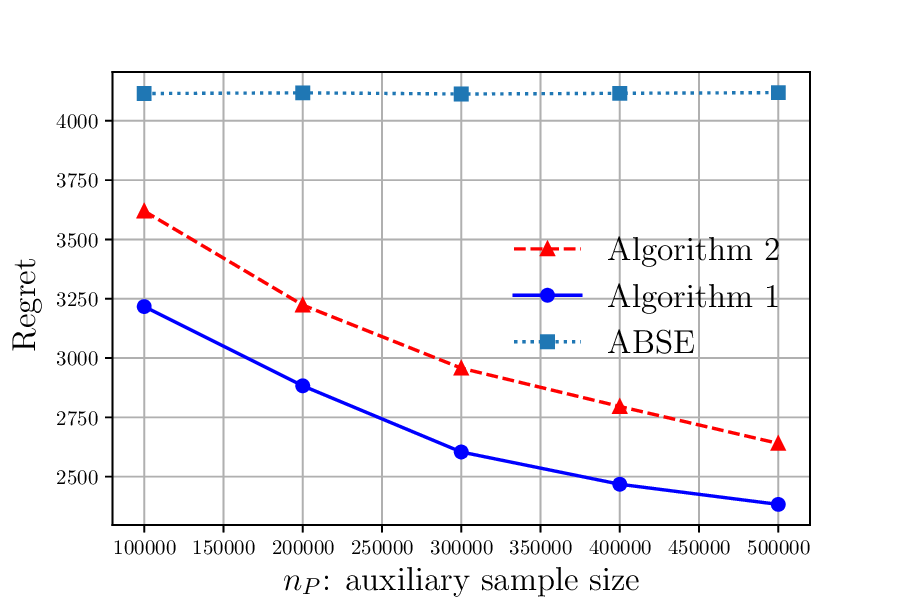} 
& \includegraphics[width=0.31\textwidth]{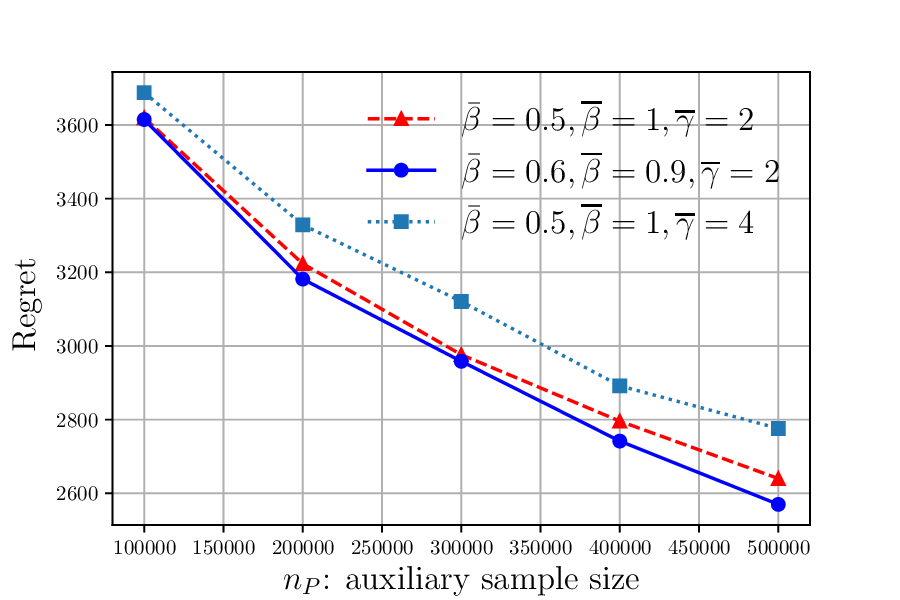} 
\tabularnewline
(a) & (b) 
& (c) \tabularnewline
\end{tabular}
\caption{(a) Regret vs.~auxiliary sample size $n_P$ with $\beta = 0.6$; (b) Regret vs.~auxiliary sample size $n_P$ with $\beta = 0.8$; (c) Regret of Algorithm~\ref{alg:UCB-TL-adaptive} vs.~auxiliary sample size $n_P$ for different parameter bounds with $\beta = 0.8$. Here, $d=2, K=4,  \gamma = 1$,  $\kappa = 1$, and $n_Q = 1 \times 10^{5}$. \label{fig:adapt}}
\end{figure}

\paragraph*{Adaptive algorithm}
Lastly, let us perform a comparison of the numerical performance among three algorithms: the proposed adaptive algorithm (Algorithm~\ref{alg:UCB-TL-adaptive}), the minimax optimal algorithm (Algorithm~\ref{alg:UCB-TL}), and the ABSE algorithm. Assuming that both Algorithm~\ref{alg:UCB-TL} and the ABSE algorithm possess complete knowledge of the parameters $\beta$ and $\gamma$, Figure~\ref{fig:adapt} (a) and (b) illustrate the regret vs.~the source sample size $n_P$ for $\beta = 0.6$ and $\beta = 0.8$, respectively. Here, the parameter bounds for the adaptive algorithm are set as $\underline{\beta} = 0.5, \overline{\beta} = 1, \overline{\gamma} = 2$. In addition, Figure~\ref{fig:adapt} (c) plots the regret of Algorithm~\ref{alg:UCB-TL-adaptive} vs.~the source sample size $n_P$ for different choices of parameter bounds for $\beta$ and $\gamma$. Encouragingly, Figure~\ref{fig:adapt} demonstrates that the regrets of Algorithm~\ref{alg:UCB-TL-adaptive} are reasonably close to those of Algorithm~\ref{alg:UCB-TL} for different values of $\beta$ and that they remain relatively insensitive to the parameter bounds, as long as they contain the true values.  This confirms the practical applicability of our adaptive procedure in real-world scenarios.

\section{Proof of Theorem \ref{thm:upper-bound}}
\label{sec:Proof-of-Upper-Bound}

Inspired by the analytical framework developed in \citet{perchet2013multi}, let us first introduce several notations. 
To begin with, let $\mathcal{T}_{t}$ denote the partition tree at time $t$ for any $t\geq 1$. Next, for any bin $B$, we define the parent of $B$ by
\begin{align}  
\mathsf{p} (B) \defn
\begin{cases} 
\{B': B \in \mathsf{child}(B') \}, & \text{if } B \neq \mathcal{X}, \\
 \varnothing, & \text{if } B = \mathcal{X}.
\end{cases} 
\end{align}
Here, we recall that $\mathsf{child}(B)$ denotes the set of children of any bin $B$, namely, $\mathsf{child}(B) \defn \{ B' \in \mathcal{B}_{i + 1} : B' \subset B \}$ for any $B \in \mathcal{B}_{i}$ with $i \geq 0$. Denote $\mathsf{p}^1(B) \defn \mathsf{p}(B)$ and $\mathsf{p}^k(B) \defn \mathsf{p}(\mathsf{p}^{k-1}(B))$. The set of ancestors of any bin $B$ is defined by
\begin{align}  
\mathsf{anc}(B) \coloneqq \{B' \in \mathcal{T} : B' = \mathsf{p}^k (B) \text{ for some } k \geq 1\}.
\end{align}
In addition, since we use the complete source dataset $\mathcal{D}^{P}$ in Algorithm~\ref{alg:UCB-TL}, we drop the notation $\mathcal{D}^{P}$ in $n_{k}^{P}(B;\mathcal{D}^{P})$, $\overline{Y}_{k}^{P}(B;\mathcal{D}^{P})$ and $\tau_{k}^{\star}(B;\mathcal{D}^{P})$ throughout the proof of Theorem~\ref{thm:upper-bound} in Appendix~\ref{sec:Proof-of-Upper-Bound}.

With the notations in hand, we begin the proof of Theorem \ref{thm:upper-bound}.
For any bin $B$, let us denote
\begin{align}
R^{\tree}(B) & \defn\sum_{t=1}^{n_{Q}}\big(f_{\pi^{\star}(X_{t}^{Q})}(X_{t}^{Q})-f_{\pi_{t}(X_{t}^{Q})}(X_{t}^{Q})\big)\mathds{1}\{X_{t}^{Q}\in B\} \mathds{1}\{B\in\mathcal{T}_{t}\},\label{eq:R-nonleaf-B} \\
R^{\lf}(B) & \defn\sum_{t=1}^{n_{Q}}\big(f_{\pi^{\star}(X_{t}^{Q})}(X_{t}^{Q})-f_{\pi_{t}(X_{t}^{Q})}(X_{t}^{Q})\big)\mathds{1}\{X_{t}^{Q}\in B\} \mathds{1}\{B\in\mathcal{L}_{t}\},\label{eq:R-leaf-B}
\end{align}
where we recall that $\mathcal{L}_{t}$ is the set of the leaf nodes of the tree $\mathcal{T}_{t}$ at time $t$, in particular, the partition of the covariate space $\mathcal{X}$ at time $t$. 
Note that $R_{n_{Q}}(\pi)=R^{\tree}(\mathcal{X})$ since $X_{t}^{Q}\in\mathcal{X}\in\mathcal{T}_{t}$ for any $t\geq1$. Therefore, we shall control $R^{\tree}(\mathcal{X})$ in what follows.

For any bin $B$, one can decompose $R^{\tree}(B)$ as
\begin{align}
R^{\tree}(B)=R^{\lf}(B)+\sum_{B'\in\mathsf{child}(B)}R^{\tree}(B').\label{eq:R-leaf-nonleaf-relation}
\end{align}
In the bin $B$, we shall focus on the arms whose reward gaps relative to the optimal arm are of order $|B|^{\beta}$. To precisely quantify this, for each bin $B$, let us define two sets of arms:
\begin{align}
\underline{\mathcal{I}}_{B} & \defn\big\{ k\in[K]:\sup_{x\in B}\{f_{(1)}(x)-f_{k}(x)\}\leq c_{0}|B|^{\beta}\big\}, \label{eq:arm_set_lb}\\
\overline{\mathcal{I}}_{B} & \defn\big\{ k\in[K]:\sup_{x\in B}\{f_{(1)}(x)-f_{k}(x)\}\leq8c_{0}|B|^{\beta}\big\}, \label{eq:arm_set_ub}
\end{align}
where we define $c_{0}\defn2C_{\beta}$. In addition, let $\mathcal{I}_{B}$ denote the set of active arms in bin $B$ at the end of rounds before $B$ is replaced by its children in the partition. Now we define the ``good'' event 
\begin{align}
\label{eq:good-event-B}
\mathcal{E}_{B}\defn\{\underline{\mathcal{I}}_{B}\subset\mathcal{I}_{B}\subset\overline{\mathcal{I}}_{B}\}.
\end{align}
By (\ref{eq:R-leaf-nonleaf-relation}), we can decompose
\begin{align*}
R^{\tree}(B) & = R^{\tree}(B) \mathds{1}\{\mathcal{E}_B^c \} + R^{\tree}(B) \mathds{1}\{\mathcal{E}_B \} \\
& = R^{\tree}(B) \mathds{1}\{\mathcal{E}_B^c \} + \bigg( R^{\lf}(B)+\sum_{B'\in\mathsf{child}(B)}R^{\tree}(B') \bigg) \mathds{1}\{\mathcal{E}_B \} \\ 
& = R^{\tree}(B)\mathds{1}\{\mathcal{E}_{B}^c\}+R^{\lf}(B)\mathds{1}\{\mathcal{E}_{B}\}+\sum_{B'\in\mathsf{child}(B)}R^{\tree}(B')\mathds{1}\{\mathcal{E}_{B}\}\\
 & =R^{\tree}(B)\mathds{1}\{\mathcal{E}_{B}^c\}+R^{\lf}(B)\mathds{1}\{\mathcal{E}_{B}\}\\
 & \quad+\sum_{B'\in\mathsf{child}(B)}\bigg(R^{\tree}(B')\mathds{1}\{\mathcal{E}_{B'}^{c}\}+R^{\lf}(B')\mathds{1}\{\mathcal{E}_{B'}\}+\sum_{B''\in\mathsf{child}(B')}R^{\tree}(B'')\mathds{1}\{\mathcal{E}_{B'}\}\bigg)\mathds{1}\{\mathcal{E}_{B}\}\\
 & =R^{\tree}(B)\mathds{1}\{\mathcal{E}_{B}^c\}+\sum_{B'\in\mathsf{child}(B)}R^{\tree}(B')\mathds{1}\{\mathcal{E}_{B'}^{c}\cap\mathcal{E}_{B}\} \\
 & \quad+R^{\lf}(B)\mathds{1}\{\mathcal{E}_{B}\}+\sum_{B'\in\mathsf{child}(B)}R^{\lf}(B')\mathds{1}\{\mathcal{E}_{B'}\cap\mathcal{E}_{B}\}\\
 & \quad+\sum_{B'\in\mathsf{child}(B)}\sum_{B''\in\mathsf{child}(B')}R^{\tree}(B'')\mathds{1}\{\mathcal{E}_{B'}\cap\mathcal{E}_{B}\}.
\end{align*}
Let us define 
\begin{align}
\label{eq:good-event-ancB}
    \mathcal{G}_{B}\defn\bigcap_{B'\in\mathsf{anc}(B)}\mathcal{E}_{B'},
\end{align}
and adopt the convention $\mathds{1}\{\mathcal{G}_\mathcal{X}\}=1$.
Then, by deduction, one can express the regret by
\begin{align}
R^{\tree}(\mathcal{X}) &=  \sum_{0\leq i< l^{\star}}\underbrace{\sum_{B\in\mathcal{B}_{i}}R^{\tree}(B)\mathds{1}\{\mathcal{E}_{B}^{c}\cap\mathcal{G}_{B}\}}_{=:\,R^{\tree}_{i}}
+\sum_{0\leq i< l^{\star}}\underbrace{\sum_{B\in\mathcal{B}_{i}}R^{\lf}(B)\mathds{1}\{\mathcal{E}_{B}\cap\mathcal{G}_{B}\}}_{=:\,R^{\lf}_{i}} \nonumber \\
&\quad +\underbrace{\sum_{B\in\mathcal{B}_{l^{\star}}}R^{\tree}(B)\mathds{1}\{\mathcal{G}_{B}\}}_{=:\,R_{l^{\star}}}. \label{eq:regret-oracle-decomp}
\end{align}
Here, $l^{\star}\in\mathbb{N}$ is chosen to be
\begin{align}
\label{eq:max_depth}
    l^{\star} \defn \bigg\lceil {\frac{1}{d+2\beta}}\log_{2}\Big(\frac{ n_{Q}}{K}\Big) \bigg\rceil \vee \bigg\lceil {\frac{1}{d+2\beta+\gamma}\log_{2} \Big(\frac{c^{\star}\kappa n_{P}}{K}\Big)}\bigg\rceil,
\end{align} 
where $c^{\star}\defn \frac{1}{4}\underline{q}c_{\gamma}C_{\beta}^2$. In addition, let us define $l_{4} \coloneqq \big\lceil
\frac{1}{d+\gamma} \log_{2} \big( \frac{\underline{q}c_{\gamma}\kappa n_{P}}{80K\log(n_{P})} \big)\big\rceil$, and $l_{5}\defn\big\lceil \frac{1}{d+2\beta+\gamma}\log(\frac{c^{\star}\kappa n_{P}}{K})\big\rceil$.

In what follows, we shall control $R^{\tree}_{i}$, $R^{\lf}_{i}$, and $R_{l^{\star}}$ separately, which are established in Lemmas~\ref{lemma:non-leaf-upper-oracle-born}--\ref{lemma:leaf-upper-oracle}.
To begin with, Lemma \ref{lemma:non-leaf-upper-oracle-born} (below) provides the upper bounds for $R^{\tree}_{i}$.
\begin{lemma}\label{lemma:non-leaf-upper-oracle-born} 
For any $i\geq0$, one has
\begin{align}
\label{eq:regret-non-leaf-oracle-born-i-large}
 \mathbb{E}[R^{\tree}_{i}] \lesssim   K2^{(d+2\beta-(1+\alpha)\beta )i}\log^{+}\big(n_{Q}2^{-(d+2\beta)i}\big).
\end{align}
Moreover, for any $0\leq i< l_{4} \wedge l_{5}$, one further has
\begin{align}
\mathbb{E}[R^{\tree}_{i}] \lesssim
& Kn_{Q}2^{-(1+\alpha)\beta i} \exp\left(-\frac{c^{\star}\kappa}{K}n_{P}2^{-(d+2\beta+\gamma)i}\right)  
\nonumber \\
& + Kn_{Q}2^{-(1+\alpha)\beta i} \bigg(\frac{1}{\kappa n_{P}}2^{(d+2\beta+\gamma) i} \wedge \frac{1}{n_{Q}} 2^{(d+2\beta)i} + \frac{1}{n^{10}_{P}}\bigg).
% & + Kn_{Q}2^{-(1+\alpha)\beta i} \frac{1}{\kappa n_{P}}2^{(d+2\beta+\gamma) i}.
 \label{eq:regret-non-leaf-oracle-born-i-small}
\end{align}
\end{lemma}
\begin{proof}
    See Appendix \ref{subsec:Proof-of-lemma:non-leaf-upper-oracle-born}.
\end{proof}
Next, Lemma \ref{lemma:non-leaf-upper-oracle-live} (below) gives the upper bounds for $R^{\lf}_{i}$.
\begin{lemma}\label{lemma:non-leaf-upper-oracle-live}
For any $i\geq0$, we have
\begin{align}
\label{eq:regret-non-leaf-oracle-live-i-large}
 \mathbb{E}[R^{\lf}_{i}] \lesssim   K2^{(d+2\beta-(1+\alpha)\beta )i}\log^{+}\big(n_{Q}2^{-(d+2\beta)i}\big).
\end{align}
Furthermore, the following holds for any $0\leq i< l_{4}\wedge l_{5}$:
\begin{align}
\mathbb{E}[R^{\lf}_{i}]
&\lesssim Kn_{Q}2^{-(1+\alpha)\beta i}\exp\left(-\frac{c^{\star}\kappa}{K}n_{P}2^{-(d+2\beta+\gamma)i}\right) \nonumber \\
&+ \frac{K}{n^{10}_P}2^{(d+2\beta-(1+\alpha)\beta)i}\log^{+}\big(n_{Q}2^{-(d+2\beta)i}\big). \label{eq:regret-non-leaf-oracle-live-i-small}
\end{align}
\end{lemma}
\begin{proof}
    See Appendix \ref{sec:Proof-of-lemma:non-leaf-upper-oracle-live}.
\end{proof}
In addition, Lemma \ref{lemma:leaf-upper-oracle} (below) controls the regret accumulated on the nodes of depth~$l^{\star}$.
\begin{lemma}\label{lemma:leaf-upper-oracle}The regret incurred
on the nodes of depth $l^{\star}$ obeys
\begin{align}
\mathbb{E}[R_{ l^{\star}}]\lesssim n_{Q}2^{- (1+\alpha)\beta  l^{\star}}.\label{eq:regret-leaf-oracle}
\end{align}
\end{lemma}
\begin{proof}
    See Appendix \ref{subsec:Proof-of-lemma:leaf-upper-oracle}.
\end{proof}

With Lemmas~\ref{lemma:non-leaf-upper-oracle-born}--\ref{lemma:leaf-upper-oracle} in hand, we can readily upper bound the regret.
Combining (\ref{eq:regret-oracle-decomp}), (\ref{eq:regret-non-leaf-oracle-born-i-large}), (\ref{eq:regret-non-leaf-oracle-live-i-large}), and (\ref{eq:regret-leaf-oracle}) yields the following bound:
\begin{align}
\mathbb{E}\big[R_{n_{Q}}(\pi)\big] &= \mathbb{E} \big[R^{\tree}(\mathcal{X})\big] = \sum_{0\leq i< l^{\star}}\mathbb{E}[R^{\tree}_{i}] +\sum_{0\leq i< l^{\star}}\mathbb{E}[R^{\lf}_{i}] + \mathbb{E}[R_{ l^{\star}}] \nonumber\\
 & \lesssim \sum_{0 \leq i< l^{\star}}K2^{(d+2\beta-(1+\alpha)\beta )i}\log^{+}\big(n_{Q}2^{-(d+2\beta)i}\big) + n_{Q}2^{- (1+\alpha)\beta  l^{\star}}. \label{eq:regret-oracle-ub-crude-temp}
\end{align} 
Meanwhile, we can also plug (\ref{eq:regret-non-leaf-oracle-born-i-large})--(\ref{eq:regret-leaf-oracle}) into (\ref{eq:regret-oracle-decomp}) to obtain a more refined upper bound:
\begin{align}
\mathbb{E}\big[R_{n_{Q}}(\pi)\big] &= \mathbb{E} \big[R^{\tree}(\mathcal{X})\big]  =\sum_{0\leq i< l^{\star}}\mathbb{E}[R^{\tree}_{i}] +\sum_{0\leq i< l^{\star}}\mathbb{E}[R^{\lf}_{i}] + \mathbb{E}[R_{ l^{\star}}] \nonumber\\
 & \lesssim \sum_{0\leq i< l_{4}\wedge l_{5} \wedge l^{\star}} Kn_{Q}2^{-(1+\alpha)\beta i} \exp\left(-\frac{c^{\star}\kappa}{K}n_{P}2^{-(d+2\beta+\gamma)i}\right) \nonumber \\
 & \quad+\sum_{0\leq i< l_{4}\wedge l_{5} \wedge l^{\star}} Kn_{Q}2^{-(1+\alpha)\beta i} \bigg(\frac{1}{\kappa n_{P}}2^{(d+2\beta+\gamma) i} \wedge \frac{1}{n_{Q}}2^{(d+2\beta)i}+\frac{1}{n^{10}_{P}}\bigg) \nonumber\\
  % & \quad+\sum_{0\leq i< l_{4}\wedge l_{5} \wedge l^{\star}} Kn_{Q}2^{-(1+\alpha)\beta i} \frac{1}{\kappa n_{P}}2^{(d+2\beta+\gamma) i} \nonumber\\
 & \quad + \sum_{0\leq i< l_{4}\wedge l_{5} \wedge l^{\star}}\frac{1}{n^{10}_P}K2^{(d+2\beta-(1+\alpha)\beta)i}\log^{+}\big(n_{Q}2^{-(d+2\beta)i}\big) \nonumber \\
 & \quad + \sum_{l_{4}\wedge l_{5} \wedge l^{\star} \leq i< l^{\star}}K2^{(d+2\beta-(1+\alpha)\beta )i}\log^{+}\big(n_{Q}2^{-(d+2\beta)i}\big) + n_{Q}2^{- (1+\alpha)\beta  l^{\star}}. \label{eq:regret-oracle-ub-temp}
\end{align}
In what follows, we shall exploit \eqref{eq:regret-oracle-ub-crude-temp} and \eqref{eq:regret-oracle-ub-temp} to bound the regret differently according to the relationship between $n_{P}$ and $n_{Q}$.

\begin{itemize}
\item 
We begin with the scenario $\frac{n_Q}{K} \geq (\frac{c^{\star}\kappa n_{P}}{K})^{\frac{d+2\beta}{d+2\beta+\gamma}}$, where one has $l^{\star}=\big\lceil \frac{1}{d+2\beta} \log_{2}(\frac{n_{Q}}{K})\big\rceil$ by (\ref{eq:max_depth}), and we shall invoke (\ref{eq:regret-oracle-ub-crude-temp}) to control the regret. 
In this case, by the conditions that $\alpha\beta \leq d$ and $l^{\star}=\big\lceil \frac{1}{d+2\beta} \log_{2}(\frac{n_{Q}}{K})\big\rceil$, applying Lemma~\ref{lemma:series-1} yields
\begin{align}
\label{eq:sum-temp1}
    \sum_{0 \leq i< l^{\star}} & 2^{(d+2\beta- (1+\alpha)\beta )i}\log^{+}\big(n_{Q}2^{-(d+2\beta)i}\big) \lesssim 2^{(d + 2\beta - (1+\alpha)\beta )l^{\star}}.
\end{align}
As a consequence, collecting (\ref{eq:regret-oracle-ub-crude-temp}) and \eqref{eq:sum-temp1} together yields
\begin{align*}
\mathbb{E}\big[R_{n_{Q}}(\pi)\big] & \lesssim K 2^{(d + 2\beta - (1+\alpha)\beta )l^{\star}} + n_{Q}2^{-(1+\alpha)\beta  l^{\star}} \\
& \lesssim n_{Q} \Big(\frac{n_{Q}}{K}\Big)^{-\frac{(1+\alpha)\beta }{d+2\beta}} \asymp n_{Q}^{1-\frac{(1+\alpha)\beta }{d+2\beta}}.
\end{align*}
where the last line follows from $l^{\star}=\big\lceil \frac{1}{d+2\beta} \log_{2} (\frac{n_{Q}}{K})\big\rceil$ and $K \asymp 1$.

\item 
Next, let us turn to the case $\frac{n_Q}{K}< (\frac{c^{\star}\kappa n_{P}}{K})^{\frac{d+2\beta}{d+2\beta+\gamma}}$, where we have $l^{\star} = \big\lceil \frac{1}{d+2\beta+\gamma}\log_{2}(\frac{c^{\star}\kappa n_{P}}{K})\big\rceil$ from (\ref{eq:max_depth}), and we shall apply (\ref{eq:regret-oracle-ub-temp}) to bound the regret. As a reminder, we define $l_{4} \coloneqq \big\lceil
\frac{1}{d+\gamma} \log_{2} \big( \frac{\underline{q}c_{\gamma}\kappa n_{P}}{80K\log(n_{P})} \big)\big\rceil$ and $l_{5}\defn\big\lceil \frac{1}{d+2\beta+\gamma}\log_{2}(\frac{c^{\star}\kappa n_{P}}{K})\big\rceil$.
Since $\beta > 0$, one also has $l^{\star} = l_{5} \leq l_{4}$ for sufficiently large $n_{P}$. Therefore, we can invoke (\ref{eq:regret-oracle-ub-temp}) to obtain
\begin{align*}
\mathbb{E}\big[R_{n_{Q}}(\pi)\big]
 & \lesssim \sum_{0\leq i< l^{\star}} Kn_{Q}2^{-(1+\alpha)\beta i} \exp\left(-\frac{c^{\star}\kappa}{K}n_{P}2^{-(d+2\beta+\gamma)i}\right)  \\
 & \quad+\sum_{0\leq i< l^{\star}} Kn_{Q}2^{-(1+\alpha)\beta i} \bigg(\frac{1}{\kappa n_{P}}2^{(d+2\beta+\gamma) i} \wedge \frac{1}{n_{Q}}2^{(d+2\beta)i}+\frac{1}{n^{10}_{P}}\bigg) \\
 & \quad + \frac{1}{n^{10}_P}\sum_{0\leq i< l^{\star}}K2^{(d+2\beta-(1+\alpha)\beta)i}\log^{+}\big(n_{Q}2^{-(d+2\beta)i}\big) + n_{Q}2^{- (1+\alpha)\beta  l^{\star}}  \\
 & \overset{(\mathrm{i})}{\lesssim} K n_{Q} 2^{-(1+\alpha)\beta l^{\star}}
  + K 2^{-(1+\alpha)\beta l^{\star}} \bigg(\frac{n_{Q}}{\kappa n_{P}}2^{(d+2\beta+\gamma) l^{\star}} \wedge 2^{(d+2\beta)l^{\star}}+\frac{1}{n^{10}_{P}}\bigg) \\
 & \quad + K 2^{-(1+\alpha)\beta l^{\star}} \frac{1}{n_{P}^{10}} 2^{(d+2\beta)l^{\star}} \\
  % \log^{+}\big(n_{Q}2^{-(d+2\beta)l^{\star}}\big)  \\
 & \asymp K \Big(n_{Q}+  \frac{ n_{Q}}{\kappa n_{P}} 2^{(d+2\beta+\gamma)l^{\star}} \wedge  2^{(d+2\beta) l^{\star}} \Big)2^{- (1+\alpha)\beta  l^{\star}}\\
 & \overset{(\mathrm{ii})}{\lesssim} K\Big(n_{Q}+ \frac{n_{Q}}{\kappa n_{P}} \frac{c^{\star}\kappa n_{P}}{K} \wedge \Big(\frac{c^{\star}\kappa n_{P}}{K}\Big)^{\frac{d+2\beta}{d+2\beta+\gamma}} \Big) \Big(\frac{c^{\star}\kappa n_{P}}{K}\Big)^{-\frac{ (1+\alpha)\beta }{d+2\beta+\gamma}} \\
 & \overset{(\mathrm{iii})}{\asymp} n_{Q} (\kappa n_{P})^{-\frac{ (1+\alpha)\beta }{d+2\beta+\gamma}}.
\end{align*}
Here, (i) uses \eqref{eq:sum-temp1}, Lemma~\ref{lemma:series-2}, and $l^{\star} = \big\lceil \frac{1}{d+2\beta+\gamma}\log_{2}(\frac{c^{\star}\kappa n_{P}}{K})\big\rceil$; (ii) uses use $l^{\star} = \big\lceil \frac{1}{d+2\beta+\gamma}\log_{2}(\frac{c^{\star}\kappa n_{P}}{K})\big\rceil$;
(iii) is due to $\frac{n_Q}{K}< (\frac{c^{\star}\kappa n_{P}}{K})^{\frac{d+2\beta}{d+2\beta+\gamma}}$ and $K\asymp 1$.

\item
Combining the two cases above, we arrive at the advertised bound
\begin{align*}
\mathbb{E}\big[R_{n_{Q}}(\pi)\big]\lesssim n_{Q}\big( n_{Q} + ( \kappa n_{P})^{\frac{d+2\beta}{d+2\beta+\gamma}} \big)^{-\frac{ (1+\alpha)\beta }{d+2\beta}}.
\end{align*}
This finishes the proof of Theorem~\ref{thm:upper-bound}.

\end{itemize}

\subsection{Proof of Lemma \ref{lemma:non-leaf-upper-oracle-born}}

\label{subsec:Proof-of-lemma:non-leaf-upper-oracle-born}
Fix an arbitrary bin $B \in \mathcal{B}_{i}$ such that $0 \leq i < l^{\star}$ and $Q_{X}(B) > 0$.
Recall that Procedure~\ref{alg:EA-TL} in Algorithm~\ref{alg:UCB-TL} operated in bin $B$ is restricted to observations of which the covariates fall in bin $B$. For any arm $k\in[K]$, denote by $\big(Y^{Q,(k)}_{i}(B)\big)_{i \geq 1}$ and $\big(Y^{P,(k)}_{i}(B)\big)_{i \geq 1}$ the associated random rewards observed by successive pulls of arm $k$ in the $Q$-bandit and $P$-data, respectively. The average reward of arm $k$ that combines the observed rewards of $\tau$ pulls in the $Q$-bandit and the samples in the $P$-data is denoted by
\begin{align}
    \overline{Y}_{\tau}^{(k)}(B) \coloneqq \frac{1}{\tau + n_{k}^{P}(B)} \Bigg( \sum_{i=1}^{\tau} Y^{Q,(k)}_{i}(B) + \sum_{i=1}^{n_{k}^{P}(B)} Y^{P,(k)}_{i}(B) \Bigg).
\end{align}

On the event $\mathcal{G}_{B}$, we know that $\mathcal{I}_{\mathsf{p}(B)}\subset\overline{\mathcal{I}}_{\mathsf{p}(B)}$.
By the definition of $\overline{\mathcal{I}}_{\mathsf{p}(B)}$ in \eqref{eq:arm_set_ub}, for any arm $k\in\mathcal{I}_{\mathsf{p}(B)}$, one has
\begin{align*}
\sup_{x\in\mathsf{p}(B)}\{f_{(1)}(x)-f_{k}(x)\}\leq8c_{0}|\mathsf{p}(B)|^{\beta}=2^{3+\beta}c_{0}|B|^{\beta} =: c_{1} |B|^{\beta}.
\end{align*}
This implies that for any arm $k\in\mathcal{I}_{\mathsf{p}(B)}$ and $x\in B$,
\begin{align}
\big(f_{(1)}(x)-f_{k}(x)\big)\mathds{1}\{\mathcal{G}_{B}\}\leq c_{1}|B|^{\beta}\mathds{1}\{0<f_{(1)}(x)-f_{(2)}(x)\leq c_{1}|B|^{\beta}\}\mathds{1}\{\mathcal{G}_{B}\}.\label{eq:reward-gap-event-ancestor}
\end{align}
Therefore, we can bound
\begin{align*}
& \mathbb{E}[R^{\tree}(B)\mathds{1}\{\mathcal{E}_{B}^{c}\cap\mathcal{G}_{B}\}]\\
& \quad=\mathbb{E}\bigg[\mathds{1}\{\mathcal{E}_{B}^{c}\cap\mathcal{G}_{B}\}\sum_{t=1}^{n_{Q}}\big(f_{\pi^{\star}(X_{t}^{Q})}(X_{t}^{Q})-f_{\pi_{t}(X_{t}^{Q})}(X_{t}^{Q})\big)\mathds{1}\{X_{t}^{Q}\in B\}\mathds{1}\{B\in\mathcal{T}_{t}\}\bigg]\\
& \quad\leq\mathbb{E}\bigg[\mathds{1}\{\mathcal{E}_{B}^{c}\cap\mathcal{G}_{B}\}\sum_{t=1}^{n_{Q}} c_{1}|B|^{\beta}\mathds{1}\{0<f_{(1)}(X_{t}^{Q})-f_{(2)}(X_{t}^{Q})\leq c_{1}|B|^{\beta}\}\mathds{1}\{X_{t}^{Q}\in B\}\bigg]\\
& \quad \lesssim n_{Q}|B|^{\beta}Q_{X}(0<f_{(1)}(X)-f_{(2)}(X)\leq c_{1}|B|^{\beta},X\in B)\mathbb{P}\{\mathcal{E}_{B}^{c}\cap\mathcal{G}_{B}\}.
\end{align*}
% where the last line holds because $Q_{X}(B)\leq\overline{q}|B|^{d}$ due to Assumption \ref{assumption:bounded-density}. 
Summing over all $B\in\mathcal{B}_{i}$ yields
\begin{align}
\mathbb{E}[R^{\tree}_{i}] & =\sum_{B\in\mathcal{B}_{i}}\mathbb{E}[R^{\tree}(B)\mathds{1}\{\mathcal{E}_{B}^{c}\cap\mathcal{G}_{B}\}] \nonumber\\
 & \lesssim n_{Q}2^{-\beta i}\max_{B\in\mathcal{B}_{i}}\mathbb{P}\{\mathcal{E}_{B}^{c}\cap\mathcal{G}_{B}\}\sum_{B\in\mathcal{B}_{i}}Q_{X}(0<f_{(1)}(X)-f_{(2)}(X)\leq c_{1}|B|^{\beta},X\in B).\label{eq:regret-nonleaf-good-event-temp1}
\end{align}
By the margin condition (Assumption \ref{assumption:margin}), we can upper bound
\begin{align}
\sum_{B\in\mathcal{B}_{i}}  Q_{X}(0<f_{(1)}(X)-f_{(2)}(X)\leq c_{1}|B|^{\beta},X\in B)
% & =\sum_{B\in\mathcal{B}_{i}}\frac{1}{Q_{X}(B)}Q_{X}(0<f_{(1)}(X)-f_{(2)}(X)\leq c_{1}|B|^{\beta},X\in B)\nonumber \\
 & = Q_{X}(0<f_{(1)}(X)-f_{(2)}(X)\leq c_{1}|B|^{\beta})\nonumber \\
 & \leq C_{\alpha }( c_{1}|B|^{\beta})^{\alpha} \nonumber \\
 & \lesssim |B|^{\alpha\beta} = 2^{-\alpha\beta i}.\label{eq:prob-Q-gap-ub}
\end{align}
Combining (\ref{eq:regret-nonleaf-good-event-temp1}) and (\ref{eq:prob-Q-gap-ub}) shows that
\begin{align}
\mathbb{E}[R^{\tree}_{i}]\lesssim n_{Q}2^{-(1+\alpha)\beta  i}\max_{B\in\mathcal{B}_{i}}\mathbb{P}\{\mathcal{E}_{B}^{c}\cap\mathcal{G}_{B}\}.\label{eq:regret-nonleaf-good-event-temp}
\end{align}

In addition, we claim that for any $i \geq 0$,
\begin{align}
\max_{B\in\mathcal{B}_{i}}\mathbb{P}\{\mathcal{E}_{B}^{c}\cap\mathcal{G}_{B}\}\lesssim
\frac{K}{n_{Q}}2^{(d+2\beta)i}\log^{+}\big(n_{Q}2^{-(d+2\beta)i}\big), \label{eq:prob-Ec-F-i-large}
\end{align}
and for any $0\leq i<l_{4}\wedge l_{5}$,
\begin{align}
\max_{B\in\mathcal{B}_{i}}\mathbb{P}\{\mathcal{E}_{B}^{c}\cap\mathcal{G}_{B}\}\lesssim
K\exp\left(-\frac{c^{\star}\kappa}{K}n_{P}2^{-(d+2\beta+\gamma)i}\right) + \frac{K2^{(d+2\beta+\gamma)i}}{\kappa n_{P}} 
\wedge \frac{K2^{(d+2\beta)i}}{n_{Q}} +\frac{1}{n^{10}_{P}}
; \label{eq:prob-Ec-F-i-small}
\end{align}
with the proof postponed to the end of the section. Combining (\ref{eq:regret-nonleaf-good-event-temp}), (\ref{eq:prob-Ec-F-i-large}), and (\ref{eq:prob-Ec-F-i-small}) leads to the advertised bounds (\ref{eq:regret-non-leaf-oracle-born-i-large}) and (\ref{eq:regret-non-leaf-oracle-born-i-small}).

Therefore, the remainder of the proof boils down to establishing (\ref{eq:prob-Ec-F-i-large}) and (\ref{eq:prob-Ec-F-i-small}).
By definition of $\mathcal{E}_{B}$, we can express
\begin{align*}
\mathbb{P}\{\mathcal{E}_{B}^{c}\cap\mathcal{G}_{B}\} & =\mathbb{P}\big\{\{\underline{\mathcal{I}}_{B}\not\subset\mathcal{I}_{B}\}\cap\mathcal{G}_{B}\big\}+\mathbb{P}\big\{\{\underline{\mathcal{I}}_{B}\subset\mathcal{I}_{B}\}\cap\{\mathcal{I}_{B}\not\subset\overline{\mathcal{I}}_{B}\}\cap\mathcal{G}_{B}\big\}.
\end{align*}
In what follows, we shall control the terms on the right-hand side separately.
\begin{itemize}
\item We start with the first term. On the event $\{ \underline{\mathcal{I}}_{B}\not\subset\mathcal{I}_{B} \}$, we know that there exists an arm $k_{0}\in\underline{\mathcal{I}}_{B}$ that gets eliminated before $\tau_{k_{0}}^{\star}(B)$ rounds in bin $B$. By the elimination criterion in Algorithm~\ref{alg:EA-TL}, we know that there exists some $j_{0}\in\mathcal{I}_{\mathsf{p}(B)}$ and $\tau_{k_0},\tau_{j_0}\geq0$ such that $\overline{Y}_{\tau_{j_0}}^{(j_{0})}(B)-U_{j_{0}}(\tau_{j_0},B)>\overline{Y}_{\tau_{k_0}}^{(k_{0})}(B)+U_{k_{0}}(\tau_{k_0},B)$.
On the other hand, as arm $k_{0}\in\underline{\mathcal{I}}_{B}$, one knows
\begin{align*}
    \bar{f}^{Q}_{j_{0}}(B)-\bar{f}^{Q}_{k_{0}}(B) & \leq \bar{f}^{Q}_{(1)}(B)-\bar{f}^{Q}_{k_{0}}(B) \\
    & = \bE \big[ f_{(1)}(X^{Q})-f_{k_{0}}(X^{Q}) \,|\,X^{Q}\in B\big] \\
    & \leq c_{0}|B|^\beta \leq\frac{1}{2}U_{k_{0}}(\tau_{k_{0}},B)+\frac{1}{2}U_{j_{0}}(\tau_{j_{0}},B),
\end{align*}
where the first inequality follows from the definition of $\underline{\mathcal{I}}_{B}$ in (\ref{eq:arm_set_lb}), and the last step holds due to the definition of $U_{k}^{\star}(\tau,B)$ in (\ref{eq:UCB}). Similarly, we also have
\begin{align*}
\bar{f}^{P}_{j_{0}}(B)-\bar{f}^{P}_{k_{0}}(B)&\leq\bar{f}^{P}_{(1)}(B)-\bar{f}^{P}_{k_{0}}(B)\leq c_{0}|B|^{\beta}\leq\frac{1}{2}U_{k_{0}}(\tau_{k_{0}},B)+\frac{1}{2}U_{j_{0}}(\tau_{j_{0}},B).
\end{align*}
Combining these two observations, we know from the triangle inequality that there exists some $k \in \{k_{0}, j_{0}\}$ such that $\big|\overline{Y}_{\tau}^{(k)}(B)-\mathbb{E}[\overline{Y}_{\tau}^{(k)}(B)]\big|>\frac{1}{2}U_{k}(\tau,B)$ for some $0\leq\tau\leq\tau_{k}^{\star}(B)$. Therefore, we find that
\begin{align}
\mathbb{P}\big\{&\{\underline{\mathcal{I}}_{B}\not\subset\mathcal{I}_{B}\}\cap\mathcal{G}_{B}\big\} \nonumber \\
& \leq\mathbb{P}\big\{\exists k\in\mathcal{I}_{\mathsf{p}(B)},0\leq\tau\leq\tau_{k}^{\star}(B):\big|\overline{Y}_{\tau}^{(k)}(B)-\mathbb{E}[\overline{Y}_{\tau}^{(k)}(B)]\big|\geq U_{k}(\tau,B)/2\big\} \nonumber \\
 & \leq\sum_{k\in[K]}\mathbb{P}\big\{\exists0\leq\tau\leq\tau_{k}^{\star}(B):\big|\overline{Y}_{\tau}^{(k)}(B)-\mathbb{E}[\overline{Y}_{\tau}^{(k)}(B)]\big|\geq U_{k}(\tau,B)/2\big\} \label{eq:bad-event-1-oracle}.
\end{align}
\item We proceed to control the second term. On the event $\{\mathcal{I}_{B}\not\subset\overline{\mathcal{I}}_{B}\}$, there exist an arm $j_0\in\mathcal{I}_{B}$ and a point $x_{0}\in B$ such that $f_{(1)}(x_{0})-f_{j_0}(x_{0})>8c_{0}|B|^{\beta}$.
Let $k_{0}\in[K]$ be an optimal arm at $x_{0}$, i.e.~$f_{k_{0}}(x_{0})=f_{(1)}(x_{0})$.
It is not hard to verify that $k_{0}\in\underline{\mathcal{I}}_{B}$.
Indeed, the smoothness assumption yields 
\begin{align*}
\sup_{x\in B}|f_{k_{0}}(x)-f_{k_{0}}(x_{0})|&\leq C_\beta \sup_{x\in B}\|x-x_{0}\|_{\infty}^{\beta}\leq\frac{1}{2}c_{0}|B|^{\beta}, \\
\sup_{x\in B}|f_{(1)}(x)-f_{(1)}(x_{0})|&\leq C_\beta \sup_{x\in B}\|x-x_{0}\|_{\infty}^{\beta}\leq\frac{1}{2}c_{0}|B|^{\beta}.
\end{align*}
As a result, applying the triangle inequality shows that
\begin{align*}
\sup_{x\in B}|f_{(1)}(x)-f_{k_{0}}(x)| & \leq\sup_{x\in B}|f_{(1)}(x)-f_{(1)}(x_{0})|+|f_{(1)}(x_{0})-f_{k_{0}}(x_{0})| \\
& \quad +\sup_{x\in B}|f_{k_{0}}(x_{0})-f_{k_{0}}(x)| \\
& \leq c_{0}|B|^{\beta}.
\end{align*}
It then follows from the definition of $\underline{\mathcal{I}}_{B}$ in \eqref{eq:arm_set_lb} that $k_{0}\in\underline{\mathcal{I}}_{B}$. 
Therefore, on the event $\{\underline{\mathcal{I}}_{B}\subset\mathcal{I}_{B}\}\cap\{\mathcal{I}_{B}\not\subset\overline{\mathcal{I}}_{B}\}$, arms $j_{0}$ and $k_{0}$ remain active before bin $B$ gets split. Consequently, by the elimination procedure in Algorithm~\ref{alg:EA-TL}, we obtain
\begin{align}
 \big| \overline{Y}_{\tau^{\star}_{k_{0}}(B)}^{(k_{0})}(B) - \overline{Y}_{\tau_{j_{0}}^{\star}(B)}^{(j_0)}(B) \big| \leq  U_{k_{0}}(\tau^{\star}_{k_{0}}(B),B) + U_{j_{0}}(\tau^{\star}_{j_{0}}(B),B).\label{eq:Y_bar_gap}
\end{align}
In addition, by the smoothness condition, it is easy to see that for any $k\in[K]$ and $x\in B$, the average reward satisfies
\begin{align*}
    \big|\bar{f}^{Q}_{k}(B)-f_{k}(x)\big| &= \bigg|\frac{1}{Q(B)} \int_{B} \big(f_{k} (y) - f_{k}(x) \big) \,\mathrm{d}Q_{X}(y) \bigg| \\
    & \leq \frac{1}{Q(B)} \int_{B} \big| f_{k} (y) - f_{k}(x) \big| \,\mathrm{d}Q_{X}(y) \\
    & \leq \frac{1}{Q(B)} \int_{B} \frac{1}{2}c_{0}|B|^{\beta} \,\mathrm{d}Q_{X}(y) \\
    & = \frac{1}{2}c_{0}|B|^{\beta}.
\end{align*}
Consequently, we can lower bound
\begin{align}
\bar{f}^{Q}_{k_{0}}(B) & \geq f_{k_{0}}(x_{0})-\frac{1}{2} c_{0}|B|^{\beta}=f_{(1)}(x_{0})-\frac{1}{2}c_{0}|B|^{\beta}\nonumber \\
 & >f_{j_0}(x_{0})+\frac{15}{2}c_{0}|B|^{\beta}>\bar{f}^{Q}_{j_0}(B)+7c_{0}|B|^{\beta}\nonumber \\
 & \geq\bar{f}^{Q}_{j_0}(B)+\frac{3}{2}U_{j_{0}}(\tau^{\star}_{j_{0}}(B),B)+ \frac{3}{2}U_{k_{0}}(\tau^{\star}_{k_{0}}(B),B),\label{eq:f_bar_gap}
\end{align}
where the last step holds due to (\ref{eq:tau-star}). Clearly, the above inequality also holds for $\bar{f}^{P}_{k_{0}}(B) - \bar{f}^{P}_{j_{0}}(B)$.
Combined with (\ref{eq:Y_bar_gap}), this implies that the following holds for either $k=k_{0}$ or $k=j_{0}$:
\begin{align*}
\big|\overline{Y}_{\tau_{k}^{\star}(B)}^{(k)}(B)-\mathbb{E}[\overline{Y}_{\tau_{k}^{\star}(B)}^{(k)}(B)]\big|>\frac{1}{2}U_{k}(\tau^{\star}_{k}(B),B).
\end{align*}
Therefore, we obtain that
\begin{align}
 & \mathbb{P}\big\{\{\underline{\mathcal{I}}_{B}\subset\mathcal{I}_{B}\}\cap\{\mathcal{I}_{B}\not\subset\overline{\mathcal{I}}_{B}\}\cap\mathcal{G}_{B}\big\}\nonumber \\
 & \quad\leq\mathbb{P}\big\{\exists k\in\mathcal{I}_{\mathsf{p}(B)}:\big|\overline{Y}_{\tau_{k}^{\star}(B)}^{(k)}(B)-\mathbb{E}[\overline{Y}_{\tau_{k}^{\star}(B)}^{(k)}(B)]\big|>U_{k}(\tau^{\star}_{k}(B),B)/2\big\} \nonumber \\
 & \quad\leq \sum_{k\in[K]}\mathbb{P}\big\{\big|\overline{Y}_{\tau_{k}^{\star}(B)}^{(k)}(B)-\mathbb{E}[\overline{Y}_{\tau_{k}^{\star}(B)}^{(k)}(B)]\big|>U_{k}(\tau^{\star}_{k}(B),B)/2\big\} \label{eq:bad-event-2-oracle}.
\end{align}
\item Combining \eqref{eq:bad-event-1-oracle} and \eqref{eq:bad-event-2-oracle}, we find that
\begin{align}
\mathbb{P}\big\{\mathcal{E}_{B}^{c}\cap\mathcal{G}_{B}\big\} & \leq\sum_{k\in[K]}\mathbb{P}\big\{\exists0\leq\tau\leq\tau_{k}^{\star}(B):\big|\overline{Y}_{\tau}^{(k)}(B)-\mathbb{E}[\overline{Y}_{\tau}^{(k)}(B)]\big|\geq U_{k}(\tau,B)/2\big\} \nonumber\\
 & \quad+\sum_{k\in[K]}\mathbb{P}\big\{\big|\overline{Y}_{\tau_{k}^{\star}(B)}^{(k)}(B)-\mathbb{E}[\overline{Y}_{\tau_{k}^{\star}(B)}^{(k)}(B)]\big|>U_{k}(\tau^{\star}_{k}(B),B)/2\big\} \nonumber \\ 
&\leq 2\sum_{k\in[K]}\mathbb{P}\big\{\exists0\leq\tau\leq\tau_{k}^{\star}(B):\big|\overline{Y}_{\tau}^{(k)}(B)-\mathbb{E}[\overline{Y}_{\tau}^{(k)}(B)]\big|\geq U_{k}(\tau,B)/2\big\}
 \label{eq:prob-Ec-F-temp}.
\end{align}

As a result, it suffices to bound $\mathbb{P}\big\{\exists0\leq\tau\leq\tau_{k}^{\star}(B):\big|\overline{Y}_{\tau}^{(k)}(B)-\mathbb{E}[\overline{Y}_{\tau}^{(k)}(B)]\big|\geq U_{k}(\tau,B)/2\big\}$, which is established in Lemma \ref{lemma:prob-Ec-F} below, with the proof deferred to Appendix~\ref{sec:Proof of lemma:prob-Ec-F}.

\begin{lemma}
\label{lemma:prob-Ec-F}
Instate the assumptions of Theorem~\ref{thm:upper-bound}. For any fixed $k\in[K]$ and $B\in \mathcal{B}_i$ with $0 \leq i < l^{\star}$, one has
\begin{align}
& \mathbb{P}\big\{\exists0\leq\tau\leq\tau_{k}^{\star}(B):\big|\overline{Y}_{\tau}^{(k)}(B)-\mathbb{E}[\overline{Y}_{\tau}^{(k)}(B)]\big|\geq U_{k}(\tau,B)/2\big\} \nonumber  \\
& \qquad \lesssim
\frac{1}{n_{Q}}2^{(d+2\beta)i}\log^{+}\big(n_{Q}2^{-(d+2\beta)i}\big),
\end{align}
Moreover, if $0\leq i<l_{4}\wedge l_{5}$, one further has
\begin{align}
\mathbb{P}\big\{&\exists0\leq\tau\leq\tau_{k}^{\star}(B):\big|\overline{Y}_{\tau}^{(k)}(B)-\mathbb{E}[\overline{Y}_{\tau}^{(k)}(B)]\big|\geq U_{k}(\tau,B)/2\big\} \nonumber \\
& \lesssim \exp\left(-\frac{c^{\star}\kappa}{K}n_{P}2^{-(d+2\beta+\gamma)i}\right) + \frac{2^{(d+2\beta+\gamma)i}}{\kappa n_{P}}
\wedge \frac{2^{(d+2\beta)i}}{n_{Q}} + \frac{1}{n^{10}_{P}}.
\end{align}

% In addition, the inequalities above also hold for $\mathbb{P}\big\{\big|\overline{Y}_{\tau_{k}^{\star}(B)}^{(k)}(B)-\mathbb{E}[\overline{Y}_{\tau_{k}^{\star}(B)}^{(k)}(B)]\big|>U_{k}(\tau^{\star}_{k}(B),B)/2\big\}$.
\end{lemma}

With Lemma~\ref{lemma:prob-Ec-F} in hand, applying it to (\ref{eq:prob-Ec-F-temp}) completes the proofs for (\ref{eq:prob-Ec-F-i-large}) and (\ref{eq:prob-Ec-F-i-small}).

\end{itemize}

\subsection{Proof of Lemma \ref{lemma:prob-Ec-F}}
\label{sec:Proof of lemma:prob-Ec-F}
Fix an arbitrary arm $k\in[K]$ and an arbitrary bin $B \in \mathcal{B}_{i}$ such that $0\leq i < l^{\star}$ and $Q_X(B)>0$. 
% \begin{itemize}
% \item
% Let us start with $\mathbb{P}\big\{ \exists0\leq\tau\leq\tau_{k}^{\star}(B):\big|\overline{Y}_{\tau}^{(k)}(B)-\mathbb{E}[\overline{Y}_{\tau}^{(k)}(B)]\big|\geq U_{k}(\tau,B)/2\big\}$.  
Conditioned on $\{X_{t}^{P}\}_{t=1}^{n_{P}}$, we know from Lemma~\ref{lemma:exist-t-AH} that
\begin{align}
& \mathbb{P} \big\{\exists0\leq\tau\leq\tau_{k}^{\star}(B):\big|\overline{Y}_{\tau}^{(k)}(B)-\mathbb{E}[\overline{Y}_{\tau}^{(k)}(B)]\big|\geq U_{k}(\tau,B)/2\,\big|\,\{X_{t}^{P}\}_{t=1}^{n_{P}}\big\} \nonumber \\
& \quad \lesssim\frac{\tau_{k}^{\star}(B)}{n_{Q}|B|^{d}} + \frac{1}{\kappa n_{P}|B|^{d+2\beta+\gamma}} 
\wedge \frac{1}{n_{Q}|B|^{d+2\beta}}
.\label{eq:regret-b-prob-ub}
\end{align}
This suggests we need to control $\tau_{k}^{\star}(B)$, which is defined to the smallest nonnegative integer such that $U_{k}(\tau,B) \leq 2C_{\beta}|B|^{\beta}$ .

To this end, recall the definition of $U_{k}(\tau,B)$ in \eqref{eq:UCB}. Since $n_{k}^{P}(B) \geq 0$, it is straightforward to upper bound
\begin{align}
\tau_{k}^{\star}(B)\leq c_{2}|B|^{-2\beta}\log^{+}\big(n_{Q}|B|^{d+2\beta}\big),\label{eq:tau-B-k-non-concen}
\end{align}
where $c_{2}\defn 2/C_{\beta}^2 \vee 1$. Meanwhile, as $\tau \geq 0$, we also have the following upper bound
\begin{align}
\tau^{\star}_{k}(B) \leq n_{Q}|B|^{d}\exp\left(-\frac{1}{2}C_{\beta}^2n^{P}_{k}(B)|B|^{2\beta}\right), \label{eq:tau-B-k-exp}
\end{align}
provided $\frac{1}{2}C_{\beta}^2n^{P}_{k}(B)|B|^{2\beta}\geq 1$.
Let us define the event 
\begin{align}
\cA_{B} \defn\Big\{n_{k}^{P}(B)\geq\frac{ \underline{q}c_{\gamma}\kappa n_{P}}{2K}|B|^{d+\gamma}\Big\}.\label{eq:event-aux-sample-size-lb}
\end{align} 
On the event $\cA_{B}$, recalling $c^{\star}\defn \frac{1}{4}\underline{q}c_{\gamma}C_{\beta}^2$, one has
\begin{align}
\tau_{k}^{\star}(B) & \leq n_{Q}|B|^d \exp\left(-\frac{c^{\star}\kappa}{K}n_{P}|B|^{d+2\beta+\gamma}\right). \label{eq:tau-B-k-concen}
\end{align}
provided $\frac{1}{2}C_{\beta}^2n^{P}_{k}(B)|B|^{2\beta} \geq \frac{c^{\star}\kappa}{K}n_{P}|B|^{d+2\beta+\gamma} = \frac{c^{\star}\kappa}{K}n_{P}2^{-(d+2\beta+\gamma)i} \geq 1$, or equivalently, $0\leq i < l_{5}\defn\big\lceil \frac{1}{d+2\beta+\gamma}\log(\frac{c^{\star}\kappa n_{P}}{K})\big\rceil$.
Moreover, note that $n_{k}^{P}(B)$ is a sum of $n_{P}$ i.i.d.~zero-mean Bernoulli random variables. By Assumption~\ref{assumption:bounded-density} and Definitions~\ref{def:transfer}--\ref{def:explore-coef}, one can lower bound the expectation of $n_{k}^{P}(B)$ by
\begin{align*}
\mathbb{E}[n_{k}^{P}(B)] & \geq P_{X}(B)\frac{ \kappa  n_{P}}{K}\geq c_{\gamma}|B|^{\gamma}Q_{X}(B)\frac{ \kappa  n_{P}}{K}\geq\frac{\underline{q} c_{\gamma} \kappa n_{P}}{K}|B|^{d+\gamma} = \frac{\underline{q} c_{\gamma} \kappa n_P}{K}2^{-(d+\gamma)i}.
\end{align*}
Recall $l_{4} \defn \big\lceil\frac{1}{d+\gamma}\log_{2}(\frac{\underline{q}c_{\gamma} \kappa n_{P}}{80K\log(n_{P})})\big\rceil$. If $0 \leq i < l_{4}$, one has $\mathbb{E}[n_{k}^{P}(B)]\geq80\log(n_{P})$. 
As a consequence, we can invoke the Chernoff bound to find
\begin{align}
\mathbb{P}\{\cA_{B}^{c}\} & =\mathbb{P}\Big\{ n_{k}^{P}(B)<\frac{\underline{q}c_{\gamma}\kappa n_{P}}{2K}|B|^{d+\gamma}\Big\} \leq \mathbb{P}\Big\{ n_{k}^{P}(B)< \frac{1}{2} \mathbb{E}[ n_{k}^{P}(B)]\Big\} \nonumber \\
 &  \leq \exp(-\mathbb{E}[n_{k}^{P}(B)]/8) \leq n^{-10}_{P} .\label{eq:prob-G-c}
\end{align}

With these bounds in place, we can readily control $\mathbb{P}\big\{ \exists0\leq\tau\leq\tau_{k}^{\star}(B):\big|\overline{Y}_{\tau}^{(k)}(B)-\mathbb{E}[\overline{Y}_{\tau}^{(k)}(B)]\big|\geq U_{k}(\tau,B)/2\big\}$. 
In light of (\ref{eq:regret-b-prob-ub}), we can upper bound
\begin{align}
 \mathbb{P}\big\{& \exists0\leq\tau\leq\tau_{k}^{\star}(B):\big|\overline{Y}_{\tau}^{(k)}(B)-\mathbb{E}[\overline{Y}_{\tau}^{(k)}(B)]\big| \geq U_{k}(\tau,B)/2\big\} \nonumber  \\
&  \overset{}{\lesssim} \frac{\tau_{k}^{\star}(B)}{n_{Q}|B|^{d}} + \frac{1}{\kappa n_{P}} |B|^{-(d+2\beta+\gamma)} 
\wedge \frac{1}{n_{Q}} |B|^{-(d+2\beta)} 
\nonumber \\
& \overset{(\mathrm{i})}{\lesssim}\frac{1}{n_{Q}} |B|^{-(d+2\beta)}\log^{+}\big(n_{Q}|B|^{d+2\beta}\big) + \frac{1}{\kappa n_{P}} |B|^{-(d+2\beta+\gamma)} 
\wedge \frac{1}{n_{Q}} |B|^{-(d+2\beta)} 
\nonumber \\ 
&   \asymp \frac{1}{n_{Q}}2^{(d+2\beta)i}\log^{+} (n_{Q}2^{-(d+2\beta)i} ) . \label{eq:eq:regret-b-prob-ub-large-i}
\end{align}
where (i) utilizes the upper bound of $\tau_{k}^{\star}(B)$ in (\ref{eq:tau-B-k-non-concen}). 

Moreover, if $0\leq i < l_{4}\wedge l_{5}$,  we can further bound
\begin{align}
\mathbb{P}\big\{ & \exists0\leq\tau\leq\tau_{k}^{\star}(B):\big|\overline{Y}_{\tau}^{(k)}(B)-\mathbb{E}[\overline{Y}_{\tau}^{(k)}(B)]\big|\geq U_{k}(\tau,B)/2\big\} \nonumber \\
 & = \mathbb{P}\big\{\exists0\leq\tau\leq\tau_{k}^{\star}(B):\big|\overline{Y}_{\tau}^{(k)}(B)-\mathbb{E}[\overline{Y}_{\tau}^{(k)}(B)]\big|\geq U_{k}(\tau,B)/2\,\big|\,\cA_{B}\big\} \nonumber \\
 & \quad+\mathbb{P}\big\{\exists0\leq\tau\leq\tau_{k}^{\star}(B):\big|\overline{Y}_{\tau}^{(k)}(B)-\mathbb{E}[\overline{Y}_{\tau}^{(k)}(B)]\big|\geq U_{k}(\tau,B)/2\,\big|\,\cA_{B}^{c}\big\}\mathbb{P}\{\cA_{B}^{c}\}\nonumber \\
 & \leq  \mathbb{P}\big\{\exists0\leq\tau\leq\tau_{k}^{\star}(B):\big|\overline{Y}_{\tau}^{(k)}(B)-\mathbb{E}[\overline{Y}_{\tau}^{(k)}(B)]\big|\geq U_{k}(\tau,B)/2\,\big|\,\cA_{B}\big\} +\mathbb{P}\{\cA_{B}^{c}\}\nonumber \\
 & \overset{(\mathrm{i})}{\lesssim} \frac{\tau_{k}^{\star}(B)}{n_{Q}|B|^{d}} + \frac{1}{\kappa n_{P}|B|^{d+2\beta+\gamma}} 
 \wedge \frac{1}{n_{Q}|B|^{d+2\beta}} 
 + \frac{1}{n^{10}_{P}}\nonumber \\
 & \overset{(\mathrm{ii})}{\lesssim} \exp\left(-\frac{c^{\star}\kappa}{K}n_{P}|B|^{d+2\beta+\gamma}\right)  + \frac{1}{\kappa n_{P}|B|^{d+2\beta+\gamma}} 
 \wedge \frac{1}{n_{Q}|B|^{d+2\beta}} 
 + \frac{1}{n^{10}_{P}} \nonumber \\
 & = \exp\left(-\frac{c^{\star}\kappa}{K}n_{P}2^{-(d+2\beta+\gamma)i}\right) + \frac{2^{(d+2\beta+\gamma)i}}{\kappa n_{P}} 
 \wedge \frac{2^{(d+2\beta)i}}{n_{Q}} + \frac{1}{n^{10}_{P}}
 . \label{eq:eq:regret-b-prob-ub-small-i}
\end{align}
Here, (i) follows from (\ref{eq:regret-b-prob-ub}), (\ref{eq:prob-G-c}) and the fact that the event $\cA_{B}$ belongs to the $\sigma$-algebra generated by $\{X_{t}^{P}\}_{t=1}^{n_{P}}$; (ii) is arises from the upper bound of $\tau_{k}^{\star}(B)$ in (\ref{eq:tau-B-k-concen}).
Putting (\ref{eq:eq:regret-b-prob-ub-large-i}) and (\ref{eq:eq:regret-b-prob-ub-small-i}) together completes the proof of Lemma~\ref{lemma:prob-Ec-F}.

\subsection{Proof of Lemma \ref{lemma:non-leaf-upper-oracle-live}}
\label{sec:Proof-of-lemma:non-leaf-upper-oracle-live}

Recall the definition of $R^{\lf}(B)$ in (\ref{eq:R-leaf-B}). For any bin $B \in \mathcal{B}_{i}$ such that $0 \leq i < l^{\star}$ and $Q_{X}(B) > 0$, one can upper bound
\begin{align}
 \mathbb{E}& [R^{\lf}(B)\mathds{1}\{\mathcal{E}_{B}\cap\mathcal{G}_{B}\}] \nonumber \\
 & \quad=\mathbb{E}\bigg[\mathds{1}\{\mathcal{E}_{B}\cap\mathcal{G}_{B}\}\sum_{t=1}^{n_{Q}}\big(f_{\pi^{\star}(X_{t}^{Q})}(X_{t}^{Q})-f_{\pi_{t}(X_{t}^{Q})}(X_{t}^{Q})\big)\mathds{1}\{X_{t}^{Q}\in B,B\in\mathcal{L}_{t}\}\bigg] \nonumber \\
 & \quad\overset{(\mathrm{i})}{\leq}\mathbb{E}\bigg[\sum_{t=1}^{n_{Q}} c_{1}|B|^{\beta}\mathds{1}\{0<f_{(1)}(X_{t}^{Q})-f_{(2)}(X_{t}^{Q})\leq c_{1}|B|^{\beta}\}\mathds{1}\{X_{t}^{Q}\in B,B\in\mathcal{L}_{t}\}\bigg] \nonumber \\
 & \quad\overset{(\mathrm{ii})}{\lesssim} |B|^{\beta}Q_{X}(0<f_{(1)}(X)-f_{(2)}(X)\leq c_{1}|B|^{\beta}\,|\,X\in B)\sum_{k=1}^{K}\mathbb{E}[\tau_{k}^{\star}(B)]. \label{eq:non-leaf-upper-oracle-live-temp}
\end{align}
Here, (i) follows from (\ref{eq:reward-gap-event-ancestor}) and (ii) holds due to the round limits in Algorithm~\ref{alg:UCB-TL}. 
Note that we can upper bound
\begin{align}
\sum_{B\in\mathcal{B}_{i}:Q_{X}(B) > 0} & Q_{X}(0<f_{(1)}(X)-f_{(2)}(X)\leq c_{1}|B|^{\beta}\,|\,X\in B) \nonumber \\
& =\sum_{B\in\mathcal{B}_{i}:Q_{X}(B) > 0}\frac{1}{Q_{X}(B)}Q_{X} \big(0<f_{(1)}(X)-f_{(2)}(X)\leq c_{1}|B|^{\beta},X\in B \big)\nonumber \\
 & \overset{(\mathrm{i})}{\leq}\frac{1}{\underline{q}|B|^{d}}Q_{X} \big(0<f_{(1)}(X)-f_{(2)}(X)\leq c_{1}|B|^{\beta} \big)\nonumber \\
 & \overset{(\mathrm{ii})}{\leq}\frac{1}{\underline{q}|B|^{d}} C_{\alpha}(c_{1}|B|^{\beta})^{\alpha}\nonumber  \\
 & \lesssim|B|^{\beta\alpha-d} \label{eq:cond-prob-Q-gap-ub}
\end{align} 
where (i) follows from Assumption~\ref{assumption:bounded-density} and (ii) arises from Assumption~\ref{assumption:margin}.
Clearly, regret is only incurred in a bin $B$ such that $Q_{X}(B) > 0$. As a result, summing over all bins in $\mathcal{B}_{i}$, we combine (\ref{eq:non-leaf-upper-oracle-live-temp}) and (\ref{eq:cond-prob-Q-gap-ub}) to obtain that 
\begin{align}
\mathbb{E}[R^{\lf}_{i}] & =\sum_{B\in\mathcal{B}_{i}}\mathbb{E}\big[R^{\lf}(B)\mathds{1}\{\mathcal{E}_{B}\cap\mathcal{G}_{B}\}\big]\nonumber \\
 % & \leq \sum_{B\in\mathcal{B}_{i}} c_{1}|B|^{\beta}Q_{X}(0<f_{(1)}(X)-f_{(2)}(X)\leq c_{1}|B|^{\beta}\,|\,X\in B)K\max_{k\in[K]}\mathbb{E}[\tau_{k}^{\star}(B)]\nonumber \\
 & \lesssim K|B|^{\beta}\max_{B\in\mathcal{B}_{i},k\in[K]}\mathbb{E}[\tau_{k}^{\star}(B)] \nonumber \\
 & \qquad \sum_{B\in\mathcal{B}_{i}, Q_{X}(B) > 0}Q_{X}(0<f_{(1)}(X)-f_{(2)}(X)\leq c_{1}|B|^{\beta}\,|\,X\in B)\nonumber \\
 & \lesssim K2^{(d-(1+\alpha)\beta )i}\max_{B\in\mathcal{B}_{i},k\in[K]}\mathbb{E}[\tau_{k}^{\star}(B)]. \label{eq:regret-leaf-bad-event}
\end{align}

Therefore, it remains to upper bound $\max_{B\in\mathcal{B}_{i},k\in[K]}\mathbb{E}[\tau_{k}^{\star}(B)]$. Towards this, fix an arbitrary bin $B\in\mathcal{B}_{i}$ and an arbitrary arm $k\in[K]$. In view of (\ref{eq:tau-B-k-non-concen}), one can bound directly
\begin{align}
\max_{B\in\mathcal{B}_{i},k\in[K]}\mathbb{E}[\tau_{k}^{\star}(B)]\lesssim2^{2\beta i}\log^{+}\big(n_{Q}2^{-(d+2\beta)i}\big). \label{eq:tau-ub-exp-i-large}
\end{align}
for any $i\geq0$.
Moreover, if $0\leq i< l_{4}\wedge l_{5}$, recalling the event $\cA_{B}$ defined in \eqref{eq:event-aux-sample-size-lb}, we can combine (\ref{eq:tau-B-k-non-concen}), (\ref{eq:tau-B-k-concen}), and (\ref{eq:prob-G-c}) to find that
\begin{align}
\mathbb{E}[\tau_{k}^{\star}(B)] & =\mathbb{E}[\tau_{k}^{\star}(B)\,|\,\cA_{B}]\mathbb{P}\{\cA_{B}\}+\mathbb{E}[\tau_{k}^{\star}(B)\,|\,\cA_{B}^{c}]\mathbb{P}\{\cA_{B}^{c}\} \nonumber \\
& \leq\mathbb{E}[\tau_{k}^{\star}(B)\,|\,\cA_{B}]+\mathbb{E}[\tau_{k}^{\star}(B)\,|\,\cA_{B}^{c}]\mathbb{P}\{\cA_{B}^{c}\}\nonumber\\
& \lesssim n_{Q}|B|^d \exp\left(-\frac{c^{\star}\kappa}{K}n_{P}|B|^{d+2\beta+\gamma}\right)
 +\frac{1}{n^{10}_P}|B|^{-2\beta}\log^{+}\big(n_{Q}|B|^{d+2\beta}\big) \nonumber\\
& = n_{Q}2^{-di}\exp\left(-\frac{c^{\star}\kappa}{K}n_{P}2^{-(d+2\beta+\gamma)i}\right)+ \frac{1}{n^{10}_P}2^{2\beta i}\log^{+}\big(n_{Q}2^{-(d+2\beta)i}\big) \label{eq:tau-ub-exp-i-small}.
\end{align}

Plugging the upper bounds (\ref{eq:tau-ub-exp-i-small}) and (\ref{eq:tau-ub-exp-i-large}) back into (\ref{eq:regret-leaf-bad-event}) completes the proof of Lemma~\ref{lemma:non-leaf-upper-oracle-live}.

\subsection{Proof of Lemma \ref{lemma:leaf-upper-oracle}}

\label{subsec:Proof-of-lemma:leaf-upper-oracle}

For any bin $B\in\mathcal{B}_{l^{\star}}$, it is straightforward to bound
\begin{align*}
\mathbb{E}\big[R^{\tree}(B)\mathds{1}\{\mathcal{G}_{B}\}\big] & = \mathbb{E}\bigg[\mathds{1}\{\mathcal{G}_{B}\}\sum_{t=1}^{n_{Q}}\big(f_{\pi^{\star}(X_{t}^{Q})}(X_{t}^{Q})-f_{\pi_{t}(X_{t}^{Q})}(X_{t}^{Q})\big)\mathds{1}\{X_{t}^{Q}\in B,\,B\in\mathcal{T}_{t}\}\bigg]\\
 & \leq\mathbb{E}\bigg[\sum_{t=1}^{n_{Q}} c_{1}|B|^{\beta}\mathds{1}\{0<f_{(1)}(X_{t}^{Q})-f_{(2)}(X_{t}^{Q})\leq c_{1}|B|^{\beta}\}\mathds{1}\{X_{t}^{Q}\in B\}\bigg]\\
 & \lesssim \sum_{t=1}^{n_{Q}} |B|^{\beta}Q_{X}\big(0<f_{(1)}(X)-f_{(2)}(X)\leq c_{1}|B|^{\beta},X\in B\big) \\
& = \sum_{t=1}^{n_{Q}} 2^{-\beta l^{\star}} Q_{X}\big(0<f_{(1)}(X)-f_{(2)}(X)\leq c_{1}2^{-\beta l^{\star}},X\in B\big)
\end{align*}
where the first inequality arises from (\ref{eq:reward-gap-event-ancestor}).
As a consequence, we can sum over all bins $B\in\mathcal{B}_{l^{\star}}$ to obtain
\begin{align*}
\mathbb{E}[R_{l^\star}] & =\sum_{B\in\mathcal{B}_{ l^{\star}}}\mathbb{E}\big[R^{\tree}(B)\mathds{1}\{\mathcal{G}_{B}\}\big] 
\lesssim \sum_{t=1}^{n_{Q}} 2^{-\beta l^{\star}}Q_{X}\big(0<f_{(1)}(X)-f_{(2)}(X)\leq c_{1}2^{-\beta l^{\star}}\big) \\
 & \leq n_{Q} 2^{-\beta l^{\star}} C_{\alpha} \big( c_{1}2^{-\beta l^{\star}} \big)^{\alpha} \lesssim n_{Q}2^{- (1+\alpha)\beta  l^{\star}},
\end{align*}
where we use the margin condition (Assumption~\ref{assumption:margin}) in the last line. This completes the proof of Lemma \ref{lemma:leaf-upper-oracle}.

\section{Proof for Theorems~\ref{thm:lower-bound} and Theorem~\ref{thm:lower-bound-adaptive}}

\label{sec:Proof-of-Lower-Bound}

Note that the self-similar function space $\Pi(K,\beta,\alpha,\gamma,\kappa,l_{0},b)$ is a subset of the general function space $\Pi(K,\beta,\alpha,\gamma,\kappa)$.
Therefore, it suffices to prove Theorem~\ref{thm:lower-bound-adaptive}, from which Theorem~\ref{thm:lower-bound} follows as an immediate consequence.

To this end, we shall leverage the lower-bound construction ideas in \citet{audibert2007fast,kpotufe2021marginal}. A significant technical difference lies in the use of self-similar functions for the construction of the reward functions.

\subsection*{Step 1: constructing a collection of hypotheses}

Fix some $r\in(0,1)$ to be specified later. Let $G=\{x_{1},\cdots,x_{N^{d}}\}$ be a regular grid in $\mathcal{X}$ where $N\defn\lfloor r^{-1}\rfloor$, i.e.
\begin{align*}
G=\{x\in\mathcal{\mathcal{X}}:x_{i}=(k_{i}-1/2)r,\,k_{i}\in[N],\,i\in[d]\}.
\end{align*}
Let us define the integer $m\defn\lfloor c_{m}r^{\alpha\beta-d}\rfloor$ for some sufficiently small constant $c_{m}>0$ such that $1\leq m\leq N^{d}$.
Note that this is feasible due to the condition $d\geq\alpha\beta$.
In view of the celebrated Varshamov--Gilbert bound \citep[Lemma 4.7]{massart2007concentration}, we can find a set of well-separated vectors $\Omega_{m}=\{\omega^{(i)}\}_{i=1}^{M}\subset\{-1,1\}^{m}$ such that 
\begin{align}
\log_{2}(M)\geq\frac{m}{8},\quad\text{and}\quad\rho(\omega^{(i)},\omega^{(j)})\geq\frac{m}{8},\quad\forall1\leq i\neq j\leq M.\label{eq:V-G-bound}
\end{align}
where $\rho(\omega,\omega')\coloneqq|\{k\in[m]:\omega_{k}\neq\omega_{k}'\}|$.
In what follows, we shall construct a collection $\mathcal{H}=\big\{(Q^{\omega},P^{\omega},\{\mu^{\omega}(\cdot\,|\,x)\}_{x\in\mathcal{X}}):\omega\in\Omega_{m}\big\}$ of probability distributions that characterize contextual $K$-armed bandits. Here, $Q^{\omega}$ (resp.~$P^{\omega}$) is a probability distribution over $\mathcal{X}\times[0,1]^{K}$ representing the target bandit (resp.~source bandit), and $\{\mu^{\omega}(\cdot\,|\,x)\}_{x\in\mathcal{X}}$ is a family of probability distributions over $[K]$ that describe the source arm selection policy.

\paragraph*{Constructing covariate distributions}
We choose the covariate distribution to be independent of $\omega$ under both $P$ and $Q$, namely, $P_{X}^{\omega}\equiv P_{X}$ and $Q_{X}^{\omega}\equiv Q_{X}$ for any $\omega\in\Omega_{m}$.

The density function of the covariate distribution $Q_{X}$, denoted by $q_{X}$, satisfies
\begin{align}
q_{X}(x)=\begin{cases}
q_{1}, & \text{if }x\in\bigcup_{i=1}^{m}B(x_{i},r/8),\\
q_{0}, & \text{if }x\in\mathcal{X}\setminus\bigcup_{i=1}^{m}B(x_{i},r/2),\\
0 & \text{otherwise.}
\end{cases}\label{eq:Q-density-lb}
\end{align}
Here, we define
\begin{align}
\label{eq:Q-density-value}
q_{1}\coloneqq\frac{w}{\mathsf{Leb}\big(B(x_{1},r/8)\big)},\quad\text{and}\quad q_{0}\coloneqq\frac{1-mw}{\mathsf{Leb}\big(\mathcal{X}\setminus\bigcup_{i=1}^{m}B(x_{i},r/2)\big)},
\end{align}
where $w\defn c_{w}r^{d}$ for some constant $c_{w}>0$, and $\mathsf{Leb}$ denote the Lebesgue measure.
As for $P_{X}$, we set the density function $p_{X}$ by
\begin{align}
\label{eq:P-density-value}
p_{X}(x)=\begin{cases}
c_{\gamma}r^{\gamma}q_{1}, & \text{if }x\in\bigcup_{i=1}^{m}B(x_{i},r/8),\\
\frac{1-c_{\gamma}4^{-d}mr^{d+\gamma}q_{1}-(1-mr^{d})q_{0}}{mr^{d}(1-2^{-d})},
 & \text{if }x\in\bigcup_{i=1}^{m}\big(B(x_{i},r/2)\setminus B(x_{i},r/4)\big),\\
q_{0}, & \text{if }x\in\mathcal{X}\setminus\bigcup_{i=1}^{m}B(x_{i},r/2),\\
0, & \text{otherwise.}
\end{cases}
\end{align}
% Note that this is feasible because $c_{\gamma},r\in(0,1]$.

\paragraph*{Constructing reward distributions}
Recall that under the covariate shift model, the reward distributions are identical under $P$ and $Q$. For any arm $k\in [K]$, we choose the random reward of arm $k$ conditioned on the covariate $X$ to be a Bernoulli random variable with parameter $f_{k}^{\omega}(X)$.
 The expected reward functions $\{f_{k}^{\omega}\}_{k\in[K]}$ are constructed as follows. We first define the function $\phi:\mathbb{R}_{+}\rightarrow[0,1]$ by
\begin{align*}
\phi(x)=\begin{cases}
1, & \text{if }0\leq x<1/8,\\
2-8x, & \text{if }1/8\leq x<1/4,\\
0, & \text{otherwise},
\end{cases}
\end{align*}
and the function $\varphi:\mathbb{R}^{d}\rightarrow[0,1/4]$ via
\begin{align*}
\varphi(x)\coloneqq\widetilde{C}_{\beta}r^{\beta}\phi^{\beta}(\|x\|_{\infty}/r),
\end{align*}
where $\widetilde{C}_{\beta}=C_{\beta}8^{-\beta}\wedge(1/4)$. Subsequently, for any $\omega\in\Omega$, we define the reward function $f_{k}^{\omega}:\mathcal{X}\rightarrow[0,1]$ by
\begin{align}
\label{eq:reward-func-lb}
f_{k}^{\omega}(x)=
\begin{cases}
	1/2+\sum_{i=1}^{m}\omega_{i}\varphi(x-x_{i})\mathds{1}\{x\in B(x_{i},r/2)\}, & \text{ if }\,\, k = 1, \\
	1/2, & \text{ if }\,\, k \neq 1.
\end{cases}
\end{align}
Let us denote $\delta_{\beta}\coloneqq\widetilde{C}_{\beta}r^{\beta}\in(0,1/4]$.
We make a few remarks as follows.
\begin{itemize}
\item For any $\omega\in\Omega_{m}$, the reward gap satisfies
\begin{align}
f_{(1)}^{\omega}(x)-f_{(2)}^{\omega}(x) = \delta_{\beta},
 \label{eq:reward-gap-lb}
\end{align}
if $x\in\bigcup_{i=1}^{m}B(x_{i},r/8)$. In addition, for any $x\in\mathcal{X}\setminus\bigcup_{i=1}^{m}B(x_{i},r/4)$, the reward functions are identical, i.e.~$f_{(1)}^{\omega}(x)-f_{(2)}^{\omega}(x)=0$.
\item For any two different $\omega\neq\omega'\in\Omega_{m}$, their corresponding optimal policies $\pi^{\omega,\star}(x)$ and $\pi^{\omega',\star}(x)$ differ if and only if $x\in\{\bigcup_{i}B(x_{i},r/4):i\in[m] \,\,\text{ such that }\,\, \omega_{i}\neq\omega_{i}'\}$.
\end{itemize}

\paragraph*{Constructing source policies}
The source arm selection policy $\{\mu^{\omega,P}(\cdot\,|\,x)\}_{x\in\mathcal{X}}$ is also chosen to be independent of $\omega$. For any $\omega\in\Omega_{m}$, if $x\in\bigcup_{i=1}^{m}B(x_{i},r/8)$, we set $\mu^{\omega,P}(\cdot\,|\,x)$ by
\begin{align*}
\mu^{\omega,P}(k\,|\,x)=
\begin{cases}
	\frac{\kappa}{K}, &\text{ if }\,\, k = 1, \\
	\frac{1}{K-1}-\frac{\kappa}{K(K-1)}, &\text{ if }\,\, k \neq 1.
\end{cases}
\end{align*}
Otherwise, we set $\mu^{\omega,P}(\cdot\,|\,x)$ to be the uniform distribution over $[K]$.

\paragraph*{Verifying assumptions}
Finally, let us verify the constructed hypotheses satisfy the assumptions and definitions introduced in Section~\ref{sec:Problem-formulation}.
\begin{itemize}
\item \emph{Smoothness assumption}. By construction, it suffices to show that $\varphi(x)\in\mathcal{H}(\beta,C_{\beta})$. Towards this, straightforward calculation yields
\begin{align*}
|\varphi(x)-\varphi(\widetilde{x})| & =\widetilde{C}_{\beta}r^{\beta}\big|\phi^{\beta}(\|x\|_{\infty}/r)-\phi^{\beta}(\|\widetilde{x}\|_{\infty}/r)\big|\\
 & \overset{(\mathrm{i})}{\leq}\widetilde{C}_{\beta}r^{\beta}|\phi(\|x\|_{\infty}/r)-\phi(\|\widetilde{x}\|_{\infty}/r)|^{\beta}\\
 & \overset{(\mathrm{ii})}{\leq}\widetilde{C}_{\beta}r^{\beta}|8(\|x\|_{\infty}/r-\|\widetilde{x}\|_{\infty}/r)|^{\beta}\\
 & \overset{(\mathrm{iii})}{\leq}\widetilde{C}_{\beta}8^{\beta}\|x-\widetilde{x}\|_{\infty}^{\beta}\\
 & \overset{(\mathrm{iv})}{\leq}C_{\beta}\|x-\widetilde{x}\|_{\infty}^{\beta}.
\end{align*}
Here, (i) is true since $|a^{\beta}-b^{\beta}|\leq|a-b|^{\beta}$ holds any $a,b\geq0$ and $0<\beta\leq1$; (ii) follows from the fact that $\phi$ is a Lipschitz function with Lipschitz constant $8$; (iii) arises from the triangle inequality; (iv) is true due to the choice of $\widetilde{C}_{\beta}=C_{\beta}8^{-\beta}\wedge(1/4)$.
This demonstrates that for any $\omega\in\Omega_{m}$, the reward functions satisfy the smoothness assumption (Assumption~\ref{assumption:smooth}).
\item \emph{Marginal assumption}. Recall that the marginal distribution $Q_{X}\omega\equiv Q_{X}$ is independent of $\omega$. By (\ref{eq:Q-density-lb}) and (\ref{eq:reward-gap-lb}), we know that for any $0<\delta\leq\delta_{\beta}$,
\begin{align*}
Q_{X}\Big\{0<f_{(1)}^{\omega}(X)-f_{(2)}^{\omega}(X)<\delta\Big\}=0.
\end{align*}
On the other hand, for any $\delta>\delta_{\beta}$, one know from (\ref{eq:Q-density-lb}), (\ref{eq:Q-density-value}), and (\ref{eq:reward-gap-lb}) that
\begin{align*}
Q_{X}\Big\{0<f_{(1)}^{\omega}(X)-f_{(2)}^{\omega}(X)<\delta\Big\} & \leq\sum_{i=1}^{m}Q_{X}\big(B(x_{i},r/8)\big)=mw\\
 & \leq c_{m}c_{w}r^{\alpha\beta}=c_{m}c_{w}(\delta_{\beta}/\widetilde{C}_{\beta})^{\alpha}\\
 & \leq c_{m}c_{w}\widetilde{C}_{\beta}^{-\alpha}\delta^{\alpha}=c_{m}c_{w}(C_{\beta}^{-\alpha}8^{\alpha\beta}\wedge4^{\alpha})\delta^{\alpha}\\
 & \leq C_{\alpha}\delta^{\alpha}
\end{align*}
as long as $c_{m}$ is sufficiently small. As a result, the reward functions $\{f_{k}^{\omega}\}_{k\in[K]}$ obey the margin condition (Assumption~\ref{assumption:margin}).
\item \emph{Bounded density assumption}. It is straightforward to see that
\begin{align*}
q_{1} & =\frac{w}{(r/4)^{d}}=\frac{c_{w}r^{d}}{(r/4)^{d}}=4^{d}c_{w},\\
q_{0} & =\frac{1-mw}{1-mr^{d}}=\frac{1-c_{m}r^{\alpha\beta-d}\cdot c_{w}r^{d}}{1-c_{m}r^{\alpha\beta-d}\cdot r^{d}}=\frac{1-c_{m}c_{w}r^{\alpha\beta}}{1-c_{m}r^{\alpha\beta}}=1+o(1),
\end{align*}
where the last step holds as long as $r=o(1)$. As a result, we can choose $c_{w}\in[4^{-d}\underline{q},4^{-d}\overline{q}]$ so that the constructed $Q_{X}$ satisfy the bounded density assumption (Assumption~\ref{assumption:bounded-density}).
\item \emph{Transfer exponent}. Fix an arbitrary $x$ such that $x\in B(x_{i},r/8)$ for some $i\in[m]$, and fix an arbitrary $\widetilde{x}\in B(x,r/8)$.
For any $\widetilde{r}\in(0,3/8r]$, we know that $B(\widetilde{x},\widetilde{r})\subset B(x_{i},r/2)$.
It follows that
\begin{align*}
P_{X}\big(B(\widetilde{x},\widetilde{r})\big) & \geq P_{X}\big(B(\widetilde{x},\widetilde{r})\cap B(x_{i},r/8)\big)=q_{1}c_{\gamma}r^{\gamma}\mathsf{Leb}\big(B(\widetilde{x},\widetilde{r})\cap B(x_{1},r/8)\big)\\
 & =q_{1}c_{\gamma}r^{\gamma}\frac{Q_{X}\big(B(\widetilde{x},\widetilde{r})\cap B(x_{i},r/8)\big)}{q_{1}}=c_{\gamma}r^{\gamma}Q_{X}\big(B(\widetilde{x},\widetilde{r})\big)\\
 & \geq c_{\gamma}\widetilde{r}^{\gamma}Q_{X}\big(B(\widetilde{x},\widetilde{r})\big).
\end{align*}
Similarly, for any $x\in\mathcal{X}\setminus\bigcup_{i=1}^{m}B(x_{i},r/2)$, $\widetilde{x}\in B(x,r/2)$ and $\widetilde{r}\in(0,3r/8]$, one has $B(\widetilde{x},\widetilde{r})\subset\mathcal{X}\setminus\bigcup_{i=1}^{m}B(x_{i},r/2)$, and thus
\begin{align*}
P_{X}\big(B(\widetilde{x},\widetilde{r})\big) & =c_{\gamma}Q_{X}\big(B(\widetilde{x},\widetilde{r})\big)\geq c_{\gamma}\widetilde{r}^{\gamma}Q_{X}\big(B(\widetilde{x},\widetilde{r})\big).
\end{align*}
As $Q_{X}(B(x,r/2)) = P_{X}(B(x,r/2))$ for any $x \in G$, one can also verify that the above inequalities also hold for $3r/8 < \widetilde{r} \leq 1$.
As a consequence, by Definition~\ref{def:transfer}, the transfer exponent of $P_{X}$ with respect to $Q_{X}$ is equal to $\gamma$ with constant $c_{\gamma}$.
\item \emph{Exploration coefficient}. By construction, it is easy to see that $\min_{k\in[K]}\mu^{\omega,P}(k\,|\,x)=\kappa/K$ if $x\in\bigcup_{i=1}^{m}B(x_{i},r/8)$, and $\min_{k\in[K]}\mu^{\omega,P}(k\,|\,x)=1/K$ otherwise. Therefore, by Definition~\ref{def:explore-coef}, the exploration exponent of the constructed source policy with respect to $Q_{X}$ is equal to $\kappa$.
\item \emph{Self-similarity assumption}. We shall apply a similar argument as in the proof of Example in Section~\ref{subsec:The-self-similarity-assumption} to show that for any $\omega\in\Omega_{m}$, the reward function $f_{1}^{\omega}(x)$ of arm $1$ defined in (\ref{eq:reward-func-lb}) is self-similar. By construction, it is not hard to see that it suffices to show that $\phi^{\beta}(\|x\|_{\infty})$ is self-similar. To this end, for any $\beta>0$ and integer $l\geq0$, standard calculation yields
\begin{align*}
\int_{[0,2^{-l}]^{d}}\Big(\min_{1\leq i\leq d}x_{i}\Big)^{\beta}\,\mathrm{d}x_{1}\cdots\mathrm{d}x_{d}=\frac{d!}{\prod_{i=1}^{d}(\beta+i)}2^{-(d+\beta)l}.
\end{align*}
Therefore, let us fix an arbitrary integer $l\geq3$. Let bin $B_{0}\defn[4^{-1}-2^{-l},4^{-1}]^{d}$ and $x_{0}=4^{-1}1_{d}$, where $1_{d}$ is the all-one vector in $\mathbb{R}^{d}$. It is straightforward to compute
\begin{align*}
\Gamma_{B_{0}}\phi^{\beta}(\|x_0\|_{\infty};Q_{X}) & =\frac{1}{\mathsf{Leb}(B_0)}\int_{B_0}\phi^{\beta}(\|x\|_{\infty})\,\mathrm{d}x\\
 & =2^{ld}\int_{B_0}\|2-8x\|_{\infty}^{\beta}\mathrm{d}x_{1}\cdots\mathrm{d}x_{d}\\
 & =8^{\beta}2^{ld}\int_{[0,2^{-l}]^{d}}\Big(\min_{1\leq i\leq d}x_{i}\Big)^{\beta}\,\mathrm{d}x_{1}\cdots\mathrm{d}x_{d}\\
 & =\frac{d!8^{\beta}}{\prod_{i=1}^{d}(\beta+i)}2^{-\beta l}.
\end{align*}
This means that
\begin{align*}
\sup_{B\in\mathcal{B}_{l}}\sup_{x\in B}\big|\Gamma_{B}\phi^{\beta}(\|x\|_{\infty};Q_{X})-\phi^{\beta}(\|x\|_{\infty})\big| 
& \geq\big|\Gamma_{B_{0}}\phi^{\beta}(\|x_{0}\|_{\infty};Q_{X})-\phi^{\beta}(\|x_{0}\|_{\infty})\big|\\
 & =\Gamma_{B_{0}}\phi^{\beta}(\|x_{0}\|_{\infty};Q_{X})\geq\frac{d!8^{\beta}}{\prod_{i=1}^{d}(\beta+i)}2^{-\beta l}.
\end{align*}
In addition, it is easy to see that the bound also holds for $\Gamma_{B}\phi^{\beta}(\|x\|_{\infty};P_{X|\pi=1})$.
Therefore, we conclude that for any $\omega\in\Omega_{m}$, the constructed reward functions $\{f_{k}^{\omega}\}_{k\in[K]}$ obey the self-similarity condition (Assumption~\ref{assumption:self-similar}) with $l_{0}=3$ and $b=\frac{d!8^{\beta}}{\prod_{i=1}^{d}(\beta+i)}$.
\item Putting these together justifies that the constructed set of hypotheses $\mathcal{H}$ belong to the self-similar function space $\Pi(K,\alpha,C_{\alpha},\beta,C_{\beta},\underline{q},\overline{q},\gamma,c_{\gamma},\kappa,l_{0},b)$.
\end{itemize}

\subsection*{Step 2: bounding the KL divergence between hypotheses}

Fix an arbitrary admissible policy $\pi^{Q}=\{\pi^{Q}_t\}_{t=1}^{n_{Q}}$. For ease of presentation, let us first introduce some notations. We denote by $\mathcal{D}_{n_{Q}}^{Q}\coloneqq\big\{\big(X_{t}^{Q},\pi_{t}^{Q}(X_{t}^{Q}),Y_{t}^{Q,(\pi_{t}^{Q}(X_{t}^{Q}))}\big)\big\}_{t=1}^{n_{Q}}$ and $\mathcal{D}_{n_{P}}^{P}\coloneqq\big\{\big(X_{i}^{P},\pi_{i}^{P}(X_{i}^{P}),Y_{i}^{P,(\pi_{i}^{P}(X_{i}^{P}))}\big)\big\}_{i=1}^{n_{P}}$ the $Q$-data and $P$-data, respectively.
For any $\omega\in\Omega$, let $\mathbb{P}^{\omega}$ denote the joint probability distribution of random variables $\mathcal{D}_{n_{Q}}^{Q}$ and $\mathcal{D}_{n_{P}}^{P}$, where $\big(X_{t}^{Q},Y_{t}^{Q,(1)},\dots,Y_{t}^{Q,(K)}\big)$, $t=1,\dots,n_{Q}$, and $\big(X_{i}^{P},Y_{i}^{P,(1)},\dots,Y_{i}^{P,(K)}\big)$, $i=1,\dots,n_{P}$ are i.i.d.~distributed according to $Q^{\omega}$ and $P^{\omega}$, respectively. Also, we use $\mathbb{E}^{\omega}$ to denote the corresponding expectation.

Next, for each $1\leq i\leq n_{P}$ , let us define the $\sigma$-algebra generated by the samples in the $P$-data up to time $i$, i.e.~$\mathcal{F}_{i}^{P}=\sigma\Big(\big\{\big(X_{s}^{P},\pi_{s}^{P}(X_{s}^{P}),Y_{s}^{P,(\pi_{s}^{P}(X_{s}^{P}))}\big)\big\}_{s=1}^{i-1}\Big)$.
Similarly, for each $1\leq t\leq n_{Q}$, define the $\sigma$-algebra generated by the full $P$-data and observations in the $Q$-data up to time $t$, namely $\mathcal{F}_{t}^{Q}\coloneqq\sigma\Big(\mathcal{D}_{n_{P}}^{P},\big\{\big(X_{s}^{P},\pi_{s}^{P}(X_{s}^{P}),Y_{s}^{P,(\pi_{s}^{P}(X_{s}^{P}))}\big)\big\}_{s=1}^{t-1}\Big)$. 

In addition, for any $x\in\mathcal{X}$ and $k \in [K]$, we define $n_{k}^{\omega,Q}(x)\coloneqq\sum_{t=1}^{n_{Q}}\mathbb{E}^{\omega}[\mathds{1}\{\pi_{t}^{Q}(x)=k\}]$ and $n_{k}^{\omega,P}(x)\coloneqq\sum_{i=1}^{n_{P}}\mathbb{E}^{\omega}[\mathds{1}\{\pi_{i}^{P}(x)=k\}]$.

With these definitions in hand, we can express the probability density $p^{\omega}$ of $\mathbb{P}^{\omega}$ as
\begin{align*}
p^{\omega}(\mathcal{D}_{n_{P}}^{P},\mathcal{D}_{n_{Q}}^{Q}) & =\prod_{i=1}^{n_{P}}p^{\omega}(X_{i}^{P})p^{\omega}\big(\pi_{i}^{P}(X_{i}^{P})\,|\,\mathcal{F}_{i}^{P},X_{i}^{P}\big)p^{\omega}\big(Y_{i}^{P,(\pi_{i}^{P}(X_{i}^{P}))}\,|\,X_{i}^{P},\pi_{i}^{P}(X_{i}^{P})\big)\\
 & \quad\times\prod_{t=1}^{n_{Q}}q^{\omega}(X_{t}^{Q})q^{\omega}\big(\pi_{t}^{Q}(X_{t}^{Q})\,|\,\mathcal{F}_{t}^{Q},X_{t}^{Q}\big)q^{\omega}\big(Y_{t}^{Q,(\pi_{t}^{Q}(X_{t}^{Q}))}\,|\,X_{t}^{Q},\pi_{t}^{Q}(X_{t}^{Q})\big).
\end{align*}
Note that $(P_{X}^{\omega},Q_{X}^{\omega})$ is identical for any $\omega\in\Omega_{m}$, due to the construction of marginal distributions in Step 1. Moreover, the densities involving the policies $p^{\omega}\big(\pi_{i}^{P}(X_{i}^{P})\,|\,\mathcal{F}_{i}^{P},X_{i}^{P}\big)$ and $q^{\omega}\big(\pi_{t}^{Q}(X_{t}^{Q})\,|\,\mathcal{F}_{t}^{Q},X_{t}^{Q}\big)$ are also the same since the source policy is constructed to be independent of $\omega$ and $\pi^{Q}$ is fixed. Therefore, for an arbitrary pair $(\omega,\omega')$ such that $\omega\neq\omega'\in\Omega_{m}$, we have
\begin{align*}
\log\frac{\mathrm{d}\mathbb{P}^{\omega}}{\mathrm{d}\mathbb{P}^{\omega'}}(\mathcal{D}_{n_{P}}^{P},\mathcal{D}_{n_{Q}}^{Q}) & =\sum_{t=1}^{n_{Q}}\log\frac{q^{\omega}\big(Y_{t}^{Q,(\pi_{t}^{Q}(X_{t}^{Q}))}\,|\,X_{t}^{Q},\pi_{t}^{Q}(X_{t}^{Q})\big)}{q^{\omega'}\big(Y_{t}^{Q,(\pi_{t}^{Q}(X_{t}^{Q}))}\,|\,X_{t}^{Q},\pi_{t}^{Q}(X_{t}^{Q})\big)} \\
& \quad +\sum_{i=1}^{n_{P}}\log\frac{p^{\omega}\big(Y_{i}^{P,(\pi_{i}^{P}(X_{i}^{P}))}\,|\,X_{i}^{P},\pi_{i}^{P}(X_{i}^{P})\big)}{p^{\omega'}\big(Y_{i}^{P,(\pi_{i}^{P}(X_{i}^{P}))}\,|\,X_{i}^{P},\pi_{i}^{P}(X_{i}^{P})\big)}.
\end{align*}
Taking the expectation with respect to $\mathbb{P}^{\omega}$ yields
\begin{align}
\mathsf{KL}(\mathbb{P}^{\omega}\,\|\,\mathbb{P}^{\omega'})&=\underbrace{\mathbb{E}^{\omega}\Bigg[\sum_{t=1}^{n_{Q}}\log\frac{q^{\omega}\big(Y_{t}^{Q,(\pi_{t}^{Q}(X_{t}^{Q}))}\,|\,X_{t}^{Q},\pi_{t}^{Q}(X_{t}^{Q})\big)}{q^{\omega'}\big(Y_{t}^{Q,(\pi_{t}^{Q}(X_{t}^{Q}))}\,|\,X_{t}^{Q},\pi_{t}^{Q}(X_{t}^{Q})\big)}\Bigg]}_{=:\,\mathsf{KL}^{Q}} \nonumber \\
& \quad +\underbrace{\mathbb{E}^{\omega}\Bigg[\sum_{i=1}^{n_{P}}\log\frac{p^{\omega}\big(Y_{i}^{P,(\pi_{i}^{P}(X_{i}^{P}))}\,|\,X_{i}^{P},\pi_{i}^{P}(X_{i}^{P})\big)}{p^{\omega'}\big(Y_{i}^{P,(\pi_{i}^{P}(X_{i}^{P}))}\,|\,X_{i}^{P},\pi_{i}^{P}(X_{i}^{P})\big)}\Bigg]}_{=:\,\mathsf{KL}^{P}}.\label{eq:KL-temp}
\end{align}

As a result, it remains to control $\mathsf{KL}^{Q}$ and $\mathsf{KL}^{P}$.
\begin{itemize}
	\item 
Let us start with the first term $\mathsf{KL}^{Q}$. By the tower property, we can simplify
\begin{align}
\mathsf{KL}^{Q} & =\int_{\mathcal{X}}\mathbb{E}^{\omega}\Bigg[\sum_{t=1}^{n_{Q}}\log\frac{q^{\omega}\big(Y_{t}^{Q,(\pi_{t}^{Q}(x))}\,\big|\,x,\pi_{t}^{Q}(x)\big)}{q^{\omega'}\big(Y_{t}^{Q,(\pi_{t}^{Q}(x))}\,\big|\,x,\pi_{t}^{Q}(x)\big)}\,\Big|\,X_{t}^{Q}=x\Bigg]\,\mathrm{d}Q_{X}(x)\nonumber \\
 & =\int_{\mathcal{X}}\mathbb{E}^{\omega}\Bigg[\sum_{t=1}^{n_{Q}}\mathbb{E}^{\omega}\bigg[\log\frac{q^{\omega}\big(Y_{t}^{Q,(\pi_{t}^{Q}(x))}\,\big|\,x,\pi_{t}^{Q}(x)\big)}{q^{\omega'}\big(Y_{t}^{Q,(\pi_{t}^{Q}(x))}\,\big|\,x,\pi_{t}^{Q}(x)\big)}\,\Big|\,X_{t}^{Q}=x,\pi_{t}^{Q}(x)\bigg]\Bigg]\,\mathrm{d}Q_{X}(x)\nonumber \\
 & =\int_{\mathcal{X}}\mathbb{E}^{\omega}\Bigg[\sum_{t=1}^{n_{Q}}\mathsf{KL}\Big(Q_{Y^{(\pi_{t}^{Q}(x))}|X=x}^{\omega}\,\Big\|\,Q_{Y^{(\pi_{t}^{Q}(x))}|X=x}^{\omega'}\Big)\Bigg]\,\mathrm{d}Q_{X}(x)\nonumber \\
 & =\int_{\mathcal{X}}\mathbb{E}^{\omega}\Bigg[\sum_{t=1}^{n_{Q}}\sum_{k=1}^{K}\mathds{1}\{\pi_{t}^{Q}(x)=k\}\mathsf{KL}\Big(Q_{Y^{(k)}|X=x}^{\omega}\,\Big\|\,Q_{Y^{(k)}|X=x}^{\omega'}\Big)\Bigg]\,\mathrm{d}Q_{X}(x)\nonumber \\
 & =\sum_{k=1}^{K}\int_{\mathcal{X}} n_{k}^{\omega,Q}(x)\mathsf{KL}\big(Q_{Y^{(k)}|X=x}^{\omega}\,\|\,Q_{Y^{(k)}|X=x}^{\omega'}\big)\,\mathrm{d}Q_{X}(x).\label{eq:KL-Q-data}
\end{align}
By the construction of reward distributions in Step 1, for any $k\neq 1$, the conditional distribution $Q_{Y^{(k)}|X = x}^{\omega}$ is a Bernoulli distribution with parameter $1/2$ for all $\omega\in\Omega$ and $x\in\mathcal{X}$. As a result, we can compute
\begin{align}
\mathsf{KL}^{Q} & \overset{}{=}\int_{\mathcal{X}}n_{1}^{\omega,Q}(x)\mathsf{KL}\big(Q_{Y^{(1)}|X=x}^{\omega}\,\|\,Q_{Y^{(1)}|X=x}^{\omega'}\big)\,\mathrm{d}Q_{X}(x)\nonumber \\
 & \overset{(\mathrm{i})}{\leq}\int_{\mathcal{X}}n_{Q}\mathsf{KL}\big(Q_{Y^{(1)}|X=x}^{\omega}\,\|\,Q_{Y^{(1)}|X=x}^{\omega'}\big)\,\mathrm{d}Q_{X}(x)\nonumber \\
 & \overset{(\mathrm{ii})}{=}n_{Q}\sum_{i\in[m]:\omega_{i}\neq\omega'_{i}}Q_{X}\big(B(x_{i},r/8)\big)\mathsf{KL}\big(Q_{Y^{(1)}|X=x_{i}}^{\omega}\,\|\,Q_{Y^{(1)}|X=x_{i}}^{\omega'}\big)\nonumber \\
 & \overset{(\mathrm{iii})}{\leq}n_{Q}mw\mathsf{KL}\big(\mathsf{Bern}(1/2+\delta_{\beta})\,\|\,\mathsf{Bern}(1/2-\delta_{\beta})\big)\nonumber \\
 & \overset{(\mathrm{iv})}{\leq}8n_{Q}mw\delta_{\beta}^{2} \nonumber \\
 & \lesssim n_{Q}wr^{2\beta}m.\label{eq:KL-Q-data-UB}
\end{align}
Here, (i) is true because of the simple bound $n_{k}^{\omega,Q}(x)\coloneqq\sum_{t=1}^{n_{Q}}\mathbb{E}^{\omega}[\mathds{1}\{\pi_{t}^{Q}(x)=k\}] \leq n_{Q}$; (ii) is true because
% follows from the choices of support of $Q_{X}$ in \eqref{eq:Q-density-lb} and reward distributions, which implies that 
on the support of $Q_X$, $Q_{Y^{(1)}|X=x}^{\omega}$ and $Q_{Y^{(1)}|X=x}^{\omega'}$ are different only if $x\in\cup_{i\in[m]}B(x_{i},r/8)$; (iii) arises from the density of $Q_X$ in \eqref{eq:Q-density-value} and that the reward distribution conditioned on the covariate is Bernoulli with parameter given in \eqref{eq:reward-func-lb}; (iv) arises from Lemma~\ref{lemma:KL-Bern}; the last step uses $\delta_{\beta}\coloneqq\widetilde{C}_{\beta}r^{\beta}$.
\item
In a similar way, one can also derive
\begin{align}
\mathsf{KL}^{P}=\sum_{k=1}^{K}\int_{\mathcal{X}}n_{k}^{\omega,P}(x)\mathsf{KL}\big(P_{Y^{(k)}|X=x}^{\omega}\,\|\,P_{Y^{(k)}|X=x}^{\omega'}\big)\,\mathrm{d}P_{X}(x).\label{eq:KL-P-data}
\end{align}
According to the construction of $P$, one has
\begin{align}
\mathsf{KL}^{P} &\overset{(\mathrm{i})}{=} \int_{\mathcal{X}}n_{1}^{\omega,P}(x)\mathsf{KL}\big(P_{Y^{(1)}|X=x}^{\omega}\,\|\,P_{Y^{(1)}|X=x}^{\omega'}\big)\,\mathrm{d}P_{X}(x) \nonumber \\
 &\overset{(\mathrm{ii})}{=}\frac{\kappa}{K} n_{P}\sum_{i\in[m]:\omega_{i}\neq\omega'_{i}}P_{X}(B(x_{i},r/8))\mathsf{KL}\big(P_{Y^{(1)}|X=x}^{\omega}\,\|\,P_{Y^{(1)}|X=x}^{\omega'}\big)\nonumber \\
 % & \overset{(\mathrm{ii})}{=}\kappa n_{P}\rho(\omega,\omega')wc_{\gamma}r^{\gamma}\mathsf{KL}\big(P_{Y^{(1)}|X=x_{1}}^{\omega}\,\|\,P_{Y^{(1)}|X=x_{1}}^{\omega'}\big)\nonumber \\
 & \overset{(\mathrm{iii})}{\leq}\frac{\kappa}{K} n_{P}mwc_{\gamma}r^{\gamma}\mathsf{KL}\big(\mathsf{Bern}(1/2+\delta_{\beta})\,\|\,\mathsf{Bern}(1/2-\delta_{\beta})\big)\nonumber \\
 & \overset{(\mathrm{iv})}{\leq}8\frac{\kappa}{K} n_{P}mwc_{\gamma}r^{\gamma}\delta_{\beta}^{2} \nonumber \\
 & \lesssim \kappa n_{P}r^{\gamma}wr^{2\beta}m, \label{eq:KL-P-data-UB}
\end{align}
where (i) is true since $P_{Y^{(k)}|X = x}^{\omega}$ is a Bernoulli distribution with parameter $1/2$ for any $k\neq1$, $x\in\mathcal{X}$, and $\omega\in\Omega$; (ii) holds because on the support of $P_X$, $P_{Y^{(1)}|X=x}^{\omega} \neq P_{Y^{(1)}|X=x}^{\omega'}$ only if $x\in\cup_{i\in[m]}B(x_{i},r/8)$, and $\mu^{\omega,P}(1\,|\,x)=\kappa/K$ for any $x\in\cup_{i\in[m]}B(x_{i},r/8)$, which implies that $n_{1}^{\omega,P}(x)\coloneqq\sum_{i=1}^{n_{P}}\mathbb{E}^{\omega}[\mathds{1}\{\pi_{i}^{P}(x)=1\}]=\kappa n_{P}/K$; (iii) replies on the density of $P_X$ in \eqref{eq:P-density-value} and that the conditional reward distribution is Bernoulli with parameter given in \eqref{eq:reward-func-lb}; (iv) uses Lemma~\ref{lemma:KL-Bern}; the last line is due to $\delta_{\beta}\coloneqq\widetilde{C}_{\beta}r^{\beta}$ and $K \asymp 1$.
\item 
Plugging (\ref{eq:KL-Q-data-UB}) and (\ref{eq:KL-P-data-UB}) into (\ref{eq:KL-temp}) leads to
\begin{align}
\mathsf{KL}(\mathbb{P}^{\omega}\,\|\,\mathbb{P}^{\omega'}) & =\mathsf{KL}^{Q}+\mathsf{KL}^{P}
\lesssim(n_{Q}+\kappa n_{P}r^{\gamma})wr^{2\beta}m\nonumber \\
 & \lesssim (n_{Q}+\kappa n_{P}r^{\gamma})r^{d+2\beta}m\nonumber \\
 & \lesssim (n_{Q}+\kappa n_{P}r^{\gamma})r^{d+2\beta}\log_{2}(M),\label{eq:KL-small}
\end{align}
where we recall $w\defn c_w r^d$, and the last step is due to (\ref{eq:V-G-bound}).
\end{itemize}
Now we set $r=c_{r}(n_{Q}+\kappa n_{P}^{\frac{d+2\beta}{d+2\beta+\gamma}})^{-\frac{1}{d+2\beta}}=o(1)$ for some sufficiently small constant $c_{r}>0$. This gives
\begin{align*}
\mathsf{KL}(\mathbb{P}^{\omega}\,\|\,\mathbb{P}^{\omega'})\ll\log_{2}(M).
\end{align*}

\subsection*{Step 3: applying the standard reduction scheme}

Fix an arbitrary admissible policy $\pi=\{\pi_{t}\}_{t=1}^{n_{Q}}$. By the construction of marginal distribution $Q_{X}$ and random reward distributions, at each time step $t$, regret is incurred if and only $X_{t}^{Q}\in\bigcup_{i=1}^{m}B(x_{i},r/8)$.
Therefore, from \eqref{eq:reward-gap-lb}, we can compute
\begin{align*}
R^{\omega}_{n_{Q}}(\pi) & =\sum_{t=1}^{n_{Q}}\mathbb{E}^{\omega}\Big[f^{\omega}_{\pi^{\omega,\star}(X_{t}^{Q})}(X_{t}^{Q})-f^{\omega}_{\pi_{t}(X_{t}^{Q})}(X_{t}^{Q})\Big]\\
 & \geq\delta_{\beta}\sum_{t=1}^{n_{Q}}\mathbb{E}^{\omega}\Big[\mathds{1}\big\{X_{t}^{Q}\in\cup_{i=1}^{m}B(x_{i},r/8)\big\}\mathds{1}\big\{\pi^{\omega,\star}(X_{t}^{Q})\neq\pi_{t}(X_{t}^{Q})\big\}\Big].
\end{align*}
Therefore, for any $\omega\neq\omega'\in\Omega_{m}$, we can lower bound
\begin{align}
R_{n_{Q}}^{\omega}&(\pi)+R_{n_{Q}}^{\omega'}(\pi) \nonumber \\
& \geq\delta_{\beta}\sum_{t=1}^{n_{Q}}\sum_{i=1}^{m}\mathbb{E}^{\omega}\Big[\mathds{1}\big\{X_{t}^{Q}\in B(x_{i},r/8)\big\}\mathds{1}\big\{\pi^{\omega,\star}(X_{t}^{Q})\neq\pi_{t}(X_{t}^{Q})\big\}\Big]\nonumber \\
 & \quad+\delta_{\beta}\sum_{t=1}^{n_{Q}}\sum_{i=1}^{m}\mathbb{E}^{\omega'}\Big[\mathds{1}\big\{X_{t}^{Q}\in B(x_{i},r/8)\big\}\mathds{1}\big\{\pi^{\omega',\star}(X_{t}^{Q})\neq\pi_{t}(X_{t}^{Q})\big\}\Big]\nonumber \\
 % & =\delta_{\beta}\sum_{t=1}^{n_{Q}}\sum_{i=1}^{m}\mathbb{E}_{X_{t}^{Q}\sim Q_{X}}\Big[\mathds{1}\big\{X_{t}^{Q}\in B(x_{i},r/8)\big\} \nonumber \\
 % & \qquad \qquad \cdot\big(\mathds{1}\big\{\pi^{\omega,\star}(X_{t}^{Q})\neq\pi_{t}(X_{t}^{Q})\big\}+\mathds{1}\big\{\pi^{\omega',\star}(X_{t}^{Q})\neq\pi_{t}(X_{t}^{Q})\big\}\big)\Big]\nonumber \\
 & \overset{(\mathrm{i})}{\geq}\delta_{\beta}\sum_{i=1}^{m}\sum_{t=1}^{n_{Q}}\mathbb{E}^{\omega}\Big[\mathds{1}\big\{X_{t}^{Q}\in B(x_{i},r/8)\big\}\Big]\mathds{1}\{\omega_{i}\neq\omega_{i}'\}\nonumber \\
 & =\delta_{\beta}n_{Q}Q_{X}(B(x_{1},r/8))\rho(\omega,\omega') \nonumber \\
 & \overset{(\mathrm{ii})}{\geq} 2^{-3}\delta_{\beta}n_{Q}wm \nonumber \\
 & \overset{(\mathrm{iii})}{\geq} 2^{-3}\widetilde{C}_{\beta}c_{w}c_{m}r^{(1+\alpha)\beta}n_{Q}\nonumber \\
 & \gtrsim n_{Q}\Big(n_{Q}+(\kappa n_{P})^{\frac{d+2\beta}{d+2\beta+\gamma}}\Big)^{-\frac{(1+\alpha)\beta}{d+2\beta}}.\label{eq:reg-diff-omega-lb}
\end{align}
Here, (i) holds because if $\omega_{i}\neq\omega_{i}'$ for some $i\in[m]$ and $X_{t}^{Q}\in B(x_{i},r/8)$, then one has $\pi^{\omega,\star}(X_{t}^{Q})\neq\pi^{\omega',\star}(X_{t}^{Q}$), which further implies that
\begin{align*}
\mathds{1}\big\{\pi^{\omega,\star}(X_{t}^{Q})\neq\pi_{t}(X_{t}^{Q})\big\}+\mathds{1}\big\{\pi^{\omega',\star}(X_{t}^{Q})\neq\pi_{t}(X_{t}^{Q})\big\}\geq\mathds{1}\{\omega_{i}\neq\omega_{i}'\};
\end{align*}
(ii) is due to (\ref{eq:V-G-bound}) and (\ref{eq:Q-density-value}); (iii) uses $\delta_{\beta}\defn\widetilde{C}_{\beta}r^\beta$, $w\defn c_w r^d$, and $m\defn\lfloor c_{m}r^{\alpha\beta-d}\rfloor$.

Putting (\ref{eq:KL-small}) and (\ref{eq:reg-diff-omega-lb}) together, we can readily invoke \citet[Theorem 2.5]{tsybakov2009introduction} to obtain that
\begin{align*}
\inf_{\pi}\sup_{(P,Q)\in\Pi(K,\beta,\alpha,\gamma,\kappa,l_{0},b)}\mathbb{E}\big[R_{n_{Q}}(\pi)\big]\gtrsim n_{Q}\Big(n_{Q}+(\kappa n_{P})^{\frac{d+2\beta}{d+2\beta+\gamma}}\Big)^{-\frac{(1+\alpha)\beta}{d+2\beta}}.
\end{align*}

\section{Proof of Theorem \ref{thm:upper-bound-adaptive}}

\label{sec:Proof-of-Adaptivity}

We start by presenting Lemma~\ref{lemma:smooth-est-error} below, which illustrates that the smoothness estimate $\widehat{\beta}$ returned by Procedure~\ref{alg:smoothness} in Algorithm~\ref{alg:UCB-TL-adaptive} is close to the ground truth $\beta$ with high probability. The proof can be found in Appendix~\ref{proof-lemma:smooth-est-error}.

\begin{lemma}
\label{lemma:smooth-est-error}
	Instate the assumptions of Theorem~\ref{thm:upper-bound-adaptive}. Define the event
\begin{align}
\mathcal{E}_{\se}\coloneqq\bigg\{\beta-\frac{C_{3}(d+2\overline{\beta}+\overline{\gamma})^2 \log_{2}(\log(n))}{\underline{\beta}\log_{2}(n)}\leq\widehat{\beta}\leq\beta\bigg\}.\label{eq:event-beta-est-good-event}
\end{align}
for some constant $C_{3}$ independent of $n_Q$ and $n_{P}$, where we recall $n\defn n_{P} \vee n_{Q}$. Then one has $\mathbb{P}\{\mathcal{E}_{\se}\}\geq1- O(n^{-10})$.
\end{lemma}
Before delving into the details of controlling the regret, we pause to make some remarks. 
To begin with, when the source sample size is relatively large, we split the source dataset and only use the subset $\mathcal{D}^{P}_{\mathsf{dm}} \subset \cD^P$ to assist in learning the target $Q$-bandit in Algorithm~\ref{alg:UCB-TL-adaptive}. By the choice of $s_{P}$ in the smoothness estimation stage (Procedure \ref{alg:smoothness}), we always have $|\mathcal{D}^{P}_{\mathsf{dm}}| = n_{P}-s_{P}=(1-o(1))n_{P}$. For simplicity of presentation, we drop the notation $\mathcal{D}^{P}_{\mathsf{dm}}$ in $n_{k}^{P}(B;\mathcal{D}^{P}_{\mathsf{dm}})$, $\overline{Y}_{k}^{P}(B;\mathcal{D}^{P}_{\mathsf{dm}})$, $\widehat{U}_{k}(\tau,B;\mathcal{D}^{P}_{\mathsf{dm}})$, and $\widehat{\tau}_{k}^{\star}(B;\mathcal{D}^{P}_{\mathsf{dm}})$ in the remainder of this proof.
Second, when the source sample size is relatively small, Procedure \ref{alg:smoothness} chooses arms uniformly at random for $1\leq t < s_{Q}$, in order to collect samples to estimate the smoothness. The regret collected during this stage can be simply upper bounded by $s_{Q}$. Therefore, in what follows, we shall focus on bounding the regret incurred after time $s_{Q}$.
Finally, we note that the event $\cE_\se$ is independent of the source data subset $\mathcal{D}^{P}_{\mathsf{dm}}$ and the $Q$-samples collected after time $s_{Q}$, i.e.~$\{(X_t^Q,\pi^\adp_t,Y^{Q,(\pi^\adp_t)}_t)\}_{t\geq s_Q}$.

Now, let us proceed to prove the regret bound. Similar to the proof of Theorem~\ref{thm:upper-bound} in Appendix~\ref{sec:Proof-of-Upper-Bound}, for any bin $B$, let us define
\begin{align}
R^{\tree,\adp}(B) & \defn\sum_{t=s_{Q}}^{n_{Q}}\big(f_{\pi^{\star}(X_{t}^{Q})}(X_{t}^{Q})-f_{\pi^{\adp}_{t}(X_{t}^{Q})}(X_{t}^{Q})\big)\mathds{1}\{X_{t}^{Q}\in B\} \mathds{1}\{B\in\mathcal{T}^{\adp}_{t}\},\label{eq:R-nonleaf-B-adp} \\
R^{\lf,\adp}(B) & \defn\sum_{t=s_{Q}}^{n_{Q}}\big(f_{\pi^{\star}(X_{t}^{Q})}(X_{t}^{Q})-f_{\pi^{\adp}_{t}(X_{t}^{Q})}(X_{t}^{Q})\big)\mathds{1}\{X_{t}^{Q}\in B\} \mathds{1}\{B\in\mathcal{L}^{\adp}_{t}\},\label{eq:R-leaf-B-adp}
\end{align}
where $\mathcal{T}^{\adp}_{t}$ (resp.~$\mathcal{L}^{\adp}_{t}$) denotes the partition tree (resp.~partition of the covariate space / set of the leaf nodes of $\mathcal{T}^{\adp}_{t}$) generated by Algorithm~\ref{alg:UCB-TL-adaptive} at time $t\geq s_{Q}$.
Also, for any bin $B$, we define the sets:
\begin{align}
\underline{\mathcal{I}}_{B}^\adp & \defn\big\{ k\in[K]:\sup_{x\in B}\{f_{(1)}(x)-f_{k}(x)\}\leq \olc_{0}|B|^{\hatbeta}\big\}, \label{eq:arm_set_lb-adp}\\
\overline{\mathcal{I}}_{B}^\adp & \defn\big\{ k\in[K]:\sup_{x\in B}\{f_{(1)}(x)-f_{k}(x)\}\leq8\olc_{0}|B|^{\hatbeta}\big\}. \label{eq:arm_set_ub-adp}
\end{align}
Here, we define $\olc_{0}\defn2\overline{C}_{\beta}$ where $\overline{C}_{\beta}$ is the upper bound for $C_{\beta}$. In addition, for any bin $B$, we denote by $\mathcal{I}_{B}^\adp$ the set of active arms in bin $B$ at the end of rounds before $B$ is replaced by its children in Algorithm~\ref{alg:UCB-TL-adaptive}, and we define the events
\begin{align}
\label{eq:good-event-B-adp}
\mathcal{E}_{B}^{\adp}\defn\{\underline{\mathcal{I}}_{B}^\adp\subset\mathcal{I}_{B}^\adp\subset\overline{\mathcal{I}}_{B}^\adp\} \quad \text{and}\quad \mathcal{G}^{\adp}_{B}\defn\bigcap_{B'\in\mathsf{anc}(B)}\mathcal{E}_{B'}^\adp,
\end{align}
Observe that the regret accumulated after time $s_{Q}$ is given by $R^{\tree,\adp}(\mathcal{X})$ since $X_{t}^{Q}\in\mathcal{X}$ and $\mathcal{X}\in\mathcal{T}_{t}$ for any $t\geq s_{Q}$. As a result, we shall focus on $R^{\tree,\adp}(\mathcal{X})$, which can be decomposed as
\begin{align}
R^{\tree,\adp}(\mathcal{X}) &=  \sum_{0\leq i< l^{\adp}}\underbrace{\sum_{B\in\mathcal{B}_{i}}R^{\tree,\adp}(B)\mathds{1}\{(\mathcal{E}_{B}^{\adp})^{c}\cap\mathcal{G}^{\adp}_{B}\}}_{=:\,R^{\tree,\adp}_{i}}
+\sum_{0\leq i< l^{\adp}}\underbrace{\sum_{B\in\mathcal{B}_{i}}R^{\lf,\adp}(B)\mathds{1}\{\mathcal{E}^{\adp}_{B}\cap\mathcal{G}^{\adp}_{B}\}}_{=:\,R^{\lf,\adp}_{i}} \nonumber \\
&\quad +\underbrace{\sum_{B\in\mathcal{B}_{l^{\adp}}}R^{\tree,\adp}(B)\mathds{1}\{\mathcal{G}^{\adp}_{B}\}}_{=:\,R^{\adp}_{l^{\adp}}},\label{eq:regret-adp-decomp}
\end{align}
where $l^{\adp}\in\mathbb{N}$ is set to be
\begin{align}
\label{eq:max_depth-adp}
    l^{\adp} \defn \bigg\lceil {\frac{1}{d+2\hatbeta}}\log_{2}\Big(\frac{ n_{Q}}{K}\Big) \bigg\rceil \vee \bigg\lceil {\frac{1}{d+2\hatbeta+\gamma}\log_{2} \Big(\frac{c^{\adp}\kappa n_{P}}{K}\Big)}\bigg\rceil,
\end{align} 
with $c^{\adp}\defn \frac{1}{8}\underline{q}c_{\gamma}\overline C_{\beta}^2$. Additionally, we define $l_{4}^\adp \coloneqq \big\lceil
\frac{1}{d+\gamma} \log_{2} \big( \frac{\underline{q}c_{\gamma}\kappa n_{P}}{80K\log(n_{P})} \big)\big\rceil$, and $l_{5}^\adp\defn\big\lceil \frac{1}{d+2\hatbeta+\gamma}\log(\frac{c^{\adp}\kappa n_{P}}{K})\big\rceil$.

The three terms in (\ref{eq:regret-adp-decomp}) can be controlled individually, which are presented in
Lemmas \ref{lemma:non-leaf-upper-adp-born}--\ref{lemma:leaf-upper-adp} below.
\begin{lemma}\label{lemma:non-leaf-upper-adp-born}
For any $i \geq 0$, one has
\begin{align}
\label{eq:regret-non-leaf-adp-born-i-large}
 \mathbb{E}[R^{\tree,\adp}_{i}\,|\,\hatbeta] \lesssim  K 2^{(d+2\widehat{\beta}-(1+\alpha)\hatbeta) i}\log^{+}\big(n_{Q}2^{-(d+2\hatbeta)i}\big).
\end{align}
Further, for any $0\leq i<l_{4}^{\adp} \wedge l_{5}^\adp$, we also have
\begin{align}
\mathbb{E}[R^{\tree,\adp}_{i}\,|\,\hatbeta] \lesssim
& Kn_{Q}2^{-(1+\alpha)\hatbeta i}\exp\left(-\frac{c^{\adp}\kappa}{K}n_{P}|B|^{d+2\hatbeta+\gamma}\right) \nonumber \\
& +Kn_{Q}2^{-(1+\alpha)\hatbeta i} \bigg(\frac{1}{n}2^{(d+2\hatbeta)i}+ \frac{1}{n_{P}^{5}}\bigg).\label{eq:regret-non-leaf-adp-born-i-small}
\end{align}
\end{lemma}

\begin{proof}
    See Appendix \ref{subsec:Proof-of-lemma:non-leaf-upper-adp-born}.
\end{proof}

\begin{lemma}\label{lemma:non-leaf-upper-adp-live}
For any $i \geq 0$, we have
\begin{align}
\label{eq:regret-non-leaf-adp-live-i-large}
 \mathbb{E}[R^{\lf,\adp}_{i}\,|\,\hatbeta] \lesssim   K 2^{(d+2\widehat{\beta}-(1+\alpha)\hatbeta) i}\log^{+}\big(n_{Q}2^{-(d+2\hatbeta)i}\big).
\end{align}
Moreover, the following holds for any $0\leq i<l_{4}^{\adp} \wedge l_{5}^{\adp}$:
\begin{align}
\mathbb{E}[R^{\lf,\adp}_{i}\,|\,\hatbeta] \lesssim
& Kn_{Q}2^{-(1+\alpha)\hatbeta i}\exp\left(-\frac{c^{\adp}\kappa}{K}n_{P}2^{-(d+2\hatbeta+\gamma)i}\right) \nonumber \\
& \quad + \frac{K}{n^{5}_P} 2^{(d+2\hatbeta-(1+\alpha)\hatbeta)i} \log^{+}\big(n_{Q}2^{-(d+2\hatbeta)i}\big). \label{eq:regret-non-leaf-adp-live-i-small}
\end{align}
\end{lemma}

\begin{proof}
    See Appendix \ref{sec:Proof-of-lemma:non-leaf-upper-adp-live}.
\end{proof}

\begin{lemma}\label{lemma:leaf-upper-adp}The regret incurred
on the nodes of depth $l^{\adp}$ obeys
\begin{align}
\mathbb{E}[R^{\adp}_{l^{\adp}}\,|\,\hatbeta]\lesssim n_{Q}2^{-  (1+\alpha)\hatbeta l^{\adp}}.\label{eq:regret-leaf-adp}
\end{align}
\end{lemma}

\begin{proof}
    See Appendix \ref{subsec:Proof-of-lemma:leaf-upper-adp}.
\end{proof}

Given Lemmas~\ref{lemma:non-leaf-upper-adp-born}--\ref{lemma:leaf-upper-adp}, we start to upper bound the regret.
Substituting (\ref{eq:regret-non-leaf-adp-born-i-large}), (\ref{eq:regret-non-leaf-adp-live-i-large}), and (\ref{eq:regret-leaf-adp}) into (\ref{eq:regret-adp-decomp}) shows that
\begin{align}
 \mathbb{E} \big[R^{\tree,\adp}(\mathcal{X})\,|\,\hatbeta\big] &= \sum_{0\leq i< l^{\adp}}\mathbb{E}[R^{\tree,\adp}_{i}\,|\,\hatbeta] +\sum_{0\leq i< l^{\adp}}\mathbb{E}[R^{\lf,\adp}_{i}\,|\,\hatbeta] + \mathbb{E}[R_{ l^{\adp}}\,|\,\hatbeta] \nonumber\\
 & \lesssim \sum_{0 \leq i< l^{\adp}}K2^{(d+2\hatbeta-(1+\alpha)\hatbeta )i}\log^{+}\big(n_{Q}2^{-(d+2\hatbeta)i}\big) + n_{Q}2^{- (1+\alpha)\hatbeta  l^{\adp}}. \label{eq:regret-adp-ub-crude-temp}
\end{align} 
Meanwhile, we can also plug (\ref{eq:regret-non-leaf-adp-born-i-large})--(\ref{eq:regret-leaf-adp}) into (\ref{eq:regret-adp-decomp}) to obtain the following more refined upper bound:
\begin{align}
 \mathbb{E} \big[R^{\tree,\adp}(\mathcal{X})\,|\,\hatbeta\big] & =\sum_{0\leq i< l^{\adp}}\mathbb{E}[R^{\tree,\adp}_{i}\,|\,\hatbeta] +\sum_{0\leq i< l^{\adp}}\mathbb{E}[R^{\lf,\adp}_{i}] + \mathbb{E}[R_{ l^{\adp}}\,|\,\hatbeta] \nonumber\\
 & \lesssim \sum_{0\leq i< l_{4}^{\adp}\wedge l_{5}^{\adp} \wedge l^{\adp}} Kn_{Q}2^{-(1+\alpha)\hatbeta i} \exp\left(-\frac{c^{\adp}\kappa}{K}n_{P}2^{-(d+2\hatbeta+\gamma)i}\right) \nonumber \\
 & \quad+\sum_{0\leq i< l_{4}^{\adp}\wedge l_{5}^{\adp} \wedge l^{\adp}} Kn_{Q}2^{-(1+\alpha)\hatbeta i} \bigg(\frac{1}{n}2^{(d+2\hatbeta)i}+ \frac{1}{n_{P}^{5}}\bigg) \nonumber\\
 & \quad + \sum_{0\leq i< l_{4}^{\adp}\wedge l_{5}^{\adp} \wedge l^{\adp}}\frac{K}{n^{5}_P}2^{(d+2\hatbeta-(1+\alpha)\hatbeta)i}\log^{+}\big(n_{Q}2^{-(d+2\hatbeta)i}\big) \nonumber \\
 & \quad + \sum_{l_{4}^{\adp}\wedge l_{5}^{\adp} \wedge l^{\adp} \leq i< l^{\adp}}K2^{(d+2\hatbeta-(1+\alpha)\hatbeta )i}\log^{+}\big(n_{Q}2^{-(d+2\hatbeta)i}\big) + n_{Q}2^{- (1+\alpha)\hatbeta  l^{\adp}}. \label{eq:regret-oracle-ub-temp-adp}
\end{align}
Next, let us control the regret according to the relationship between $n_{P}$ and $n_{Q}$. 

\begin{itemize}
\item 
Let us first consider the scenario $\frac{n_Q}{K} \geq (\frac{c^{\adp}\kappa n_{P}}{K})^{\frac{d+2\hatbeta}{d+2\hatbeta+\gamma}}$, where we know $l^{\adp}=\big\lceil \frac{1}{d+2\hatbeta} \log_{2}(\frac{n_{Q}}{K})\big\rceil$ by (\ref{eq:max_depth-adp}).
Combining this with the condition $\alpha\beta \leq d$ and Lemma~\ref{lemma:series-1} yields
\begin{align}
\label{eq:sum-temp1-adp}
    \sum_{0 \leq i< l^{\adp}} & 2^{(d+2\hatbeta- (1+\alpha)\hatbeta )i}\log^{+}\big(n_{Q}2^{-(d+2\hatbeta)i}\big) \lesssim 2^{(d + 2\hatbeta - (1+\alpha)\hatbeta )l^{\adp}}.
\end{align}
As a consequence, it follows from (\ref{eq:regret-adp-ub-crude-temp}) that
\begin{align}
 \mathbb{E} \big[R^{\tree,\adp}(\mathcal{X})\,|\,\hatbeta\big] & \lesssim K 2^{(d + 2\hatbeta - (1+\alpha)\hatbeta )l^{\adp}} + n_{Q}2^{-(1+\alpha)\hatbeta  l^{\adp}} \nonumber \\
& \lesssim n_{Q} \Big(\frac{n_{Q}}{K}\Big)^{-\frac{(1+\alpha)\hatbeta }{d+2\hatbeta}} \nonumber \\
& \lesssim  n_{Q}^{1-\frac{(1+\alpha)\beta }{d+2\beta}} \log^{C_{1}}(n) \label{eq:regret-ub-nP-small-temp}
\end{align}
for some constant $C_{1}>0$ independent of $n_{Q}$ and $n_{P}$. Here, the second line follows from $l^{\adp}=\big\lceil \frac{1}{d+2\hatbeta} \log_{2} (\frac{n_{Q}}{K})\big\rceil$, and the last step follows from the error bound of $\widehat{\beta}$ on the event $\cE_\se$ in (\ref{eq:event-beta-est-good-event}) and $K \asymp 1$.
Combined with the regret incurred in the smoothness estimation stage, we find that
\begin{align}
\bE [R_{n_{Q}}(\pi_{\mathsf{a}})] & = \mathbb{E}\Bigg[\sum_{t=1}^{s_{Q}-1}\big(f^{\star}(X_{t}^{Q})-f_{\pi_{t}^\adp}(X_{t}^{Q})\big)\Bigg] \nonumber\\
 & \quad+\mathbb{E}\Bigg[\sum_{t=s_{Q}}^{n_{Q}}\big(f^{\star}(X_{t}^{Q})-f_{\pi_{t}^\adp}(X_{t}^{Q})\big) \,\Big|\, \mathcal{E}_{\se}\Bigg]\bP\{\mathcal{E}_{\se}\}\nonumber \\
 & \quad+\mathbb{E}\Bigg[\sum_{t=s_{Q}}^{n_{Q}}\big(f^{\star}(X_{t}^{Q})-f_{\pi_{t}^\adp}(X_{t}^{Q})\big) \,\Big|\, \mathcal{E}_{\se}^{c}\Bigg]\bP\{\mathcal{E}_{\se}^{c}\}\nonumber \\
 & \leq s_{Q}+\mathbb{E} \big[R^{\tree,\adp}(\mathcal{X})\,|\, \mathcal{E}_{\se}\big]+n_{Q}\mathbb{P}\{\cE_\se^c\}\nonumber \\
 & \leq n_{Q}^{\frac{\underline{\beta}}{d+2\overline{\beta}}}+n_{Q}^{1-\frac{(1+\alpha)\beta}{d+2\beta}}\log^{C_{1}}(n)+n_{Q}O(n^{-10})\nonumber \\
 & \asymp n_{Q}^{1-\frac{(1+\alpha)\beta}{d+2\beta}}\log^{C_{1}}(n),\label{eq:regret-ub-nP-small}
\end{align}
where the last step arises from the conditions $\alpha\beta\leq d$ and $\underline{\beta} \leq \beta \leq \overline{\beta}$.

\item 
Next, let us turn to the case $\frac{n_Q}{K}< (\frac{c^{\adp}\kappa n_{P}}{K})^{\frac{d+2\hatbeta}{d+2\hatbeta+\gamma}}$, where $l^{\adp} = \big\lceil \frac{1}{d+2\hatbeta+\gamma}\log_{2}(\frac{c^{\adp}\kappa n_{P}}{K})\big\rceil$ from (\ref{eq:max_depth-adp}).
% , and we shall apply (\ref{eq:regret-oracle-ub-temp}) to bound the regret. As a reminder, we define $l_{4} \coloneqq \big\lceil
% \frac{1}{d+\gamma} \log_{2} \big( \frac{\underline{q}c_{\gamma}\kappa n_{P}}{80K\log(n_{P})} \big)\big\rceil$ and $l_{5}\defn\big\lceil \frac{1}{d+2\hatbeta+\gamma}\log_{2}(\frac{c^{\adp}\kappa n_{P}}{K})\big\rceil$.
%
On the event $\cE_\se$, we know that $\hatbeta \geq \beta - o(1) > 0$, and consequently, $l^{\adp} = l_{5}^\adp \leq l_{4}^\adp$ for sufficiently large $n_{P}$. Therefore, we can apply (\ref{eq:regret-oracle-ub-temp-adp}) to obtain
\begin{align}
\mathbb{E} \big[R^{\tree,\adp}(\mathcal{X})\,|\,\hatbeta\big]
 & \lesssim \sum_{0\leq i< l^{\adp}} Kn_{Q}2^{-(1+\alpha)\hatbeta i} \exp\left(-\frac{c^{\adp}\kappa}{K}n_{P}2^{-(d+2\hatbeta+\gamma)i}\right) \nonumber \\
 & \quad+\sum_{0\leq i< l^{\adp}} Kn_{Q}2^{-(1+\alpha)\hatbeta i} \bigg(\frac{1}{n}2^{(d+2\hatbeta)i}+ \frac{1}{n_{P}^{5}}\bigg) \nonumber\\
 & \quad + \sum_{0\leq i< l^{\adp}}\frac{K}{n^{5}_P}2^{(d+2\hatbeta-(1+\alpha)\hatbeta)i}\log^{+}\big(n_{Q}2^{-(d+2\hatbeta)i}\big)+ n_{Q}2^{- (1+\alpha)\hatbeta  l^{\adp}}. \nonumber \\
 & \overset{(\mathrm{i})}{\lesssim} K n_{Q} 2^{-(1+\alpha)\hatbeta l^{\adp}}
  + K n_{Q} 2^{-(1+\alpha)\hatbeta l^{\adp}} \frac{1}{n}2^{(d+2\hatbeta)l^\adp} \nonumber \\
  & \quad + \frac{K}{n_{P}^{5}} 2^{-(1+\alpha)\hatbeta l^{\adp}}  2^{(d+2\hatbeta)l^{\adp}} \nonumber\\
 & \overset{(\mathrm{ii})}{\asymp} K n_{Q} 2^{-(1+\alpha)\hatbeta l^{\adp}} \asymp K n_{Q} \Big(\frac{c^\adp \kappa n_{P}}{K}\Big)^{-\frac{ (1+\alpha)\hatbeta }{d+2\hatbeta+\gamma}} \nonumber\\
 & \lesssim n_{Q} (\kappa n_{P})^{-\frac{ (1+\alpha)\beta }{d+2\beta+\gamma}} \log^{C_{2}}(n),
 \label{eq:regret-ub-nP-large-temp}
\end{align}
for some constant $C_2 > 0$ independent of $n_P$ and $n_Q$.
(i) arises from Lemma~\ref{lemma:series-2} and $l^{\adp} = \big\lceil \frac{1}{d+2\hatbeta+\gamma}\log_{2}(\frac{c^{\adp}\kappa n_{P}}{K})\big\rceil$, $d \geq \alpha\beta$, and (\ref{eq:sum-temp1-adp}); (ii) uses $2^{(d+\hatbeta)l^\adp}=o(n_{P})$ by the choice of $l^{\adp}$; \eqref{eq:regret-ub-nP-large-temp} follows from the estimation error of $\widehat{\beta}$ on the event $\cE_\se$ in (\ref{eq:event-beta-est-good-event}) and $K\asymp 1$.
In addition, as $K,\kappa\asymp1$, we have $n_{P}\geq n_{Q}$ for sufficiently large $n_{P}$, and hence $s_{Q}=1$ according to Procedure~\ref{alg:smoothness}.
Therefore, we find that
\begin{align}
\bE \big[R_{n_{Q}}(\pi_{\mathsf{a}})\big] & =\mathbb{E}\Bigg[\sum_{t=s_{Q}}^{n_{Q}}\big(f^{\star}(X_{t}^{Q})-f_{\pi_{t}^\adp}(X_{t}^{Q})\big) \,\Big|\, \mathcal{E}_{\se}\Bigg]\bP\{\mathcal{E}_{\se}\}\nonumber \\
 & \quad+\mathbb{E}\Bigg[\sum_{t=s_{Q}}^{n_{Q}}\big(f^{\star}(X_{t}^{Q})-f_{\pi_{t}^\adp}(X_{t}^{Q})\big) \,\Big|\, \mathcal{E}_{\se}^{c}\Bigg]\bP\{\mathcal{E}_{\se}^{c}\}\nonumber \\
 & \leq \mathbb{E} \big[R^{\tree,\adp}(\mathcal{X})\,|\, \mathcal{E}_{\se}\big]+n_{Q}\mathbb{P}\{\cE_\se^c\}\nonumber \\
 & \leq n_{Q} (\kappa n_{P})^{-\frac{ (1+\alpha)\beta }{d+2\beta+\gamma}} \log^{C_{2}}(n)+n_{Q}O(n^{-10})\nonumber \\
 & \asymp n_{Q} (\kappa n_{P})^{-\frac{ (1+\alpha)\beta }{d+2\beta+\gamma}} \log^{C_{2}}(n).\label{eq:regret-ub-nP-large}
\end{align}
where the last step holds due to the condition $d \geq \alpha \beta$ and $\kappa \asymp 1$.
\item
Combining these two cases above, we conclude that
\begin{align*}
\mathbb{E}\big[R_{n_{Q}}(\pi^{\adp})\big]\lesssim n_{Q}\big( n_{Q} + ( \kappa n_{P})^{\frac{d+2\beta}{d+2\beta+\gamma}} \big)^{-\frac{ (1+\alpha)\beta }{d+2\beta}} \log^{C}(n),
\end{align*}
for some constant $C>0$ independent of $n_{Q}$ and $n_{P}$. This finishes the proof of Theorem~\ref{thm:upper-bound-adaptive}.

\end{itemize}

\subsection{Proof of Lemma \ref{lemma:non-leaf-upper-adp-born}}
\label{subsec:Proof-of-lemma:non-leaf-upper-adp-born}

As the event $\cE_\se$ is independent of $\{(X_t^Q,\pi^\adp_t,Y^{Q,(\pi^\adp_t)}_t)\}_{t\geq s_Q}$, we abuse the notations by using $\bP\{\cdot\}$ and $\bE[\cdot]$ to represent the probability and expectation conditioned on $\hatbeta$ for simplicity of presentation. We shall apply a similar argument as in the proof of Lemma~\ref{lemma:non-leaf-upper-oracle-born} in Appendix~\ref{subsec:Proof-of-lemma:non-leaf-upper-oracle-born} to prove Lemma \ref{lemma:non-leaf-upper-adp-born}.

Fix an arbitrary bin $B \in \mathcal{B}_{i}$ such that $0 \leq i < l^{\adp}$ and $Q_{X}(B) > 0$.
On the event $\mathcal{G}_{B}^\adp$, one has $\mathcal{I}_{\mathsf{p}(B)}^\adp\subset\overline{\mathcal{I}}_{\mathsf{p}(B)}^\adp$.
Recall the definition of $\overline{\mathcal{I}}_{\mathsf{p}(B)}^\adp$ in \eqref{eq:arm_set_ub-adp}. This means that for any $x\in\mathsf{p}(B)$ and $k\in\mathcal{I}_{\mathsf{p}(B)}^\adp$, we have
\begin{align*}
f_{(1)}(x)-f_{k}(x)\leq8\olc_{0}|\mathsf{p}(B)|^{\hatbeta}=2^{3+\hatbeta}\olc_{0}|B|^{\hatbeta} =: \olc_{1} |B|^{\hatbeta}.
\end{align*}
It follows that for any arm $k\in\mathcal{I}_{\mathsf{p}(B)}$ and $x\in B$,
\begin{align}
\big(f_{(1)}(x)-f_{k}(x)\big)\mathds{1}\{\mathcal{G}_{B}^\adp\}\leq \olc_{1}|B|^{\hatbeta}\mathds{1}\{0<f_{(1)}(x)-f_{(2)}(x)\leq \olc_{1}|B|^{\hatbeta}\}\mathds{1}\{\mathcal{G}_{B}^\adp\}.\label{eq:reward-gap-event-ancestor-adp}
\end{align}
Therefore, we can upper bound
\begin{align*}
& \mathbb{E}[R^{\tree,\adp}(B)\mathds{1}\{(\mathcal{E}^{\adp}_{B})^{c}\cap\mathcal{G}^{\adp}_{B}\}]\\
& \quad=\mathbb{E}\bigg[\mathds{1}\{(\mathcal{E}^{\adp}_{B})^{c}\cap\mathcal{G}^{\adp}_{B}\}\sum_{t=s_{Q}}^{n_{Q}}\big(f_{\pi^{\star}(X_{t}^{Q})}(X_{t}^{Q})-f_{\pi^{\adp}_{t}(X_{t}^{Q})}(X_{t}^{Q})\big)\mathds{1}\{X_{t}^{Q}\in B\}\mathds{1}\{B\in\mathcal{T}^{\adp}_{t}\}\Big]\\
& \quad\leq\mathbb{E}\bigg[\mathds{1}\{(\mathcal{E}^{\adp}_{B})^{c}\cap\mathcal{G}^{\adp}_{B}\}\sum_{t=s_{Q}}^{n_{Q}} \olc_{1}|B|^{\hatbeta}\mathds{1}\{0<f_{(1)}(X_{t}^{Q})-f_{(2)}(X_{t}^{Q})\leq \olc_{1}|B|^{\hatbeta}\}\bigg]\\
& \quad\lesssim n_{Q}|B|^{\hatbeta}Q_{X}(0<f_{(1)}(X)-f_{(2)}(X)\leq \olc_{1}|B|^{\hatbeta},X\in B)\mathbb{P}\{(\mathcal{E}^{\adp}_{B})^{c}\cap\mathcal{G}^{\adp}_{B}\}.
\end{align*}
This further leads to
\begin{align}
\mathbb{E}[R^{\tree,\adp}_{i}] & =\sum_{B\in\mathcal{B}_{i}}\mathbb{E}[R^{\tree,\adp}(B)\mathds{1}\{(\mathcal{E}^{\adp}_{B})^{c}\cap\mathcal{G}^{\adp}_{B}\}] \nonumber\\
 & \lesssim n_{Q}2^{-\hatbeta i}\max_{B\in\mathcal{B}_{i}}\mathbb{P}\{(\mathcal{E}^{\adp}_{B})^{c}\cap\mathcal{G}^{\adp}_{B}\}\sum_{B\in\mathcal{B}_{i}}Q_{X}(0<f_{(1)}(X)-f_{(2)}(X)\leq \olc_{1}|B|^{\hatbeta},X\in B).\label{eq:regret-nonleaf-good-event-temp1-adp}
\end{align}
We can use the margin condition to upper bound
\begin{align}
\sum_{B\in\mathcal{B}_{i}}  Q_{X}(0<f_{(1)}(X)-f_{(2)}(X)\leq \olc_{1}|B|^{\hatbeta},X\in B)
 & = Q_{X}(0<f_{(1)}(X)-f_{(2)}(X)\leq \olc_{1}|B|^{\hatbeta})\nonumber \\
 & \leq C_{\alpha }( \olc_{1}|B|^{\beta})^{\alpha} \lesssim |B|^{\alpha \beta} = 2^{-\alpha\beta i}.\label{eq:prob-Q-gap-ub-adp}
\end{align}
Putting(\ref{eq:regret-nonleaf-good-event-temp1-adp}) and (\ref{eq:prob-Q-gap-ub-adp}) together yields
\begin{align}
\mathbb{E}[R^{\tree,\adp}_{i}]\lesssim n_{Q}2^{- (1+\alpha)\hatbeta i}\max_{B\in\mathcal{B}_{i}}\mathbb{P}\{(\mathcal{E}^{\adp}_{B})^{c}\cap\mathcal{G}^{\adp}_{B}\}.\label{eq:regret-nonleaf-good-event-temp-adp}
\end{align}

In what follows, we shall prove that for any $i\geq 0$, one has
\begin{align}
\max_{B\in\mathcal{B}_{i}}\mathbb{P}\{(\mathcal{E}^{\adp}_{B})^{c}\cap\mathcal{G}^{\adp}_{B}\}\lesssim
\frac{K}{n_{Q}}2^{(d+2\hatbeta)i}\log^{+}\big(n_{Q}2^{-(d+2\hatbeta)i} \big). \label{eq:prob-Ec-F-i-large-adp}
\end{align}
Moreover, for any $0\leq i<l_{4}^{\adp}\wedge l_{5}^\adp$, the following further holds
\begin{align}
\max_{B\in\mathcal{B}_{i}}\mathbb{P}\{(\mathcal{E}^{\adp}_{B})^{c}\cap\mathcal{G}^{\adp}_{B}\}\lesssim
K\exp\left(-\frac{c^{\adp}\kappa}{K}n_{P}2^{-(d+2\hatbeta+\gamma)i}\right) + K\bigg(\frac{1}{n}2^{(d+2\hatbeta)i}+ \frac{1}{n_{P}^{5}}\bigg); \label{eq:prob-Ec-F-i-small-adp}
\end{align}
Putting (\ref{eq:regret-nonleaf-good-event-temp-adp})--(\ref{eq:prob-Ec-F-i-small-adp}) together results in the advertised bounds (\ref{eq:regret-non-leaf-adp-born-i-large}) and (\ref{eq:regret-non-leaf-adp-born-i-small}).

Therefore, it remains to establish (\ref{eq:prob-Ec-F-i-large-adp}) and (\ref{eq:prob-Ec-F-i-small-adp}).
Towards this, by definition of $\mathcal{E}^{\adp}_{B}$, we can express
\begin{align*}
\mathbb{P}\{(\mathcal{E}^{\adp}_{B})^{c}\cap\mathcal{G}^{\adp}_{B}\} & =\mathbb{P}\big\{\{\underline{\mathcal{I}}^{\adp}_{B}\not\subset\mathcal{I}^{\adp}_{B}\}\cap\mathcal{G}^{\adp}_{B}\big\}+\mathbb{P}\big\{\{\underline{\mathcal{I}}^{\adp}_{B}\subset\mathcal{I}^{\adp}_{B}\}\cap\{\mathcal{I}^{\adp}_{B}\not\subset\overline{\mathcal{I}}^{\adp}_{B}\}\cap\mathcal{G}^{\adp}_{B}\big\},
\end{align*}
and let us bound these two terms individually.
\begin{itemize}
\item 
For the first term, we can apply a similar argument as in \eqref{eq:bad-event-1-oracle} to get
\begin{align}
\mathbb{P}&\big\{\{\underline{\mathcal{I}}_{B}^\adp\not\subset\mathcal{I}_{B}^\adp\}\cap\mathcal{G}^{\adp}_{B}\big\} \nonumber \\
&\leq\mathbb{P}\big\{\exists k\in\mathcal{I}_{\mathsf{p}(B)},0\leq\tau\leq\widehat{\tau}_{k}^{\star}(B):\big|\overline{Y}_{\tau}^{(k)}(B)-\mathbb{E}[\overline{Y}_{\tau}^{(k)}(B)]\big|\geq \widehat{U}_{k}(\tau,B)/2\big\} \nonumber \\
&\leq \sum_{k\in [K]}\mathbb{P}\big\{\exists0\leq\tau\leq\widehat{\tau}_{k}^{\star}(B):\big|\overline{Y}_{\tau}^{(k)}(B)-\mathbb{E}[\overline{Y}_{\tau}^{(k)}(B)]\big|\geq \widehat{U}_{k}(\tau,B)/2\big\} \label{eq:bad-event-1-adp}.
\end{align}
\item 
For the second term, one can also use a similar argument as in \eqref{eq:bad-event-2-oracle} to derive
\begin{align}
 & \mathbb{P}\big\{\{\underline{\mathcal{I}}_{B}^\adp\subset\mathcal{I}_{B}^\adp\}\cap\{\mathcal{I}_{B}^\adp\not\subset\overline{\mathcal{I}}_{B}^\adp\}\cap\mathcal{G}^{\adp}_{B}\big\} \nonumber \\
 & \quad\leq\mathbb{P}\big\{\exists k\in\mathcal{I}_{\mathsf{p}(B)}:\big|\overline{Y}_{\widehat{\tau}_{k}^{\star}(B)}^{(k)}(B)-\mathbb{E}[\overline{Y}_{\widehat{\tau}_{k}^{\star}(B)}^{(k)}(B)]\big|>\widehat{U}_{k}(\widehat{\tau}^{\star}_{k}(B),B)/2\big\} \nonumber \\
 & \quad\leq \sum_{k\in [K]}\mathbb{P}\big\{\big|\overline{Y}_{\widehat{\tau}_{k}^{\star}(B)}^{(k)}(B)-\mathbb{E}[\overline{Y}_{\widehat{\tau}_{k}^{\star}(B)}^{(k)}(B)]\big|>\widehat{U}_{k}(\widehat{\tau}^{\star}_{k}(B),B)/2\big\}.\label{eq:bad-event-2-adp}
\end{align}
\item Collecting \eqref{eq:bad-event-1-adp} and \eqref{eq:bad-event-2-adp} together reveals that
\begin{align}
\mathbb{P}\big\{(\mathcal{E}^{\adp}_{B})^{c}\cap\mathcal{G}^{\adp}_{B}\big\} & \leq \sum_{k\in[K]}\mathbb{P}\big\{\exists0\leq\tau\leq\widehat{\tau}_{k}^{\star}(B):\big|\overline{Y}_{\tau}^{(k)}(B)-\mathbb{E}[\overline{Y}_{\tau}^{(k)}(B)]\big|\geq \widehat{U}_{k}(\tau,B)/2\big\} \nonumber\\
 & \quad+\sum_{k\in[K]}\mathbb{P}\big\{\big|\overline{Y}_{\widehat{\tau}_{k}^{\star}(B)}^{(k)}(B)-\mathbb{E}[\overline{Y}_{\widehat{\tau}_{k}^{\star}(B)}^{(k)}(B)]\big|>\widehat{U}_{k}(\widehat{\tau}^{\star}_{k}(B),B)/2\big\} \nonumber \\
&\leq 2\sum_{k\in[K]}\mathbb{P}\big\{\exists0\leq\tau\leq\widehat{\tau}_{k}^{\star}(B):\big|\overline{Y}_{\tau}^{(k)}(B)-\mathbb{E}[\overline{Y}_{\tau}^{(k)}(B)]\big|\geq \widehat{U}_{k}(\tau,B)/2\big\}.
  \label{eq:prob-Ec-F-temp-adp}
\end{align}

Lemma \ref{lemma:prob-Ec-F-adp} below provides the upper bound for controlling the probability on the right-hand side of \eqref{eq:prob-Ec-F-temp-adp}. The proof can be found in Appendix~\ref{sec:Proof of lemma:prob-Ec-F-adp}.

\begin{lemma}
\label{lemma:prob-Ec-F-adp}
Instate the assumptions of Theorem~\ref{thm:upper-bound-adaptive}. For any fixed arm $k\in[K]$ and bin $B\in \mathcal{B}_i$ with $i \geq 0$, we have
\begin{align}
& \mathbb{P}\big\{\exists0\leq\tau\leq\widehat{\tau}_{k}^{\star}(B):\big|\overline{Y}_{\tau}^{(k)}(B)-\mathbb{E}[\overline{Y}_{\tau}^{(k)}(B)]\big|\geq \widehat{U}_{k}(\tau,B)/2\big\} \nonumber  \\
& \qquad \lesssim
\frac{1}{n_{Q}}2^{(d+2\hatbeta)i}\log^{+}\big(n_{Q}2^{-(d+2\hatbeta)i} \big),
\end{align}
In addition, if $0\leq i<l_{4}^{\adp}$, one can further bound
\begin{align}
\mathbb{P}\big\{&\exists0\leq\tau\leq\widehat{\tau}_{k}^{\star}(B):\big|\overline{Y}_{\tau}^{(k)}(B)-\mathbb{E}[\overline{Y}_{\tau}^{(k)}(B)]\big|\geq \widehat{U}_{k}(\tau,B)/2\big\} \nonumber \\
& \lesssim \exp\left(-\frac{c^{\adp}\kappa}{K}n_{P}2^{-(d+2\hatbeta+\gamma)i}\right) + \frac{1}{n}2^{(d+2\hatbeta)i}+ \frac{1}{n_{P}^{5}}.
\end{align}

% In addition, the inequalities above also hold for $\mathbb{P}\big\{\big|\overline{Y}_{\widehat{\tau}_{k}^{\star}(B)}^{(k)}(B)-\mathbb{E}[\overline{Y}_{\widehat{\tau}_{k}^{\star}(B)}^{(k)}(B)]\big|>\widehat{U}_{k}(\widehat{\tau}^{\star}_{k}(B),B)/2\big\}$.
\end{lemma}

Therefore, applying Lemma~\ref{lemma:prob-Ec-F-adp} to (\ref{eq:prob-Ec-F-temp-adp}) establishes (\ref{eq:prob-Ec-F-i-large-adp}) and (\ref{eq:prob-Ec-F-i-small-adp}).

\end{itemize}

\subsection{Proof of Lemma \ref{lemma:prob-Ec-F-adp}}
\label{sec:Proof of lemma:prob-Ec-F-adp}

Fix an arbitrary arm $k\in[K]$ and an arbitrary bin $B \in \mathcal{B}_{i}$ such that $i\geq0$ and $Q_X(B)>0$. 
% It suffices to $\mathbb{P}\big\{ \exists0\leq\tau\leq\widehat{\tau}_{k}^{\star}(B):\big|\overline{Y}_{\tau}^{(k)}(B)-\mathbb{E}[\overline{Y}_{\tau}^{(k)}(B)]\big|\geq \widehat{U}_{k}(\tau,B)/2\big\}$.  
Conditioned on $\cD^P_\dm$, one can invoke Lemma~\ref{lemma:exist-t-AH} to find
\begin{align}
& \mathbb{P} \big\{\exists0\leq\tau\leq\widehat{\tau}_{k}^{\star}(B):\big|\overline{Y}_{\tau}^{(k)}(B)-\mathbb{E}[\overline{Y}_{\tau}^{(k)}(B)]\big|\geq \widehat{U}_{k}(\tau,B)/2\,\big|\,\cD^P_\dm\big\} \lesssim\frac{\widehat{\tau}_{k}^{\star}(B)}{n_{Q}|B|^{d}} + \frac{1}{n|B|^{d+2\widehat{\beta}}}.\label{eq:regret-b-prob-ub-adp}
\end{align}
In order to control $\widehat{\tau}_{k}^{\star}(B)$, recall the definition of $\widehat{U}_{k}(\tau,B)$ in \eqref{eq:UCB-adapt}. Given that $n_{k}^{P}(B) \geq 0$, we can upper bound
\begin{align}
\widehat{\tau}_{k}^{\star}(B)\leq \olc_{2}|B|^{-2\hatbeta}\log^{+}\big(n_{Q}|B|^{d+2\hatbeta}\big),\label{eq:tau-B-k-non-concen-adp}
\end{align}
where $\olc_{2}\defn 2/ \overline{C}_{\beta}^2 \vee 1$. Moreover, since $\tau \geq 0$, when $\overline{C}_{\beta}^2n^{P}_{k}(B)|B|^{2\hatbeta}\geq2$ holds, we can also bound
\begin{align}
\widehat{\tau}^{\star}_{k}(B) \leq n_{Q}|B|^{d}\exp\left(-\frac{1}{2}\overline{C}_{\beta}^2n^{P}_{k}(B)|B|^{2\hatbeta}\right), \label{eq:tau-B-k-exp-adp}
\end{align}
Define the event 
\begin{align}
\cA_{B}^{\adp} \defn\Big\{n_{k}^{P}(B)\geq\frac{ \underline{q}c_{\gamma}\kappa n_{P}}{4K}|B|^{d+\gamma}\Big\}.\label{eq:event-aux-sample-size-lb-adp}
\end{align} 
On the event $\cA_{B}^{\adp}$, if $0 \leq i < l_{5}^{\adp}\defn\big\lceil \frac{1}{d+2\hatbeta+\gamma}\log(\frac{c^{\adp}\kappa n_{P}}{K})\big\rceil$, one has
\begin{align}
\widehat{\tau}_{k}^{\star}(B) & \leq n_{Q}|B|^d \exp\left(-\frac{c^{\adp}\kappa}{K}n_{P}|B|^{d+2\hatbeta+\gamma}\right), \label{eq:tau-B-k-concen-adp}
\end{align}
because $\frac{1}{2} \overline{C}_{\beta}^2n^{P}_{k}(B)|B|^{2\hatbeta} \geq \frac{\overline{C}_{\beta}^2 \underline{q}c_{\gamma}\kappa n_{P}}{8K}|B|^{d+2\hatbeta+\gamma} = \frac{c^{\adp}\kappa n_{P}}{K}2^{-(d+2\hatbeta+\gamma)i} \geq 1$.
Moreover, observe that $n_{k}^{P}(B)$ is a sum of i.i.d.~zero-mean Bernoulli random variables with sample size $|\cD^P_\dm|$. As a reminder, the value of $s_{P}$ in Procedure \ref{alg:smoothness} guarantees that $|\mathcal{D}^{P}_{\mathsf{dm}}| = n_{P}-s_{P}=(1-o(1))n_{P}$. Therefore, we can use Assumption~\ref{assumption:bounded-density} and Definitions~\ref{def:transfer}--\ref{def:explore-coef} to lower bound
\begin{align*}
\mathbb{E}[n_{k}^{P}(B)] & \geq P_{X}(B)\frac{ \kappa |\cD^P_\dm|}{K}\geq c_{\gamma}|B|^{\gamma}Q_{X}(B)\frac{ \kappa  n_{P}}{2K}\geq\frac{\underline{q} c_{\gamma} \kappa n_{P}}{2K}|B|^{d+\gamma} = \frac{\underline{q} c_{\gamma} \kappa n_P}{2K}2^{-(d+\gamma)i}.
\end{align*}
By the choice $l_{4}^{\adp} \defn \big\lceil\frac{1}{d+\gamma}\log_{2}(\frac{\underline{q}c_{\gamma} \kappa n_{P}}{80K\log(n_{P})})\big\rceil$, one has $\mathbb{E}[n_{k}^{P}(B)]\geq40\log(n_{P})$ for any $0 \leq i < l_{4}^{\adp}$. 
Then we can apply the Chernoff bound to obtain
\begin{align}
\mathbb{P}\{(\cA_{B}^{\adp})^{c}\} & =\mathbb{P}\Big\{ n_{k}^{P}(B)<\frac{\underline{q}c_{\gamma}\kappa n_{P}}{2K}|B|^{d+\gamma}\Big\} \leq \mathbb{P}\Big\{ n_{k}^{P}(B)< \frac{1}{4} \mathbb{E}[ n_{k}^{P}(B)]\Big\} \nonumber \\
 &  \leq \exp(-\mathbb{E}[n_{k}^{P}(B)]/8) \leq n^{-5}_{P} .\label{eq:prob-G-c-adp}
\end{align}

Now, we are ready to bound $\mathbb{P}\big\{ \exists0\leq\tau\leq\widehat{\tau}_{k}^{\star}(B):\big|\overline{Y}_{\tau}^{(k)}(B)-\mathbb{E}[\overline{Y}_{\tau}^{(k)}(B)]\big|\geq \widehat{U}_{k}(\tau,B)/2\big\}$. 
Combining (\ref{eq:regret-b-prob-ub-adp}) and (\ref{eq:tau-B-k-non-concen-adp}), we can upper bound
\begin{align}
 \mathbb{P}\big\{& \exists0\leq\tau\leq\widehat{\tau}_{k}^{\star}(B):\big|\overline{Y}_{\tau}^{(k)}(B)-\mathbb{E}[\overline{Y}_{\tau}^{(k)}(B)]\big| \geq \widehat{U}_{k}(\tau,B)/2\big\} \nonumber  \\
&  \lesssim \frac{\widehat{\tau}_{k}^{\star}(B)}{n_{Q}|B|^{d}} + \frac{1}{n} |B|^{-(d+2\hatbeta)} \lesssim \frac{1}{n_{Q}} |B|^{-(d+2\hatbeta)}\log^{+}\big(n_{Q}|B|^{d+2\hatbeta}\big) + \frac{1}{n} |B|^{-(d+2\hatbeta)} \nonumber \\ 
&   \asymp \frac{1}{n_{Q}}2^{(d+2\hatbeta)i}\log^{+}\big(n_{Q}2^{-(d+2\hatbeta)i} \big). \label{eq:eq:regret-b-prob-ub-large-i-adp}
\end{align}
In addition, if $0\leq i < l_{4}^{\adp}\wedge l_{5}^{\adp}$, we further have
\begin{align}
\mathbb{P}\big\{ & \exists0\leq\tau\leq\widehat{\tau}_{k}^{\star}(B):\big|\overline{Y}_{\tau}^{(k)}(B)-\mathbb{E}[\overline{Y}_{\tau}^{(k)}(B)]\big|\geq \widehat{U}_{k}(\tau,B)/2\big\} \nonumber \\
 & \leq  \mathbb{P}\big\{\exists0\leq\tau\leq\widehat{\tau}_{k}^{\star}(B):\big|\overline{Y}_{\tau}^{(k)}(B)-\mathbb{E}[\overline{Y}_{\tau}^{(k)}(B)]\big|\geq \widehat{U}_{k}(\tau,B)/2\,\big|\,\cA_{B}^{\adp}\big\} +\mathbb{P}\{(\cA_{B}^{\adp})^{c}\}\nonumber \\
 & \overset{(\mathrm{i})}{\lesssim} \frac{\widehat{\tau}_{k}^{\star}(B)}{n_{Q}|B|^{d}} + \frac{1}{n}|B|^{-(d+2\hatbeta)} + \frac{1}{n^{5}_{P}}\nonumber \overset{(\mathrm{ii})}{\lesssim} \exp\left(-\frac{c^{\adp}\kappa}{K}n_{P}|B|^{d+2\hatbeta+\gamma}\right) + \frac{1}{n}|B|^{-(d+2\hatbeta)}+ \frac{1}{n_{P}^{5}} \nonumber \\
 & = \exp\left(-\frac{c^{\adp}\kappa}{K}n_{P}2^{-(d+2\hatbeta+\gamma)i}\right) + \frac{1}{n}2^{(d+2\hatbeta)i} + \frac{1}{n_{P}^{5}}, \label{eq:eq:regret-b-prob-ub-small-i-adp}
\end{align}
where (i) arises from (\ref{eq:regret-b-prob-ub-adp}), (\ref{eq:prob-G-c-adp}) and the fact that the event $\cA_{B}^{\adp}$ belongs to the $\sigma$-algebra generated by $\cD^P_\dm$; (ii) uses (\ref{eq:tau-B-k-concen-adp}).
Collecting (\ref{eq:eq:regret-b-prob-ub-large-i-adp}) and (\ref{eq:eq:regret-b-prob-ub-small-i-adp}) together finishes the proof of Lemma~\ref{lemma:prob-Ec-F-adp}.

\subsection{Proof of Lemma \ref{lemma:non-leaf-upper-adp-live}}
\label{sec:Proof-of-lemma:non-leaf-upper-adp-live}

Similar to the proof of Lemma~\ref{lemma:non-leaf-upper-adp-born}, we use $\bP\{\cdot\}$ and $\bE[\cdot]$ to represent the probability and expectation conditioned on $\hatbeta$ for the sake of simplicity. Also, we shall apply a similar argument as in the proof of Lemma~\ref{lemma:non-leaf-upper-oracle-live} in Appendix~\ref{sec:Proof-of-lemma:non-leaf-upper-oracle-live} to establish Lemma \ref{lemma:non-leaf-upper-adp-live}.

We recall the definition of $R^{\lf,\adp}(B)$ in (\ref{eq:R-leaf-B-adp}). Fix an arbitrary bin $B \in \mathcal{B}_{i}$ such that $i \geq 0$ and $Q_{X}(B) > 0$. By (\ref{eq:reward-gap-event-ancestor-adp}), we can upper bound
\begin{align}
 & \mathbb{E}[R^{\lf,\adp}(B)\mathds{1}\{\mathcal{E}^{\adp}_{B}\cap\mathcal{G}^{\adp}_{B}\}] \nonumber \\
 & \quad=\mathbb{E}\bigg[\mathds{1}\{\mathcal{E}^{\adp}_{B}\cap\mathcal{G}^{\adp}_{B}\}\sum_{t=s_Q}^{n_{Q}}\big(f_{\pi^{\star}(X_{t}^{Q})}(X_{t}^{Q})-f_{\pi^{\adp}_{t}(X_{t}^{Q})}(X_{t}^{Q})\big)\mathds{1}\{X_{t}^{Q}\in B,B\in\mathcal{L}^{\adp}_{t}\}\Big] \nonumber \\
 & \quad\leq\mathbb{E}\bigg[\sum_{t=s_Q}^{n_{Q}} \olc_{1}|B|^{\hatbeta}\mathds{1}\{0<f_{(1)}(X_{t}^{Q})-f_{(2)}(X_{t}^{Q})\leq \olc_{1}|B|^{\hatbeta}\}\mathds{1}\{X_{t}^{Q}\in B,B\in\mathcal{L}^{\adp}_{t}\}\Big] \nonumber \\
 & \quad\lesssim |B|^{\hatbeta}Q_{X}(0<f_{(1)}(X)-f_{(2)}(X)\leq \olc_{1}|B|^{\hatbeta}\,|\,X\in B)\sum_{k\in[K]}\mathbb{E}[\widehat{\tau}_{k}^{\star}(B)]. \label{eq:non-leaf-upper-adp-live-temp}
\end{align}
Here, the last step arises from the round limits in Algorithm~\ref{alg:UCB-TL-adaptive}. 
In addition, we can combine Assumptions \ref{assumption:margin}--\ref{assumption:bounded-density} to bound
\begin{align}
\sum_{B\in\mathcal{B}_{i}:Q_{X}(B) > 0} & Q_{X}(0<f_{(1)}(X)-f_{(2)}(X)\leq \olc_{1}|B|^{\hatbeta}\,|\,X\in B) \nonumber \\
% & =\sum_{B\in\mathcal{B}_{i},\,Q_{X}(B)>0}\frac{1}{Q_{X}(B)}Q_{X} \big(0<f_{(1)}(X)-f_{(2)}(X)\leq \olc_{1}|B|^{\hatbeta},X\in B \big)\nonumber \\
 & \overset{(\mathrm{i})}{\leq}\frac{1}{\underline{q}|B|^{d}}Q_{X} \big(0<f_{(1)}(X)-f_{(2)}(X)\leq \olc_{1}|B|^{\hatbeta} \big)\nonumber \\
 & \overset{(\mathrm{ii})}{\leq}\frac{1}{\underline{q}|B|^{d}} C_{\alpha}(c_{1}|B|^{\hatbeta})^{\alpha} \lesssim|B|^{\alpha\hatbeta-d} = 2^{(d-\alpha \hatbeta)i}. \label{eq:cond-prob-Q-gap-ub-adp}
\end{align} 
As a result, summing over all bins in $\mathcal{B}_{i}$, we know from (\ref{eq:non-leaf-upper-adp-live-temp}) and (\ref{eq:cond-prob-Q-gap-ub-adp}) that
\begin{align}
\mathbb{E}[R^{\lf,\adp}_{i}] & =\sum_{B\in\mathcal{B}_{i}}\mathbb{E}\big[R^{\lf,\adp}(B)\mathds{1}\{\mathcal{E}^{\adp}_{B}\cap\mathcal{G}^{\adp}_{B}\}\big]
 % & \lesssim K|B|^{\hatbeta}\max_{B\in\mathcal{B}_{i},k\in[K]}\mathbb{E}[\widehat{\tau}_{k}^{\star}(B)] \nonumber \\
 % & \qquad \sum_{B\in\mathcal{B}_{i}, Q_{X}(B) > 0}Q_{X}(0<f_{(1)}(X)-f_{(2)}(X)\leq \olc_{1}|B|^{\hatbeta}\,|\,X\in B)\nonumber \\
 \lesssim K2^{(d- (1+\alpha)\hatbeta)i}\max_{B\in\mathcal{B}_{i},k\in[K]}\mathbb{E}[\widehat{\tau}_{k}^{\star}(B)]. \label{eq:regret-leaf-bad-event-adp}
\end{align}

Consequently, it boils down to bounding $\max_{B\in\mathcal{B}_{i},k\in[K]}\mathbb{E}[\widehat{\tau}_{k}^{\star}(B)]$. To this end, by (\ref{eq:tau-B-k-non-concen-adp}), for any $i \geq 0$, one can bound directly
\begin{align}
\mathbb{E}[\widehat{\tau}_{k}^{\star}(B)]\lesssim |B|^{-2\hatbeta}\log^{+}\big(n_{Q}|B|^{d+2\hatbeta}\big) = 2^{2\hatbeta i}\log^{+}\big(n_{Q}2^{-(d+2\hatbeta)i}\big). \label{eq:tau-ub-exp-i-large-adp}
\end{align}
In addition, when $0\leq i<l_{4}^{\adp}$, let us recall the event $\mathcal{A}_{B}^{\adp}$ in \eqref{eq:event-aux-sample-size-lb-adp}. Collecting (\ref{eq:tau-B-k-non-concen-adp}), (\ref{eq:tau-B-k-concen-adp}), and (\ref{eq:prob-G-c-adp}) together reveals that
\begin{align}
\mathbb{E}[\tau_{k}^{\star}(B)] & =\mathbb{E}[\tau_{k}^{\star}(B)\,|\,\cA_{B}^\adp]\mathbb{P}\{\cA_{B}^\adp\}+\mathbb{E}[\tau_{k}^{\star}(B)\,|\,(\cA_{B}^\adp)^{c}]\mathbb{P}\{(\cA_{B}^\adp)^{c}\} \nonumber \\
& \leq\mathbb{E}[\tau_{k}^{\star}(B)\,|\,\cA_{B}^\adp]+\mathbb{E}[\tau_{k}^{\star}(B)\,|\,\cA_{B}^{c}]\mathbb{P}\{\cA_{B}^{c}\}\nonumber\\
& \lesssim n_{Q}|B|^d \exp\left(-\frac{c^{\adp}\kappa}{K}n_{P}|B|^{d+2\hatbeta+\gamma}\right)
 +\frac{1}{n^{5}_P}|B|^{-2\hatbeta}\log^{+}\big(n_{Q}|B|^{d+2\hatbeta}\big) \nonumber\\
& = n_{Q}2^{-di}\exp\left(-\frac{c^{\adp}\kappa}{K}n_{P}2^{-(d+2\hatbeta+\gamma)i}\right)+ \frac{1}{n^{5}_P}2^{2\hatbeta i}\log^{+}\big(n_{Q}2^{-(d+2\hatbeta)i}\big) \label{eq:tau-ub-exp-i-small-adp}.
\end{align}

Therefore, substituting the upper bounds (\ref{eq:tau-ub-exp-i-small-adp}) and (\ref{eq:tau-ub-exp-i-large-adp}) back into (\ref{eq:regret-leaf-bad-event-adp}) finishes the proof of Lemma~\ref{lemma:non-leaf-upper-adp-live}.

\subsection{Proof of Lemma \ref{lemma:leaf-upper-adp}}

\label{subsec:Proof-of-lemma:leaf-upper-adp}

Once again, let $\bP\{\cdot\}$ and $\bE[\cdot]$ represent the probability and expectation conditioned on $\hatbeta$ to simplify our presentation.

For any bin $B\in\mathcal{B}_{l^{\adp}}$, we can derive
\begin{align*}
\mathbb{E}\big[R^{\tree,\adp}(B)\mathds{1}\{\mathcal{G}^{\adp}_{B}\}\big] 
& =\mathbb{E}\bigg[\mathds{1}\{\mathcal{G}^{\adp}_{B}\}\sum_{t=s_Q}^{n_{Q}}\big(f_{\pi^{\star}(X_{t}^{Q})}(X_{t}^{Q})-f_{\pi^{\adp}_{t}(X_{t}^{Q})}(X_{t}^{Q})\big)\mathds{1}\{X_{t}^{Q}\in B,\,B\in\mathcal{T}^{\adp}_{t}\}\Big]\\
 & \leq\mathbb{E}\bigg[\sum_{t=s_Q}^{n_{Q}} \olc_{1}|B|^{\hatbeta}\mathds{1}\{0<f_{(1)}(X_{t}^{Q})-f_{(2)}(X_{t}^{Q})\leq \olc_{1}|B|^{\hatbeta}\}\mathds{1}\{X_{t}^{Q}\in B\}\Big]\\
 & \lesssim \sum_{t=s_Q}^{n_{Q}} |B|^{\hatbeta}Q_{X}\big(0<f_{(1)}(X)-f_{(2)}(X)\leq \olc_{1}|B|^{\hatbeta},X\in B\big) \\
& = \sum_{t=s_Q}^{n_{Q}} 2^{-\hatbeta l^{\adp}}Q_{X}\big(0<f_{(1)}(X)-f_{(2)}(X)\leq \olc_{1}2^{-\hatbeta l^{\adp}},X\in B\big)
\end{align*}
where the first inequality is due to (\ref{eq:reward-gap-event-ancestor-adp}).
As a result, summing over all bins $B\in\mathcal{B}_{l^{\adp}}$ yields
\begin{align*}
\mathbb{E}[R^{\adp}_{l^\adp}] & =\sum_{B\in\mathcal{B}_{ l^{\adp}}}\mathbb{E}\big[R^{\tree,\adp}(B)\mathds{1}\{\mathcal{G}^{\adp}_{B}\}\big] 
\lesssim \sum_{t=s_{Q}}^{n_{Q}}2^{-\hatbeta l^{\adp}}Q_{X}\big(0<f_{(1)}(X)-f_{(2)}(X)\leq \olc_{1}2^{-\hatbeta l^{\adp}}\big) \\
 & \leq n_{Q} 2^{-\hatbeta l^{\adp}} C_{\alpha} \big( \olc_{1}2^{-\hatbeta l^{\adp}} \big)^{\alpha} \lesssim n_{Q}2^{-  (1+\alpha)\hatbeta l^{\adp}}.
\end{align*}
Here, the last step follows from the margin condition (Assumption~\ref{assumption:margin}). This finishes the proof of Lemma \ref{lemma:leaf-upper-adp}.

\subsection{Proof of Lemma~\ref{lemma:smooth-est-error}}
\label{proof-lemma:smooth-est-error}

Before devolving into the details, we pause to remind readers of the key quantities defined in Procedure~\ref{alg:smoothness}. To begin with, the smoothness estimator is given by $\widehat{\beta}\coloneqq-\frac{1}{l_{1}}\log_{2}(\mathsf{b})-C_{2}\frac{\log_{2}(\log(n))}{\log_{2}(n)}$ for some constant $C_{2}>0$, where we recall $\mathsf{b}\coloneqq \max_{k\in[K],x\in\mathcal{M}}\big|\widehat{f}_{k}(x;2^{-l_{1}})-\widehat{f}_{k}(x;2^{-l_{2}})\big|$ and $n\coloneqq n_{Q}\vee n_{Q}$. Here, the reward function estimator is given by $\widehat{f}_{k}(x;2^{-l_{i}})\coloneqq\widehat{\eta}_{k}(x;B_{i}(x))$ for $i \in\{1, 2\}$, where the local regression estimator $\widehat{\eta}_{k}(x;B)$ is defined in (\ref{def:local-estimator}) for any bin $B$, and $B_{i}(x)\in\mathcal{B}_{l_{i}}$ is the bin that contains $x$. With these definitions in place, we shall control $\mathsf{b}$ in what follows.

\paragraph*{Case I: $n_{P}>n_{Q}$} Let us start with the case $n_{P}>n_{Q}$, where one has $l_{1}=\big\lceil \frac{\underline{\beta}}{(d+2\overline{\beta}+\overline{\gamma})^2}\log_{2}(n_{P})\big\rceil$ and $l_{2}=l_{1}+\big\lceil\frac{1}{d}\log_{2}(\log(n_{P}))\big\rceil$.
Given $\mathsf{b}$ is largely determined by the difference between the two reward function estimators with different bandwidths, we find it helpful to present the following two lemmas that characterize the properties of the local regression estimator $\widehat{\eta}_{k}(x;B)$. 

First, Lemma~\ref{lemma:local-poly-reg-est-error} below controls the estimation error of the local regression estimator.

\begin{lemma}\label{lemma:local-poly-reg-est-error}
Fix an arbitrary bin $B\subset\mathcal{X}$ and an arbitrary arm $k\in[K]$. Let $\{(X_{i},Y_{i})\}_{i=1}^{m}$ in $\mathcal{X}\times[0,1]$ be a sequence of i.i.d.~samples such that $X_{i}$ represents the covariate and $Y_{i}$ corresponds to the random reward generated by arm $k$. Denote by $\lambda$ the distribution of the covariate. Let $\widehat{\eta}_{k}(x;B)$ be the local regression estimator based on $\{(X_{i},Y_{i})\}_{i=1}^{m}$ as in (\ref{def:local-estimator}). Suppose that $\log(Kn)=o\big(m\lambda(B)\big)$. Then for any $x\in B$, with probability at least $1-O(K^{-1}n^{-11})$,
\begin{align*}
\big|\widehat{\eta}_{k}(x;B)-f_{k}(x)\big|\leq \big(1+o(1)\big)C_{\beta}|B|^{\beta}+O\bigg(\sqrt{\frac{1}{m\lambda(B)}\log(Kn)}\bigg).
\end{align*}
\end{lemma}

\begin{proof}
	See Appendix~\ref{proof-lemma:local-poly-reg-est-error}.
\end{proof}

Next, Lemma~\ref{lemma:local-poly-reg-poly-proj-dist} below illustrates that the local regression estimator is sufficiently close to the $L_{2}$-projection of the reward function onto the space of piecewise constant functions.

\begin{lemma}\label{lemma:local-poly-reg-poly-proj-dist}
Instate the assumptions and notations in Lemma~\ref{lemma:local-poly-reg-est-error}. Recall that $\Gamma_{B}f_{k}(\cdot ; \lambda)$ denotes the $L_{2}(\lambda)$-projection of the reward function $f_{k}$ onto the class of piecewise constant functions over the bin $B$, defined in (\ref{eq:def-poly-proj}). Then for any $x\in B$, the following holds with probability at least $1-O(K^{-1}n^{-11})$:
\begin{align*}
\big|\widehat{\eta}_{k}(x;B)-\Gamma_{B}f_{k}(x ; \lambda)\big|\lesssim\sqrt{\frac{1}{m\lambda(B)}\log(Kn)}.
\end{align*}
\end{lemma}

\begin{proof}
	See Appendix~\ref{proof-lemma:local-poly-reg-poly-proj-dist}.
\end{proof}

% In addition, recall that the partial dataset $\mathcal{D}_{\mathsf{se}}$ used for smoothness estimation has a sample size $|\mathcal{D}_{\mathsf{se}}|=T$. Lemma~\ref{lemma:sample-per-bin-lb} below guarantees that we can collect enough samples in each bin in order to estimate the smoothness parameter reliably.

% \begin{lemma}\label{lemma:sample-per-bin-lb}
% Instate the assumption of Theorem~\ref{thm:upper-bound-adaptive}. With probability at least $1-O(n^{-10})$,
% \begin{align*}
% n_{k}^{P}(B;\mathcal{D}_{\mathsf{se}})\gtrsim T2^{-(d+\gamma)l_{1}}
% \end{align*}
% holds simultaneously for any bin $B\in\mathcal{B}_{l_{1}}$ and arm $k\in[K]$, and
% \begin{align*}
% n_{k}^{P}(B;\mathcal{D}_{\mathsf{se}})\gtrsim T2^{-(d+\gamma)l_{2}}
% \end{align*}
% holds simultaneously for any bin $B\in\mathcal{B}_{l_{2}}$ and arm $k\in[K]$.
% \end{lemma}

% \begin{proof}
% 	See Appendix \ref{proof-lemma:sample-per-bin-lb}.
% \end{proof}

With Lemmas~\ref{lemma:local-poly-reg-est-error} and \ref{lemma:local-poly-reg-poly-proj-dist} in hand, now we are ready to control $\max_{k\in[K],\,x\in\mathcal{M}}\big|\widehat{f}_{k}(x;2^{-l_{1}})-\widehat{f}_{k}(x;2^{-l_{2}})\big|$. 
\begin{itemize}
	\item We begin with the upper bound.
Fix an arbitrary $k\in[K]$ and $x\in\mathcal{M}$. By the triangle inequality, we can upper bound
\begin{align*}
\big|\widehat{f}_{k}(x;2^{-l_{1}})-\widehat{f}_{k}(x;2^{-l_{2}})\big|\leq\big|\widehat{f}_{k}(x;2^{-l_{1}})-f_{k}(x)\big|+\big|\widehat{f}_{k}(x;2^{-l_{2}})-f_{k}(x)\big|.
\end{align*}
Note that the covariates $X_{i}$'s in the dataset $\mathcal{D}_{\mathsf{se}}^{(k)}$ are i.i.d.~according to $P_{X\,|\,\pi=k}$. One can lower bound the density of $P_{X\,|\,\pi=k}$ by
\begin{align*}
	p_{X\,|\,\pi=k}(x) &= \frac{p_{X}(x)\mu(k\,|\,x)}{\int_{\mathcal{X}}p_{X}(x)\mu(k\,|\,x)\,\mathrm{d}x} \\
	& \geq \frac{p_{X}(x) \kappa / K}{\int_{\mathcal{X}}p_{X}(x)(1-\kappa+\kappa/K)\,\mathrm{d}x} \\
	& = \frac{1}{K \frac{1-\kappa}{\kappa}+1} p_{X}(x)
\end{align*}
for any $x\in\mathsf{supp}(P_{X})$, where the inequality follows from an immediate consequence of Definition~\ref{def:explore-coef} that $\mu(k\,|\,x) \geq \kappa/K$ and $\mu(k\,|\,x)\leq 1-\kappa(K-1)/K$. Combined with Assumption~\ref{assumption:bounded-density} and Definition~\ref{def:transfer}, this implies for any bin $B$,
\begin{align*}
	P_{X\,|\,\pi=k}(B) & \geq \frac{1}{K \frac{1-\kappa}{\kappa}+1} P_{X}(B) \geq \frac{1}{K \frac{1-\kappa}{\kappa}+1} c_{\gamma} |B|^{\gamma} Q_{X}(B) \\
	& \geq \frac{c_{\gamma} \underline{q}}{K \frac{1-\kappa}{\kappa}+1} |B|^{d+\gamma} \gtrsim \frac{1}{K} |B|^{d+\gamma},
\end{align*}
provided $\kappa \asymp 1$.
As a result, applying Lemma~\ref{lemma:local-poly-reg-est-error} shows that for any $k\in[K]$, $B\in\mathcal{B}_{l_{1}}$ and $x\in B$, with probability at least $1-O(K^{-1}n^{-11})$,
\begin{align}
\big|\widehat{f}_{k}(x;2^{-l_{1}})-f_{k}(x)\big| & \leq\big(1+o(1)\big)C_{\beta}2^{-\beta l_{1}}+O\bigg(\sqrt{T^{-1} K 2^{(d+\gamma)l_{1}}\log(Kn)}\bigg) \nonumber \\
 & = \big(1+o(1)\big)C_{\beta}2^{-\beta l_{1}} \label{eq:f-hat-1-est-error}
\end{align}
provided $K2^{(2\beta+d+\gamma)l_{1}}\log(Kn) = o(T)$. Similarly, for any $k\in[K]$, $B\in\mathcal{B}_{l_{2}}$ and $x\in B$, we also have with probability exceeding $1-O(K^{-1}n^{-11})$,
\begin{align}
\big|\widehat{f}_{k}(x;2^{-l_{2}})-f_{k}(x)\big| & \leq\big(1+o(1)\big)C_{\beta}2^{-\beta l_{2}}+O\bigg(\sqrt{T^{-1}K 2^{(d+\gamma)l_{2}}\log(Kn)}\bigg)\nonumber \\
 & =o(2^{-\beta l_{1}})\label{eq:f-hat-2-est-error}
\end{align}
where the last line holds as long as $2^{-l_{2}} = o(2^{-l_{1}})$ and $K2^{2\beta l_{1}+(d+\gamma)l_{2}}\log(Kn) = o(T)$.
By the choice of $l_{3} = \lceil \frac{\overline{\beta}}{\underline{\beta}}l_{1}+\frac{1}{\underline{\beta}}\log_{2}(\log(n))\rceil$, one can upper bound $|\mathcal{M}|=2^{dl_{3}}\lesssim n$. Applying the union bound to (\ref{eq:f-hat-1-est-error}) and (\ref{eq:f-hat-2-est-error}) shows that with probability more than $1-O(n^{-10})$,
\begin{align}
\max_{k\in[K],x\in\mathcal{M}}\big|\widehat{f}_{k}(x;2^{-l_{1}})-\widehat{f}_{k}(x;2^{-l_{2}})\big|\leq2C_{\beta}2^{-\beta l_{1}}.\label{eq:f-hat-diff-ub}
\end{align}

\item
Next, we move on to lower bound $\max_{k\in[K],\,x\in\mathcal{M}}\big|\widehat{f}_{k}(x;2^{-l_{1}})-\widehat{f}_{k}(x;2^{-l_{2}})\big|$. By Assumption~\ref{assumption:self-similar}, we know that there exists a bin $B\in\mathcal{B}_{l_{1}}$, a point $x\in B$, and an arm $k\in[K]$ such that
\begin{align}
\big|f_{k}(x)-\Gamma_{B}f_{k}(x;P_{X\,|\,\pi=k})\big|\geq b2^{-\beta l_{1}},\label{eq:f-hat-poly-proj-x-lb}
\end{align}
for some constant $b>0$. Let $\widetilde{x}=\argmin_{x\in\mathcal{M}}\|x-\widetilde{x}\|_{\infty}$ (if there exist multiple minimizers, we choose the one that is closest to the origin). Then it suffices to lower bound $\big|\widehat{f}_{k}(\widetilde{x};2^{-l_{1}})-\widehat{f}_{k}(\widetilde{x};2^{-l_{2}})\big|$.
From (\ref{eq:f-hat-2-est-error}) and the triangle inequality, one knows that with probability at least $1-O(n^{-10})$,
\begin{align}
\big|\widehat{f}_{k}(\widetilde{x};2^{-l_{1}})-\widehat{f}_{k}(\widetilde{x};2^{-l_{2}})\big| & =\big|\widehat{f}_{k}(\widetilde{x};2^{-l_{1}})-f_{k}(\widetilde{x})+f_{k}(\widetilde{x})-\widehat{f}_{k}(\widetilde{x};2^{-l_{2}})\big|\nonumber \\
 & \geq\big|\widehat{f}_{k}(\widetilde{x};2^{-l_{1}})-f_{k}(\widetilde{x})\big|-\big|f_{k}(\widetilde{x})-\widehat{f}_{k}(\widetilde{x};2^{-l_{2}})\big|\nonumber \\
 & \geq\big|\widehat{f}_{k}(\widetilde{x};2^{-l_{1}})-f_{k}(\widetilde{x})\big|-o(2^{-\beta}l_{1}).\label{eq:eq:f-hat-diff-lb-temp}
\end{align}
This implies that it suffices to lower bound $\big|\widehat{f}_{k}(\widetilde{x};2^{-l_{1}})-f_{k}(\widetilde{x})\big|$, which can be further lower bounded by
\begin{align}
\big|\widehat{f}_{k}(\widetilde{x};2^{-l_{1}})-f_{k}(\widetilde{x})\big| & =\big|\widehat{f}_{k}(\widetilde{x};2^{-l_{1}})-\Gamma_{B}f_{k}(\widetilde{x};P_{X\,|\,\pi=k})+\Gamma_{B}f_{k}(\widetilde{x};P_{X\,|\,\pi=k})-f_{k}(\widetilde{x})\big|\nonumber \\
 & \geq\big|\Gamma_{B}f_{k}(\widetilde{x};P_{X\,|\,\pi=k})-f_{k}(\widetilde{x})\big|-\big|\widehat{f}_{k}(\widetilde{x};2^{-l_{1}})-\Gamma_{B}f_{k}(\widetilde{x};P_{X\,|\,\pi=k})\big|\label{eq:f-hat-x-tilde-est-error}
\end{align}

For the second term, applying Lemma~\ref{lemma:local-poly-reg-poly-proj-dist} demonstrates that with probability more than $1-O(n^{-10})$,
\begin{align}
\big|\widehat{f}_{k}(\widetilde{x};2^{-l_{1}})-\Gamma_{B}f_{k}(\widetilde{x};P_{X\,|\,\pi=k})\big|\lesssim\sqrt{T^{-1}K2^{(d+\gamma)l_{1}}\log(Kn)} = o(2^{-\beta l_{1}}),\label{eq:f-hat-x-tilde-est-error-term2}
\end{align}
where the last step holds as long as $K2^{(2\beta+d+\gamma)l_{1}}\log(Kn) = o(T)$.

Turning to the first term, we can use the triangle inequality again to obtain
\begin{align*}
\big|\Gamma_{B}f_{k}(\widetilde{x};P_{X\,|\,\pi=k})-f_{k}(\widetilde{x})\big| & =\big|\Gamma_{B}f_{k}(\widetilde{x};P_{X\,|\,\pi=k})-\Gamma_{B}f_{k}(x;P_{X\,|\,\pi=k}) \\
& \qquad +\Gamma_{B}f_{k}(x;P_{X\,|\,\pi=k})-f_{k}(x)+f_{k}(x)-f_{k}(\widetilde{x})\big|\\
 & \geq\big|\Gamma_{B}f_{k}(x;P_{X\,|\,\pi=k})-f_{k}(x)\big|-\big|f_{k}(x)-f_{k}(\widetilde{x})\big| \nonumber \\
 & \qquad - \big|\Gamma_{B}f_{k}(\widetilde{x};P_{X\,|\,\pi=k})-\Gamma_{B}f_{k}(x;P_{X\,|\,\pi=k})\big|.
\end{align*}
By the smoothness condition (Assumption~\ref{assumption:smooth}), construction of the grid $\mathcal{M}$, and $\underline{\beta}\leq\beta\leq\overline{\beta}$, we know that
\begin{align*}
|f_{k}(\widetilde{x})-f_{k}(x)| & \leq C_{\beta}\|\widetilde{x}-x\|_{\infty}^{\beta} \leq C_{\beta}2^{-\beta l_{3}} \\
& \leq C_{\beta} 2^{-\beta \big(\frac{\overline{\beta}}{\underline{\beta}}l_{1}+\frac{1}{\underline{\beta}}\log_{2}(\log(n)) \big) } \\
& \leq C_{\beta} 2^{-\big(\overline{\beta}l_{1}+\log_{2}(\log(n)) \big) } \\
& \leq  \frac{C_{\beta}}{\log(n)}2^{-\beta l_{1}} = o(2^{-\beta l_{1}}).
\end{align*}
Also, as both $\widetilde{x}$ and $x$ belong to $B$, one has $\Gamma_{B}f_{k}(x;P_{X\,|\,\pi=k})=\Gamma_{B}f_{k}(\widetilde{x};P_{X\,|\,\pi=k})$.
Collecting these two observations with (\ref{eq:f-hat-poly-proj-x-lb}) yields the lower bound for the first term:
\begin{align}
\big|\Gamma_{B}f_{k}(\widetilde{x};P_{X\,|\,\pi=k})-f_{k}(\widetilde{x})\big|\geq b2^{-\beta l_{1}}-o(2^{-\beta l_{1}}).\label{eq:f-hat-x-tilde-est-error-term1}
\end{align}

Combining (\ref{eq:f-hat-x-tilde-est-error-term2}) and (\ref{eq:f-hat-x-tilde-est-error-term1}) shows that with probability more than $1-O(n^{-10})$,
\begin{align*}
\big|\widehat{f}_{k}(\widetilde{x};2^{-l_{1}})-f_{k}(\widetilde{x})\big|\geq b2^{-\beta l_{1}}-o(2^{-\beta l_{1}}).
\end{align*}
Plugging this inequality back into (\ref{eq:eq:f-hat-diff-lb-temp}) gives that with probability at least $1-O(n^{-10})$,
\begin{align}
\big|\widehat{f}_{k}(\widetilde{x};2^{-l_{1}})-\widehat{f}_{k}(\widetilde{x};2^{-l_{2}})\big|\geq b2^{-\beta l_{1}}-o(2^{-\beta l_{1}})\geq\frac{b}{2}2^{-\beta l_{1}}.\label{eq:eq:f-hat-diff-lb}
\end{align}

\item 
Finally, we combine (\ref{eq:f-hat-diff-ub}) and (\ref{eq:eq:f-hat-diff-lb}) to conclude that with probability exceeding $1-O(n^{-10})$,
\begin{align*}
\frac{b}{2}2^{-\beta l_{1}}\leq\mathsf{b}\leq2C_{\beta}2^{-\beta l_{1}}.
\end{align*}
Recall that the smoothness estimator $\widehat{\beta}\defn-\frac{1}{l_{1}}\log_{2}(\mathsf{b})-C_{2}\frac{\log_{2}(\log(n))}{\log_{2}(n)}$.
The display above implies that for sufficiently large $n$,
\begin{align}
\widehat{\beta} & \leq\beta-\frac{\log_{2}(b/2)}{l_{1}}-\frac{C_{2}\log_{2}(\log(n))}{\log_{2}(n)} \nonumber \\
& \leq \beta-\frac{(d+2\overline{\beta}+\overline{\gamma})^2 \log_{2}(b/2)}{2\underline{\beta} \log_{2}(n)} -\frac{C_{2}\log_{2}(\log(n))}{\log_{2}(n)}\leq\beta, \label{eq:betahat-ub-P-large}
\end{align}
and
\begin{align}
\widehat{\beta}\geq\beta-\frac{\log_{2}(2C_{\beta})}{l_{1}}-\frac{C_{2}\log_{2}(\log(n))}{\log_{2}(n)}\geq\beta-\frac{C_{3}(d+2\overline{\beta}+\overline{\gamma})^2 \log_{2}(\log(n))}{\underline{\beta}\log_{2}(n)}. \label{eq:betahat-lb-P-large}
\end{align}
for some constant $C_{3} > 0$ independent of $n_{Q}$ and $n_{Q}$.
\end{itemize}

\paragraph*{Case II: $n_{P}\leq n_{Q}$} As for the case $n_{P}\leq n_{Q}$, one can apply a similar argument as above to show that with probability at least $1-O(n^{-10})$,
\begin{align}
\beta-\frac{C_{3}(d+2\overline{\beta})^2 \log_{2}(\log(n))}{\underline{\beta}\log_{2}(n)}\leq\widehat{\beta}\leq\beta, \label{eq:betahat-P-small}
\end{align}
for some constant $C_{3} > 0$ independent of $n_{Q}$ and $n_{Q}$. For the sake of conciseness, we omit the detailed proof.

\paragraph*{Combining Case I and Case II} Taking together \eqref{eq:betahat-ub-P-large}, \eqref{eq:betahat-lb-P-large}, and \eqref{eq:betahat-P-small}, we finish the proof of Lemma~\ref{lemma:smooth-est-error}.

\subsection{Proof of Lemma~\ref{lemma:local-poly-reg-est-error}}
\label{proof-lemma:local-poly-reg-est-error}
Fix an arm $k\in[K]$, a bin $B\subset\mathcal{X}$, and a point $x\in B$.
For notational convenience, let us denote $h\coloneqq |B|$. Recall that the local regression estimator is given by
\begin{align*}
\widehat{\eta}_{k}(x;B)=\frac{\sum_{i=1}^{m}Y_{i}\mathds{1}\{X_{i}\in B\}}{\sum_{i=1}^{m}\mathds{1}\{X_{i}\in B\}}
\end{align*}
if $\sum_{i=1}^{m}\mathds{1}\{X_{i}\in B\}>0$, and $\widehat{\eta}_{k}(x;B) = 0$ otherwise.
One can write
\begin{align*}
\widehat{\eta}_{k}(x;B)-f_{k}(x)=\frac{\sum_{i=1}^{m}\big(Y_{i}-f_{k}(x)\big)\mathds{1}\{X_{i}\in B\}}{\sum_{i=1}^{m}\mathds{1}\{X_{i}\in B\}}.
\end{align*}

In what follows, we shall control the denominator and numerator separately. 
\begin{itemize}
\item Let us start with the denominator. Define a sequence of i.i.d.~random variables
\begin{align*}
M_{i}\coloneqq \mathds{1}\{X_{i}\in B\},\quad i\in[m].
\end{align*}
For any $i\in[m]$, the expectation can be computed as $\mathbb{E}[M_{i}]=\lambda(B)$.
Also, it is straightforward to bound
\begin{align*}
|M_{i}-\mathbb{E}[M_{i}]| &\leq 1+\lambda(B)\leq2, \\
\sum_{i=1}^{m} \mathsf{Var}(M_{i}) & \leq \sum_{i=1}^{m} \mathbb{E}[M_{i}^{2}] = \sum_{i=1}^{m} \mathbb{E}[\mathds{1}\{X_{i}\in B\}]=m\lambda(B),
\end{align*}
for any $i\in[m]$, where we use the fact that $\lambda(B)\leq1$.
We can then invoke the Bernstein inequality to find that with probability at least $1-O(K^{-1}n^{-10})$,
\begin{align}
\Big|\sum_{i=1}^{m}(M_{i}-\mathbb{E}[M_{i}])\Big|&\lesssim \sqrt{m\lambda(B)\log(Kn)}+\log(Kn) \nonumber \\
& \asymp \sqrt{m\lambda(B)\log(Kn)}, \label{eq:count-est-error}
\end{align}
where the last step holds provided $m\lambda(B)\gtrsim\log(Kn)$.
Combining this with the triangle inequality, we conclude that with probability at least $1-O(K^{-1}n^{-10})$,
\begin{align}
\sum_{i=1}^{m}\mathds{1}\{X_{i}\in B\} & = \sum_{i=1}^{m}M_{i} \geq \sum_{i=1}^{m}\bE[M_{i}] - \Big|\sum_{i=1}^{m}(M_{i}-\mathbb{E}[M_{i}])\Big| \nonumber  \\
& \geq m\lambda(B)-O\Big(\sqrt{m\lambda(B)\log(Kn)}\Big)=\big(1-o(1)\big)m\lambda(B).\label{eq:lp-error-denom}
\end{align}
where the last step holds as long as $\log(Kn)/(m\lambda(B))=o(1)$.
\item Next, we turn to the numerator term. Define two sequences of i.i.d.~random variables by
\begin{align*}
N_{i} & \coloneqq \big(Y_{i}-f_{k}(X_{i})\big)\mathds{1}\{X_{i}\in B\},\quad i\in[m],\\
L_{i} & \coloneqq \big(f_{k}(X_{i})-f_{k}(x)\big)\mathds{1}\{X_{i}\in B\},\quad i\in[m].
\end{align*}

\begin{itemize}
\item It is not hard to see that for any $i\in[m]$,
\begin{align*}
	\mathbb{E}[N_{i}]= \mathbb{E}\big[ \mathds{1}\{X_{i}\in B\}\bE[Y_{i}-f_{k}(X_{i}) \,|\,X_i]\big] = 0.
\end{align*}
Also, as $Y_{i}$'s and $f_{k}$ are bounded in $[0,1]$ for any $i\in[m]$, we have $\max_{i\in[m]}|N_{i}|\leq 1$, and
\begin{align*}
\sum_{i=1}^{m}\mathsf{Var}(N_{i}) & \leq \sum_{i=1}^{m} \mathbb{E} [N_{i}^{2}] = \sum_{i=1}^{m} \mathbb{E}\big[\big(Y_{i}-f_{k}(X_{i})\big)^{2}\mathds{1}\{X_{i}\in B\}\big] \\
& \leq \sum_{i=1}^{m} \mathbb{E}\big[\mathds{1}\{X_{i}\in B\}\big] = m \lambda(B).
\end{align*}
Consequently, we can apply the Bernstein inequality to obtain that with probability at least $1-O(K^{-1}n^{-10})$,
\begin{align}
\Big|\sum_{i=1}^{m}N_{i}\Big|\lesssim \sqrt{m\lambda(B)\log(Kn)}+\log(Kn)\asymp \sqrt{m\lambda(B)\log(Kn)},\label{eq:poly-est-error-ub}
\end{align}
as long as $m\lambda(B)\gtrsim\log(Kn)$.
\item For any $i\in[m]$, we can use the smooth condition (Assumption~\ref{assumption:smooth}) to bound 
\begin{align*}
	|L_{i}| \leq C_{\beta}\|X_{i}-x\|^{\beta}_\infty\mathds{1}\{X_{i}\in B\} \leq C_\beta h^\beta\mathds{1}\{X_{i}\in B\}.
\end{align*}
Hence, for any $i\in[m]$, we can bound the expectation
\begin{align}
|\mathbb{E}[L_{i}]|
& \leq \mathbb{E}[C_{\beta}h^\beta\mathds{1}\{X_{i}\in B\}] =C_{\beta}h^{\beta}\lambda(B). \label{eq:poly-est-error-ub-exp}
\end{align}
Also, we can derive
\begin{align*}
\max_{i\in[m]}|L_{i}-\mathbb{E}[L_{i}]| & \leq\max_{i\in[m]}|L_{i}|+\max_{i\in[m]}|\mathbb{E}[L_{i}]|\\
 & \leq C_{\beta}h^{\beta}+C_{\beta}h^{\beta}\lambda(B) \leq2C_{\beta}h^{\beta},
\end{align*}
and
\begin{align*}
\sum_{i=1}^{m}\mathsf{Var}(L_{i}) & \leq \sum_{i=1}^{m} \bE [L_{i}^2] \leq \sum_{i=1}^{m} \bE [C_\beta^2 h^{2\beta}\mathds{1}\{X_{i}\in B\}] = mC_{\beta}^{2}h^{2\beta}\lambda(B).
\end{align*}
It follows from the Bernstein inequality that with probability at least $1-O(K^{-1}n^{-10})$,
\begin{align}
\bigg|\sum_{i=1}^{m}(L_{i}-\mathbb{E}[L_{i}])\bigg| &\lesssim C_{\beta}h^{\beta}\sqrt{m\lambda(B)\log(Kn)}+C_{\beta}h^{\beta}\log(Kn) \nonumber \\
& \asymp C_{\beta}h^{\beta}\sqrt{m\lambda(B)\log(Kn)}, \label{eq:poly-est-error-ub-2}
\end{align}
where the last line holds provided $m\lambda(B)\gtrsim\log(Kn)$.
\item Putting (\ref{eq:poly-est-error-ub}), (\ref{eq:poly-est-error-ub-exp}), and (\ref{eq:poly-est-error-ub-2}) together, we know from the union bound that with probability exceeding $1-O(K^{-1}n^{-10})$,
\begin{align}
\bigg| \sum_{i=1}^{m}\big(Y_{i}-f_k(x)\big)\mathds{1}\{X_{i}\in B\}\bigg| 
&\leq\sum_{i=1}^{m}|\mathbb{E}[L_{i}]|+\bigg|\sum_{i=1}^{m}(L_{i}-\mathbb{E}[L_{i}])\bigg|+\Big|\sum_{i=1}^{m}N_{i}\Big|\nonumber \\
 & \leq C_{\beta}h^{\beta}m\lambda(B)+O\Big(C_{\beta}h^{\beta}\sqrt{m\lambda(B)\log(Kn)}\Big)\nonumber \\
 & \quad+O\Big(\sqrt{m\lambda(B)\log(Kn)}\Big)\nonumber \\
 & =\big(1+o(1)\big)C_{\beta}h^{\beta}m\lambda(B)+O\Big(\sqrt{m\lambda(B)\log(Kn)}\Big)\label{eq:lp-error-numer}
\end{align}
provided $\log(Kn)/(m\lambda(B))=o(1)$.
\end{itemize}
\item Combining (\ref{eq:lp-error-denom}) and (\ref{eq:lp-error-numer}), we conclude that with probability exceeding $1-O(K^{-1}n^{-10})$,
\begin{align*}
\big|\widehat{\eta}_k(x;B)-f_k(x)\big| & =\frac{\Big|\sum_{i=1}^{n}\big(Y_{i}-f(x)\big)\mathds{1}\{X_{i}\in B\}\Big|}{\sum_{i=1}^{n}\mathds{1}\{X_{i}\in B\}}\\
 & \leq\frac{\big(1+o(1)\big)C_{\beta}h^{\beta}m\lambda(B)+O\Big(\sqrt{m\lambda(B)\log(Kn)}\Big)}{\big(1-o(1)\big)m\lambda(B)}\\
 & \leq\big(1+o(1)\big)C_{\beta}h^{\beta}+O\bigg(\sqrt{\frac{1}{m\lambda(B)}\log(Kn)}\bigg).
\end{align*}
This completes the proof of Lemma~\ref{lemma:local-poly-reg-est-error}.
\end{itemize}

\subsection{Proof of Lemma~\ref{lemma:local-poly-reg-poly-proj-dist}}
\label{proof-lemma:local-poly-reg-poly-proj-dist}
Fix an arm $k\in[K]$, a bin $B\subset\mathcal{X}$, and a point $x\in B$.
Recall the definition that
\begin{align*}
\Gamma_{B}f_{k}(x;\lambda)\defn\frac{1}{\lambda(B)}\int_{B}f_{k}(u)\,\mathrm{d}\lambda(u).
\end{align*}
We shall adopt the notations defined in the proof of Lemma~\ref{lemma:local-poly-reg-est-error} in Appendix~\ref{proof-lemma:local-poly-reg-est-error}, namely, $M_i\defn\mathds{1}\{X_{i}\in B\}$ for any $i\in[m]$. In addition, let us define a sequence of i.i.d.~random variables
\begin{align*}
\widetilde{N}_{i} & \coloneqq Y_{i}\mathds{1}\{X_{i}\in B\}, \quad i\in[m].
\end{align*}
It is easy to to see that for any $i\in[m]$,
\begin{align}
\mathbb{E}[\widetilde{N}_{i}] &= \bE\big[ \bE[ Y_i\,|\,X_i] \mathds{1}\{X_{i}\in B\}\big] = \bE[f_k(X_i)\mathds{1}\{X_{i}\in B\}] \nonumber \\
& =\int_{B}f_{k}(u)\,\mathrm{d}\lambda(u) \leq \lambda(B), \label{eq:poly-lp-dist-N-ub}
\end{align}
where the last step holds as $f_k$ is bounded in $[0,1]$.
This allows us to express
\begin{align}
\widehat{\eta}_{k}(x;B)-\Gamma_{B}f_{k}(x;\lambda) & =\frac{\sum_{i=1}^{m}Y_{i}\mathds{1}\{X_{i}\in B\}}{\sum_{i=1}^{m}\mathds{1}\{X_{i}\in B\}} -\frac{\int_{B}f_{k}(u)\,\mathrm{d}\lambda(u)}{\lambda(B)} \nonumber \\
 & =\frac{\sum_{i=1}^{m}\widetilde{N}_{i}}{\sum_{i=1}^{m}M_{i}}-\frac{\sum_{i=1}^{m}\bE[\widetilde{N}_{i}]}{\sum_{i=1}^{m}\bE[M_{i}]} \nonumber \\
 & =\frac{\sum_{i=1}^{m}(\widetilde{N}_{i}-\bE[\widetilde{N}_{i}])}{\sum_{i=1}^{n}M_{i}}+\bigg(\sum_{i=1}^{m}\bE[\widetilde{N}_{i}]\bigg)\bigg(\frac{1}{\sum_{i=1}^{m}M_{i}}-\frac{1}{\sum_{i=1}^{m}\bE[M_i]}\bigg) \label{eq:local-poly-reg-poly-proj-dist-temp}.
\end{align}
As shown in (\ref{eq:count-est-error}) and (\ref{eq:lp-error-denom}), if $\log(Kn)/(m\lambda(B))=o(1)$, the following holds with probability at least $1-O(K^{-1}n^{-10})$,
\begin{align*}
\bigg|\sum_{i=1}^{m}(M_{i}-\bE[M_{i}])\bigg| & \lesssim \sqrt{m\lambda(B)\log(Kn)},\\
\sum_{i=1}^{m}M_{i} & \gtrsim m\lambda(B).
\end{align*}
Therefore, the display above gives
\begin{align}
\bigg|\frac{1}{\sum_{i=1}^{m}M_{i}}-\frac{1}{\sum_{i=1}^{m}\bE[M_{i}]}\bigg| & =\frac{\big|\sum_{i=1}^{m}(M_{i}-\bE[M_i])\big|}{\big|\sum_{i=1}^{m}M_{i}\big|\cdot \big|\sum_{i=1}^{m}\bE[M_{i}]\big|} \nonumber \\
& \lesssim\frac{1}{m^2\lambda^2(B)} \sqrt{m\lambda(B)\log(Kn)} \nonumber \\
& =\frac{1}{m^{3/2}\lambda^{3/2}(B)} \sqrt{\log(Kn)}.\label{eq:poly-lp-dist-M-error}
\end{align}

Additionally, since $Y_{i}$'s and $f_{k}$ are bounded in $[0,1]$, it is easy to get 
and $\max_{i\in[m]}|\widetilde{N}_{i}-\mathbb{E}[\widetilde{N}_{i}]|\leq 1$, and 
\begin{align*}
\sum_{i=1}^{m}\mathsf{Var}(\widetilde{N}_{i}) & \leq\sum_{i=1}^{m} \mathbb{E}[\widetilde{N}_{i}^{2}] =\sum_{i=1}^{m}\mathbb{E}\big[\big(Y_{i}-f_{k}(X_{i})\big)^{2}\mathds{1}\{X_{i}\in B\}\big]\leq m\lambda(B).
\end{align*}
It follows from the Bernstein inequality that with probability at least $1-O(K^{-1}n^{-10})$,
\begin{align}
\Big|\sum_{i=1}^{m}(\widetilde{N}_{i}-\bE[\widetilde{N}_{i}])\Big| & \lesssim \sqrt{m\lambda(B)\log(Kn)}+\log(Kn) \nonumber \\
& \asymp \sqrt{m\lambda(B)\log(Kn)},\label{eq:poly-lp-dist-N-error}
\end{align}
as long as $m\lambda(B)\gtrsim\log(Kn)$.

With these bounds in hand, we are ready to control $\widehat{\eta}_{k}(x;B)-\Gamma_{B}f_{k}(x;\lambda)$.
Putting (\ref{eq:lp-error-denom}) and (\ref{eq:poly-lp-dist-N-error}) together and applying the union bound yields that with probability at least $1-O(K^{-1}n^{-10})$,
\begin{align*}
\bigg|\frac{\sum_{i=1}^{m}(\widetilde{N}_{i}-\bE[\widetilde{N}_{i}])}{\sum_{i=1}^{m}M_{i}}\bigg|\lesssim\frac{\sqrt{m\lambda(B)\log(Kn)}}{m\lambda(B)}=\sqrt{\frac{1}{m\lambda(B)}\log(Kn)}.
\end{align*}
In addition, combining (\ref{eq:poly-lp-dist-M-error}) and (\ref{eq:poly-lp-dist-N-ub}) shows that with probability at least $1-O(K^{-1}n^{-10})$,
\begin{align*}
\bigg(\sum_{i=1}^{m}\bE[\widetilde{N}_{i}]\bigg)\bigg(\frac{1}{\sum_{i=1}^{m}M_{i}}-\frac{1}{\sum_{i=1}^{m}\bE[M_i]}\bigg) & \lesssim m \lambda(B)\frac{1}{m^{3/2}\lambda^{3/2}(B)} \sqrt{\log(Kn)} \\
& =\sqrt{\frac{1}{m\lambda(B)}\log(Kn)}.
\end{align*}
Plugging these two inequalities above into \eqref{eq:local-poly-reg-poly-proj-dist-temp} finishes the proof of Lemma~\ref{lemma:local-poly-reg-poly-proj-dist}.

\section{Tabular case}
\label{sec:Tabular}

In this section, we briefly discuss the transfer learning for contextual multi-armed bandits in the tabular setting. The only difference from the framework introduced in Section~\ref{sec:Problem-formulation} in the main text is that the covariate herein takes on values from a finite state space $\mathcal{S} \defn \{1, \dots, S\}$. At each time step $t$, the context $X_{t}^{Q}$ is is drawn i.i.d.~with probability $Q_{X}(s)$ for each $s\in\mathcal{S}$.

In this case, the finite state space can be viewed as a partition of the covariate space in the nonparametric setting considered in Section~\ref{sec:Problem-formulation}, and a state corresponds to a bin in the partition. Therefore, it is straightforward to simplify Algorithm~\ref{alg:UCB-TL} to handle the tabular case. The high-level algorithmic idea remains unchanged: given the finite state space $\mathcal{S}$, each state $s$ is viewed as an index for a collection of static multi-armed bandit problems, and a successive elimination procedure is applied for state $s$ individually.

In order to describe the new algorithm, we need to make minor adjustments to some notations.
To begin with, similar to (\ref{eq:n-k-B-Pdata}) and (\ref{eq:Y-bar-k-B-Pdata}), for each state $s\in\mathcal{S}$ and arm $k\in [K]$, we define 
\begin{align}
	n_{k}^{P}(s) & \defn
		\sum_{(X_{i},\pi_{i},Y_{i})\in\mathcal{D}^{P}}\mathds{1}\{X_{i} = s,\pi_{i}=k\}, \label{eq:n-k-B-Pdata-tabular} \\
	\overline{Y}_{k}^{P}(s) & \defn
		\begin{cases}
		    \frac{1}{n_{k}^{P}(s)}\sum\limits_{(X_{i},\pi_{i},Y_{i})\in\mathcal{D}^{P}}Y_{i} \mathds{1}\{X_{i}=s,\pi_{i}=k\}, & \text{if }  n_{k}^{P}(s) \neq 0, \\
		    0, & \text{otherwise.}
		\end{cases} \label{eq:Y-bar-k-B-Pdata-tabular}
\end{align}
Additionally, for any non-negative integer $\tau \geq 0$, state $s\in\mathcal{S}$ and arm $k\in[K]$, we define
\begin{align}
U_{k}(\tau,s)\coloneqq
\begin{cases}
2\sqrt{\frac{2}{\tau+n_{k}^{P}(s)}\log^{+}(\frac{n_{Q}}{\tau})}, & \text{if } \tau > 0, \\
2\sqrt{\frac{2}{n_{k}^{P}(s)}\log^{+}( n_{Q} \vee n_{P})}, & \text{if } \tau = 0, 
\end{cases}
\label{eq:UCB-tabular}
\end{align}
where $\log^{+}(x)\defn\log(x)\vee1$ and we use the convention $1/0 = \infty$.

With these notations in hand, we are prepared to introduce the transfer learning algorithm for contextual multi-armed bandits in the tabular case. As before, the algorithm begins by identifying the state $s$ to which the current context $X_{t}^{Q}$ belongs. It then proceeds with a successive elimination procedure tailed to the specific state $s$, as detailed in Procedure~\ref{alg:EA-TL-tabular}. The precise steps of the algorithm are presented in Algorithm~\ref{alg:UCB-TL-tabular}. 

Finally, we would like to highlight several notable differences between this algorithm and Algorithm~\ref{alg:UCB-TL} in Section~\ref{sec:Main-Results}. Firstly, due to the finite state space, there is no need to maintain an adaptive partition of the covariate space for approximating the reward functions by their conditional expectations over bins. Consequently, the confidence bound in (\ref{eq:UCB-tabular}) does not contain the second term in (\ref{eq:UCB}) that corresponds to the approximation error. Secondly, for the same reason, Algorithm~\ref{alg:UCB-TL-tabular} directly invokes Procedure~\ref{alg:EA-TL-tabular} at each time step without the necessity of determining whether the current partition needs refinement. In addition, this also eliminates the need to assign an upper bound on the number of pulls for each arm. This is because we no longer need to balance the estimation bias and variance in each bin in order to achieve an adaptive partition. Lastly, the absence of upper bounds on the number of pulls further simplifies Procedure~\ref{alg:EA-TL-tabular} by removing the early stopping stage, which critically depends on the limit on the number of pulls for each arm.

\addtocounter{algorithm}{2}
\begin{algorithm}[t]
\caption{Transfer learning algorithm for tabular contextual multi-armed bandits}
\label{alg:UCB-TL-tabular} 
\begin{algorithmic}[1]
\State{\textbf{Input:} set of arms $\mathcal{I}$, horizon length $n_{Q}$, $P$-data $\mathcal{D}^{P}$.}
\For{$s \in \mathcal{S}$}
	\State{Initialize the policy $\widetilde{\pi}(s)$ by Procedure~\ref{alg:EA-TL-tabular}$\big(s,\mathcal{I},\mathcal{D}^{P} \big)$.}
	\State{Initialize $N(s)\gets0$.}
	    \Comment{initialize time for policy $\widetilde{\pi}(s)$}
\EndFor
\For{$t=1,\dots,n_{Q}$}
	\State{Draw a sample $X_{t}^{Q} \sim Q_{X}$.}
	\State{Denote state $s = X_{t}^{Q}$.}
	\State{Set $N(s)\gets N(s)+1$.}
		\Comment{update times $X_{t}^{Q} = s$}
	\State{Set $\pi_{t}\gets\widetilde{\pi}_{N(s)}(s)$.}
	    \Comment{choose arm by policy $\widetilde{\pi}(s)$}
\EndFor
\State{\textbf{Output:} policy $\{\pi_{t}\}_{t\geq1}$.}
\end{algorithmic}
\end{algorithm}

\addtocounter{algorithm}{-2}
\floatname{algorithm}{Procedure}
\begin{algorithm}[t]
\caption{Successive elimination procedure for a static bandit with source data}
\label{alg:EA-TL-tabular} 
\begin{algorithmic}[1]
\State{\textbf{Input:} state $s$, set of arms $\mathcal{I}$, source data $\mathcal{D}^{P}$.}
\State{Set $n_{k}^{P}(s)$ and $\overline{Y}_{k}^{P}(s)$ as in (\ref{eq:n-k-B-Pdata-tabular}) and (\ref{eq:Y-bar-k-B-Pdata-tabular}), respectively, $\forall k\in\mathcal{I}$.}
\State{Set $n_{k}^{P} \gets n_{k}^{P}(s)$ and $\overline{Y}_{k} \gets \overline{Y}_{k}^{P}(s)$, $\forall k\in\mathcal{I}$.}
\State{Initialize $t\gets0$.}
\State{Initialize $\tau_{k} \gets 0, \forall k\in\mathcal{I}$.} 
    \Comment{initialize pull counts}

\State{Initialize $\underline{Y}^{\star}\gets\max_{k\in\mathcal{I}}\big\{\overline{Y}_{k}-U_{k}(0,s)\big\}$.}
    \Comment{initialize largest reward lower bound}
\Loop
\For{$k\in\mathcal{I}$}
\If{$\overline{Y}_{k}+  U_{k}(\tau_{k},s) \geq\underline{Y}^{\star}$}
    \Comment{eliminate arm s.t.~reward upper bound is smaller than largest reward lower bound}
\State{Set $t \gets t+1$.} 
\State{Select arm $\widetilde{\pi}_{t}\gets k$ and receive reward $Y^{Q,(k)}$.}
\State{Set $\tau_{k}\gets\tau_{k} + 1$.}
    \Comment{update pull count}
\State{Set $\overline{Y}_{k}\gets\frac{1}{n_{k}^{P}+\tau_{k}}\big(Y^{Q,(k)}+(n_{k}^{P}+\tau_{k}-1)\overline{Y}_{k}\big)$.}
    \Comment{update estimated reward}
\State{Set $\underline{Y}^{\star}\gets\max_{k\in\mathcal{I}}\big\{\overline{Y}_{k}-U_{k}(\tau_{k},s)\big\}$.}
    \Comment{update largest reward lower bound}
\Else \State{Eliminate arm $k$ from active arm set: $\mathcal{I} \gets\mathcal{I}\setminus\{k\}$.}
\EndIf
\EndFor
\EndLoop
\State{\textbf{Output:} policy $\{\widetilde{\pi}_{t}\}_{t\geq1}$.}
\end{algorithmic}
\end{algorithm}

\section{Auxiliary lemmas}
This section collects several auxiliary lemmas that are useful for establishing our main theorems.

\begin{lemma}\label{lemma:exist-t-AH}Let $\{X_{i}\}_{i\geq1}$ be
a bounded martingale difference sequence with $X_{i}\in[0,1]$. Then
for any $\delta_{1},\delta_{2}>0$ and integers $T\geq0$, $n\geq1$, one has
\begin{equation}
\mathbb{P}\bigg\{\exists0\leq t\leq T:\sum_{i=1}^{n}X_{i}+\sum_{i=n+1}^{n+t}X_{i}>U(t,n,\delta_{1},\delta_{2})\bigg\}\leq 4T\delta_{1} + \delta_{2}.\label{eq:lemma:exist-t-AH}
\end{equation}
Here, $U(t,n,\delta_{1},\delta_{2})$ is given by
\begin{align*}
   U(t,n,\delta_{1},\delta_{2}) =
   \begin{cases}
    \sqrt{2(n+t)\log^{+}\big(\frac{1}{t \delta_{1}}\big)}, & \text{ if }\,\, t > 0, \\
    \sqrt{2n\log^{+}\big(\frac{1}{\delta_2}\big)}, & \text{ if }\,\, t = 0, 
\end{cases}
\end{align*}
where we recall $\log^{+}(x)\defn \log(x) \vee 1$.
\end{lemma}

\begin{proof}
    See Appendix~\ref{proof-lemma:exist-t-AH}.
\end{proof}

\begin{lemma}
\label{lemma:series-1}
Let $a,b$ be some constants. Suppose that $L>0$ is an integer such that $n2^{-bL}=c$ for some constant $c>0$. Then one has
    \begin{align}
        \sum_{i=0}^{L} 2^{ai}\log(n2^{-bi}) \leq C 2^{aL} = C n^{a/b},
    \end{align}
    for some constant $C>0$ independent of $n$.
\end{lemma}

\begin{proof}
    See Appendix~\ref{proof-lemma:series-1}.
\end{proof}

\begin{lemma}
\label{lemma:series-2}
    Let $a,b$ be some constants. Suppose that $L>0$ is an integer such that $n2^{-bL}=c$ for some constant $c>0$. Then we have
    \begin{align}
        \sum_{i=0}^{L} 2^{-ai}\exp(-n2^{-bi}) \leq C 2^{-aL} = C n^{-a/b},
    \end{align}
    for some constant $C>0$ independent of $n$.
\end{lemma}

\begin{proof}
    See Appendix~\ref{proof-lemma:series-2}.
\end{proof}

\begin{lemma}\label{lemma:KL-Bern}For any $a,b\in[0,1]$, let $\mathsf{Bern}(a)$  and $\mathsf{Bern}(b)$ denote two Bernoulli distributions with parameters $a$ and $b$, respectively. Then one has
\begin{equation}
\mathsf{KL}(\mathsf{Bern}(a) \| \mathsf{Bern}(b)) \leq \frac{(a-b)^2}{b(1-b)}.
\label{eq:lemma:KL-Bern}
\end{equation}
In addition, if $|b-1/2| \leq 1/4$, then one further has
\begin{equation}
\mathsf{KL}(\mathsf{Bern}(a) \| \mathsf{Bern}(b)) \leq 8 (a-b)^2.
\label{eq:lemma:KL-Bern-cor}
\end{equation}
\end{lemma}

\begin{proof}
    See Appendix~\ref{proof-lemma:KL-Bern}.
\end{proof}

\subsection{Proof of Lemma \ref{lemma:exist-t-AH}}
\label{proof-lemma:exist-t-AH}
We start by with the case $T=0$, where one can apply the Azuma-Hoeffding inequality to obtain
\begin{equation}
\mathbb{P}\bigg\{\sum_{i=1}^{n}X_{i}>\sqrt{2n\log^{+}\bigg(\frac{1}{\delta_{2}}\bigg)}\bigg\}\leq\delta_{2}.\label{eq:lemma:exist-t-AH-T}
\end{equation}
This proves (\ref{eq:lemma:exist-t-AH}) for $T=0$.

As for the case with $T>0$, we know from the union bound that

\begin{align*}
  \mathbb{P}&\bigg\{\exists0\leq t\leq T:\sum_{i=1}^{n}X_{i}+\sum_{i=n+1}^{n+t}X_{i}>U(t,n,\delta_{1},\delta_{2})\bigg\}\\
 & \leq\mathbb{P}\Bigg\{\sum_{i=1}^{n}X_{i}>\sqrt{2n\log^{+}\bigg(\frac{1}{\delta_{2}}\bigg)}\Bigg\}+\mathbb{P}\Bigg\{\exists1\leq t\leq T:\sum_{i=1}^{n+t}X_{i}>\sqrt{2(n+t)\log^{+}\bigg(\frac{1}{t\delta_{1}}\bigg)}\Bigg\}.
\end{align*}
The first term is upper bounded (\ref{eq:lemma:exist-t-AH-T}),
so it remains to control the second term. To this end, in view of the maximal Azuma-Hoeffding inequality, for any $\delta_{1}>0$ and integer $T\geq1$, one has
\begin{equation}
\mathbb{P}\Big\{\exists1\leq t\leq T:\sum_{i=1}^{t}X_{i}>\delta_{1}\Big\}\leq\exp\left(-\frac{2\delta_{1}^{2}}{T}\right).\label{eq:maximal Azuma-Hoeffding inequality}
\end{equation}
We can then use the peeling argument to upper bound
\begin{align*}
 & \mathbb{P}\Bigg\{\exists1\leq t\leq T:\frac{1}{n+t}\sum_{i=1}^{n+t}X_{i}>\sqrt{\frac{2}{n+t}\log^{+}\bigg(\frac{1}{t\delta_{1}}\bigg)}\Bigg\}\\
 & \quad\leq\sum_{j=1}^{\lfloor\log_{2}(T)\rfloor}\mathbb{P}\Bigg\{\exists2^{j}\leq t\leq2^{j+1}:\frac{1}{n+t}\sum_{i=1}^{n+t}X_{i}>\sqrt{\frac{2}{n+t}\log^{+}\bigg(\frac{1}{t\delta_{1}}\bigg)}\Bigg\}\\
 & \quad\leq\sum_{j=1}^{\lfloor\log_{2}(T)\rfloor}\mathbb{P}\Bigg\{\exists2^{j}\leq t\leq2^{j+1}:\sum_{i=1}^{n+t}X_{i}>(n+2^{j})\sqrt{\frac{2}{n+2^{j+1}}\log^{+}\bigg(\frac{1}{2^{j+1}\delta_{1}}\bigg)}\Bigg\}\\
 & \quad\overset{(\mathrm{i})}{\leq}\sum_{j=1}^{\lfloor\log_{2}(T)\rfloor}\exp\left(-\frac{2}{n+2^{j+1}}\frac{2(n+2^{j})^{2}}{n+2^{j+1}}\log^{+}\bigg(\frac{1}{2^{j+1}\delta_{1}}\bigg)\right)\\
 & \quad = \sum_{j=1}^{\lfloor\log_{2}(T)\rfloor}\exp\left(-4\bigg(\frac{n+2^{j}}{n+2^{j+1}}\bigg)^2\log^{+}\bigg(\frac{1}{2^{j+1}\delta_{1}}\bigg)\right)\\
 & \quad\overset{(\mathrm{ii})}{\leq}\sum_{j=1}^{\lfloor\log_{2}(T)\rfloor}\exp\left(-\log^{+}\Big(\frac{1}{2^{j+1}\delta_{1}}\Big)\right)\\
 & \quad\leq\sum_{j=1}^{\lfloor\log_{2}(T)\rfloor}2^{j+1}\delta_{1}\leq2^{\log_{2}(T)+2}\delta_{1}=4T\delta_{1},
\end{align*}
where (i) follows from (\ref{eq:maximal Azuma-Hoeffding inequality}); (ii) is true because $\frac{n+2^{j}}{n+2^{j+1}} \geq \frac{1}{2}$ for any $n\geq 0$.

Combining these two terms completes the proof of Lemma~\ref{lemma:exist-t-AH}.

\subsection{Proof of Lemma~\ref{lemma:series-1}}
\label{proof-lemma:series-1}

By defining $S_{i} = \sum_{j=0}^{i} 2^{aj} = \frac{2^{a(i+1)}-1}{2^a-1}$ for any $i\geq 0$, we can express
\begin{align*}
   2^{-bi} =  (2^{ai})^{-b/a} = \bigg(\frac{(2^a-1)S_i + 1}{2^a} \bigg)^{-b/a} = 2^b\big( (2^a-1) S_{i} +1 \big)^{-b/a}.
\end{align*}
This allows us to obtain
\begin{align*}
    \sum_{i=0}^{L} 2^{ai}\log(n2^{-bi}) & =\log(n)+ \sum_{i=1}^{L} (S_{i}-S_{i-1}) \log\Big(n2^b\big( (2^a-1) S_{i} +1 \big)^{-b/a} \Big) \\
    & = \log(n)+ \log (n2^b) \sum_{i=1}^{L} (S_{i}-S_{i-1}) \\
    & \quad -\frac{b}{a} \sum_{i=1}^{L} (S_{i}-S_{i-1}) \log\big( (2^a-1) S_{i} +1 \big) \\
    & \leq \log(n)+ S_L \log (n2^b)  -\frac{b}{a}  \int_{0}^{S_{L}} \log\big( (2^a-1) x +1 \big) \,\mathrm{d}x,
\end{align*}
where the inequality uses $S_0 = 1$ and the fact that $x \mapsto \log\big( (2^a-1) x +1 \big)$ is increasing in $x$ when $a > 0$.
Straightforward calculation gives that
\begin{align*}
    \int_{0}^{S_{L}} \log\big( (2^a-1) x +1 \big) \,\mathrm{d}x & =\Big(\frac{1}{2^a-1}+S_{L} \Big) \log\big((2^a-1)S_L+1\big) - S_L \\
    & \geq  S_{L} \log\big( 2^{a(L+1)} \big) - S_L,
\end{align*}
where we use $(2^a-1)S_L+1 = 2^{a(L+1)}$ and $a,S_L>0$ in the last step.
Combining these two displays implies that
\begin{align*}
    \sum_{i=0}^{L} 2^{ai}\log(n2^{-bi}) 
    & \leq \log(n)+S_L \log (n2^b)  - \frac{b}{a} S_L \log\big(2^{a(L+1)}\big) + \frac{b}{a} S_L  \\
    & = \log(n)+ S_L \log \big(n 2^{-bL} \big) + \frac{b}{a} S_L \\
    & \asymp \log(n) + S_L \asymp n^{a/b},
\end{align*}
where the last line arises from the condition  $n2^{-bL} = c$, and
\begin{align*}
    S_L = \frac{2^{a(L+1)}-1}{2^a-1} \lesssim 2^{aL} \asymp n^{a/b}.
\end{align*}
This finishes the proof of Lemma~\ref{lemma:series-1}.

\subsection{Proof of Lemma~\ref{lemma:series-2}}
\label{proof-lemma:series-2}

Given the condition $n2^{-bL}=c$, we can express
\begin{align*}
    \sum_{i=0}^{L} 2^{-ai}\exp\left(-n2^{-bi}\right) 
    & = \sum_{i=0}^{L} 2^{-ai}\exp\left(-c2^{b(L-i)}\right) \\
    & = 2^{-aL}\sum_{i=0}^{L} 2^{a(L-i)}\exp\left(-c2^{b(L-i)}\right) \\
    & = 2^{-aL}\sum_{i=0}^{L} 2^{ai}\exp(-c2^{bi}) \\
    & \leq 2^{-aL}\sum_{i=0}^{\infty} 2^{ai}\exp(-c2^{bi}) \\
    & \lesssim 2^{-aL},
\end{align*}
where the last line is true because the series $\sum_{i=0}^{m} 2^{ai}\exp(-c2^{bi})$ converges as $m\rightarrow \infty$ when $b>0$.
This completes the proof of Lemma~\ref{lemma:series-2}.

\subsection{Proof of Lemma \ref{lemma:KL-Bern}}
\label{proof-lemma:KL-Bern}
By definition, it is straightforward to bound
\begin{align*}
    \mathsf{KL}(\mathsf{Bern}(a)\,\|\,\mathsf{Bern}(b)) & = a \log\Big(\frac{a}{b}\Big) + (1-a) \log \Big(\frac{1-a}{1-b}\Big) \\
    & \leq a \Big(\frac{a}{b}-1\Big) + (1-a)\Big(\frac{1-a}{1-b} - 1\Big) = \frac{(a-b)^2}{b(1-b)}.
\end{align*}
where the inequality holds since $\log(1+x) \leq x$ for any $x \geq 0$. And the second claim follows as an immediate consequence because $b(1-b) \geq 3/16 $ for any $b$ such that $|b-1/2| \leq 1/4$. This finishes the proof of Lemma \ref{lemma:KL-Bern}.

\bibliographystyle{apalike}
\bibliography{bibfileRL}

\end{document}